\documentclass{llncs}

\usepackage{amssymb} % for maths
\usepackage{amsmath} % for maths
\usepackage{standalone} % for importing diagrams
\usepackage{tikz} % for making diagrams
\usetikzlibrary{fit,positioning} % for making diagrams
\usetikzlibrary{matrix} % for making diagrams
\usepackage{float} % prevent LaTeX from repositioning tables
\restylefloat{table} % prevent LaTeX from repositioning tables given [H] option
\usepackage{url} % for links

\usepackage{float} % table title above
\floatstyle{plaintop} % table title above
\restylefloat{table} % table title above

\graphicspath{{images/}{gm_bnmtf/}{greedy_search/}}

\usepackage{caption} % for limiting caption width
\usepackage{subcaption} % for figure (a) (b) in minipage
\usepackage{booktabs} % for table toprule and bottomrule

\newcommand{\R}{\boldsymbol R}
\newcommand{\E}{\boldsymbol E}
\newcommand{\U}{\boldsymbol U}
\newcommand{\V}{\boldsymbol V}
\newcommand{\F}{\boldsymbol F}
\renewcommand{\S}{\boldsymbol S}
\newcommand{\G}{\boldsymbol G}

\newcommand{\btheta}{\boldsymbol \theta}

\newcommand{\lambdak}{\lambda_{k}}
\newcommand{\lambdaUik}{\lambda_{ik}^U}

\newcommand{\lambdaVjk}{\lambda_{jk}^V}

\newcommand{\lambdaFik}{\lambda_{ik}^F}
\newcommand{\lambdaFk}{\lambda_{k}^F}
\newcommand{\lambdaSkl}{\lambda_{kl}^S}
\newcommand{\lambdaGjl}{\lambda_{jl}^G}
\newcommand{\lambdaGl}{\lambda_{l}^G}

\newcommand{\muUik}{\mu_{ik}^U}
\newcommand{\muVjk}{\mu_{jk}^V}

\newcommand{\tauUik}{\tau_{ik}^U}
\newcommand{\tauVjk}{\tau_{jk}^V}

\newcommand{\expFik}{\widetilde{F_{ik}}}
\newcommand{\expSkl}{\widetilde{S_{kl}}}
\newcommand{\expGjl}{\widetilde{G_{jl}}}

\newcommand{\expFiksqr}{\widetilde{F_{ik}^2}}
\newcommand{\expSklsqr}{\widetilde{S_{kl}^2}}
\newcommand{\expGjlsqr}{\widetilde{G_{jl}^2}}

\newcommand{\expFikp}{\widetilde{F_{ik'}}}

\newcommand{\expSkpl}{\widetilde{S_{k'l}}}
\newcommand{\expSklp}{\widetilde{S_{kl'}}}
\newcommand{\expGjlp}{\widetilde{G_{jl'}}}

\newcommand{\varFik}{\mathrm{Var}_q \left[ F_{ik} \right]}

\newcommand{\varGjl}{\mathrm{Var}_q \left[ G_{jl} \right]}

\newcommand{\sumk}{\sum_{k=1}^K}
\newcommand{\sumexclk}{\sum_{k' \neq k}}
\newcommand{\suml}{\sum_{l=1}^L}
\newcommand{\sumexcll}{\sum_{l' \neq l}}

\newcommand{\expdiffTRI}{\mathbb{E}_q \left[ ( R_{ij} - \F_i \cdot \S \cdot \G_j )^2 \right] }

\newcommand{\diffexpTRI}{ \left( R_{ij} - \sumk \suml \expFik \expSkl \expGjl \right) }

\begin{document}
	
\title{Comparative Study of Inference Methods for Bayesian Nonnegative Matrix Factorisation}%Inference Trade-Offs for \\ Bayesian Matrix Factorisation}
% Inference Trade-Offs for Bayesian Matrix Factorisation
% Effects of Inference Methods in Bayesian Matrix Factorisation
% Comparative Study of Inference Methods for Bayesian Matrix Factorisation
% Bayesian Matrix Factorisation: Study of Inference Effects
% Bayesian Matrix Factorisation: Variational Inference and Trade-Offs
% On Inference for Bayesian Matrix Factorisation
	
\author{Thomas Brouwer$^1$, Jes Frellsen$^2$, Pietro Li\'{o}$^1$}
\institute{$^1$Computer Laboratory, University of Cambridge, United Kingdom \\ $^2$Department of Computer Science, IT University of Copenhagen, Denmark}
%\email{tab43 jf519 pl219 @cam.ac.uk}

\maketitle
\begin{abstract}
	In this paper, we study the trade-offs of different inference approaches for Bayesian matrix factorisation methods, which are commonly used for predicting missing values, and for finding patterns in the data. 
	In particular, we consider Bayesian nonnegative variants of matrix factorisation and tri-factorisation, and compare non-probabilistic inference, Gibbs sampling, variational Bayesian inference, and a maximum-a-posteriori approach. The variational approach is new for the Bayesian nonnegative models. We compare their convergence, and robustness to noise and sparsity of the data, on both synthetic and real-world datasets.
	Furthermore, we extend the models with the Bayesian automatic relevance determination prior, allowing the models to perform automatic model selection, and demonstrate its efficiency.
\end{abstract}

% % % % % % % % % % % % % % % % % % % % % % % % % % % % % % % % % % % % % % % % % % % % % % % % %

\section{Introduction}
\begin{figure}[t]
	\centering
	\includegraphics[width=1\textwidth]{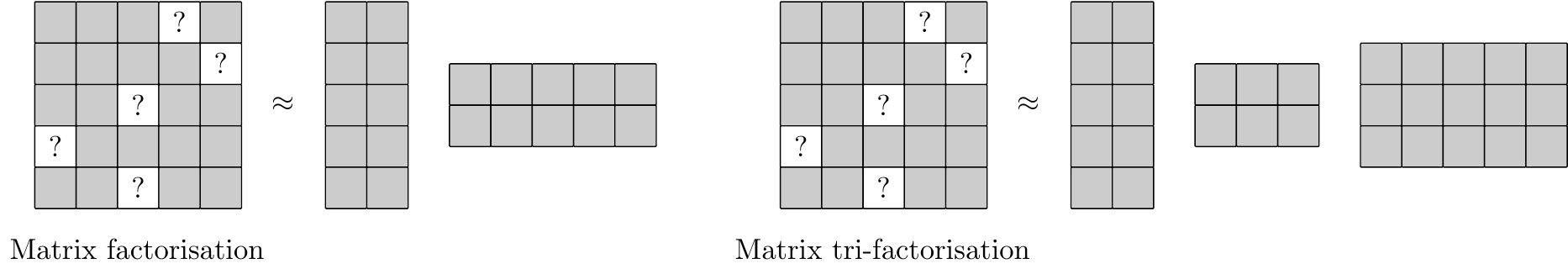}
	\captionsetup{width=1\columnwidth}
	\caption{Overview of matrix factorisation and matrix tri-factorisation methods, with missing values (?-entries).}
	\label{mf_mtf}
\end{figure}

Matrix factorisation methods have been used extensively in recent years to decompose matrices into latent factors, helping us reveal hidden structure and predict missing values. In particular we decompose a given matrix into two smaller matrices so that their product approximates the original one (see Figure \ref{mf_mtf}). Nonnegative matrix factorisation models \cite{Lee1999} have been particularly popular, as the nonnegativity constraint makes the resulting matrices easier to interpret, and is often inherent to the problem---such as in image processing or bioinformatics \cite{Lee1999,Wang2013}. 
A related problem is that of matrix tri-factorisation, first introduced by Ding et al. (2006) \cite{Ding2006}, where the observed dataset is decomposed into three smaller matrices, which again are constrained to be non-negative. 

Both matrix factorisation and tri-factorisation methods have found many applications in recent years, such as for collaborative filtering \cite{Mnih2008,Chen2009}, sentiment classification \cite{Li2009}, predicting drug-target interaction \cite{Gonen2012} and gene functions \cite{MatthiasSchubert2008}, and image analysis \cite{Zhang2005}. Methods can be categorised as either non-probabilistic or Bayesian. For the former, finding the factorisation (\textit{inference}) is commonly done using multiplicative updates, whereas for the latter we use approximate Bayesian inference methods. Non-probabilistic or maximum a posteriori (MAP) solutions give a single point estimate, which can lead to overfitting more easily and neglects uncertainty. Bayesian approaches address this issue, by instead finding a full distribution over the matrices, where we define prior distributions over the matrices and then compute their posterior after observing the actual data. This can greatly reduce overfitting. A key question that arises is: what exactly are the trade-offs between different matrix factorisation inference approaches? In particular, which perform better in terms of speed of convergence, predictive performance, and robustness to noise and sparsity? 

In this paper we answer these questions by performing a thorough empirical study to explore these trade-offs between non-probabilistic and Bayesian inference approaches, which to our knowledge had not been done before.
We consider the popular non-probabilistic matrix factorisation model from Lee and Seung (2000) \cite{Lee2000}, and a Bayesian nonnegative matrix factorisation and tri-factorisation model from Schmidt et al. (2009) \cite{Schmidt2009} and Brouwer and Li\'{o} (2017) \cite{Brouwer2017}, respectively. These models use exponential priors to enforce nonnegativity, giving Gibbs sampling algorithms for inference. The former paper also introduced a MAP algorithm, called iterated conditional modes (ICM). Both of these approaches rely on a sampling procedure to eventually converge to draws of the desired distribution---in this case the posterior of the matrices. This means that we need to inspect the values of the draws to determine when our method has converged (burn-in), and then take additional draws to estimate the posteriors.

We introduce a fourth inference technique for the Bayesian nonnegative models, based on variational Bayesian inference (VB), where instead of relying on random draws we obtain deterministic convergence to a solution. We do this by introducing a new distribution that is easier to compute, and optimise it to be as similar to the true posterior as possible. Some papers (for instance \cite{Salimans2015}) assert that variational inference gives faster but less accurate inference than sampling methods like Gibbs. One study investigating this for latent dirichlet allocation can be found in \cite{Asuncion2009}, but ours is the first paper giving a thorough empirical study of the trade-offs for matrix factorisation.
We furthermore extend the Bayesian models with automatic relevance determination (ARD), to eliminate the need for model selection.

We perform extensive experiments on both artificial and real-world data to explore the trade-offs between speed of inference, and robustness to sparsity and noise for predicting missing values. We show that Gibbs sampling is the most robust, while VB and ICM give significant run-time speedups but sacrifice some robustness, and that non-probabilistic inference tends to be fast but not robust. 
Finally, we show that ARD is an effective way of performing automatic model selection, and increases the robustness of matrix factorisation models if they are given the wrong dimensionality.

Although we study a specific Bayesian nonnegative matrix factorisation and tri-factorisation model, we believe that many of our findings and insights apply to the broad range of other matrix factorisation and tri-factorisation methods, as well as tensor and Tucker decomposition methods---their three-dimensional extensions.

% % % % % % % % % % % % % % % % % % % % % % % % % % % % % % % % % % % % % % % % % % % % % % % % % 

\section{Models}

\subsection{Nonnegative Matrix Factorisation}
	We follow the notation used by Schmidt et al. (2009) \cite{Schmidt2009} for nonnegative matrix factorisation (NMF), which can be formulated as decomposing a matrix $ \R \in \mathbb{R}^{I \times J} $ into two latent (unobserved) matrices $ \U \in \mathbb{R}_+^{I \times K} $ and $ \V \in \mathbb{R}_+^{J \times K} $, whose values are constrained to be positive. In other words, solving $ \R = \U \V^T + \E $, where noise is captured by matrix $ \E \in \mathbb{R}^{I \times J} $. The dataset $ \R $ need not be complete---the indices of observed entries can be represented by the set $ \Omega = \left\{ (i,j) \text{ $ \vert $ $ R_{ij} $ is observed} \right\} $. These entries can then be predicted by $\U \V^T$.
	
	We take a probabilistic approach to this problem. We express a likelihood function for the observed data, and treat the latent matrices as random variables. As the likelihood we assume each value of $ \R $ comes from the product of $ \U $ and $ \V $, with some Gaussian noise added,
	\begin{equation*}
		R_{ij} \sim \mathcal{N} (R_{ij} | \boldsymbol U_i \cdot \boldsymbol V_j, \tau^{-1} )
	\end{equation*}
	\noindent where $ \boldsymbol U_i, \boldsymbol V_j $ denote the $i$th and $j$th rows of $\U$ and $\V $, and $ \mathcal{N} (x|\mu,\tau) = \tau^{\frac{1}{2}} (2\pi)^{-\frac{1}{2}} \exp \left\{ -\frac{\tau}{2} (x - \mu)^2 \right\} $ is the density of the Gaussian distribution with precision $ \tau $. The set of parameters for our model is denoted $ \btheta = \left\{ \U, \V, \tau \right\} $.
	In the Bayesian approach to inference, we want to find the distributions over parameters $ \btheta $ after observing the data $ D = \lbrace R_{ij} \rbrace_{i,j \in \Omega} $. We can use Bayes' theorem, 
	\begin{equation*}
		p(\btheta|D) \propto p(D|\btheta) p(\btheta).
	\end{equation*}
	We need priors over the parameters, allowing us to express beliefs for their values---such as constraining $ \U, \V $ to be nonnegative. We can normally not compute the posterior $ p(\btheta|D) $ exactly, but some choices of priors allow us to obtain a good approximation. Schmidt et al. choose an exponential prior over $ \U $ and $ \V $, so that each element in $ U $ and $ V $ is assumed to be independently exponentially distributed with rate parameters $ \lambdaUik, \lambdaVjk > 0 $. 
	\begin{alignat*}{2}
		U_{ik} \sim \mathcal{E} ( U_{ik} | \lambdaUik)		\quad\quad		V_{jk} \sim \mathcal{E} ( V_{jk} | \lambdaVjk)
	\end{alignat*}
	\noindent where $ \mathcal{E} ( x | \lambda ) = \lambda \exp \left\{ - \lambda x \right\} u(x) $ is the density of the exponential distribution, and $u(x)$ is the unit step function. For the precision $ \tau $ we use a Gamma distribution with shape $ \alpha_{\tau} > 0 $ and rate $ \beta_{\tau} > 0 $,
	\begin{equation*}
		p(\tau) \sim \mathcal{G} (\tau | \alpha_{\tau}, \beta_{\tau} ) = \frac{{\beta_{\tau}}^{\alpha_{\tau}}}{\Gamma(\alpha_{\tau})} x^{\alpha_{\tau} -1} e^{- \beta_{\tau} x}
	\end{equation*}
	\noindent where $ \Gamma(x) = \int_{0}^{\infty} x^{t-1} e^{-x} dt $ is the gamma function. 
	
\subsection{Nonnegative Matrix Tri-Factorisation}
	The problem of nonnegative matrix tri-factorisation (NMTF) can be formulated similarly to that of nonnegative matrix factorisation, and was introduced by Brouwer and Li\'{o} (2017) \cite{Brouwer2017}. We now decompose $\R$ into three matrices $ \F \in \mathbb{R}_+^{I \times K} $, $ \S \in \mathbb{R}_+^{K \times L} $, $ \G \in \mathbb{R}_+^{J \times L} $, so that $ \R = \F \S \G^T + \E $. This decomposition has the advantage of extracting row and column factor values separately (through $\F$ and $\G$), allowing us to identify both row and column clusters. We again use a Gaussian likelihood and Exponential priors for the latent matrices.
	\begin{alignat*}{2}
	&R_{ij} \sim \mathcal{N} (R_{ij} | \F_i \cdot \S \cdot \G_j, \tau^{-1} )	 		&&\tau \sim \mathcal{G} (\tau | \alpha_{\tau}, \beta_{\tau} ) 		\\
	&F_{ik} \sim \mathcal{E} ( F_{ik} | \lambdaFik)		\quad\quad		S_{kl} \sim \mathcal{E}( S_{kl} | \lambdaSkl)		\quad\quad	&&G_{jl} \sim \mathcal{E}( G_{jl} | \lambdaGjl)
	\end{alignat*}
	
	\begin{figure}[t]
		\centering
		\includegraphics[height=130pt]{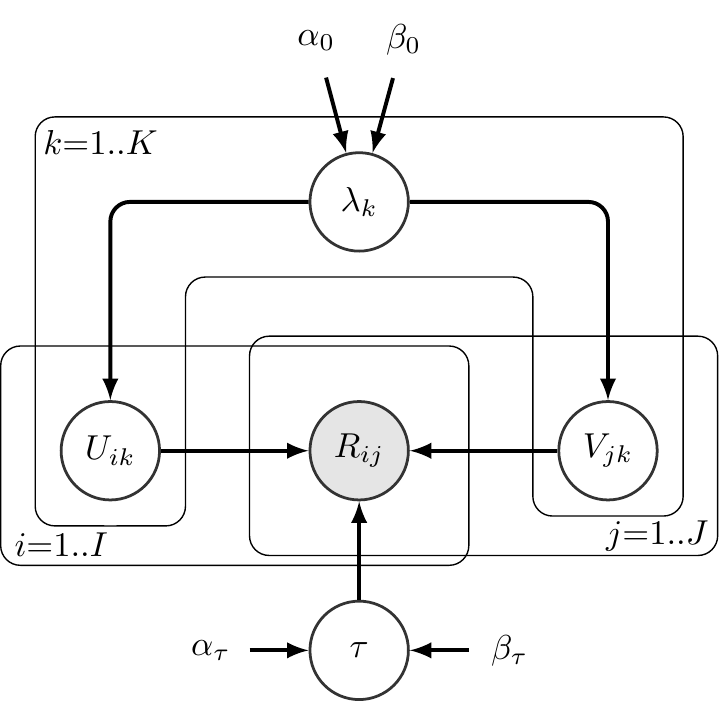}
		\includegraphics[height=135pt]{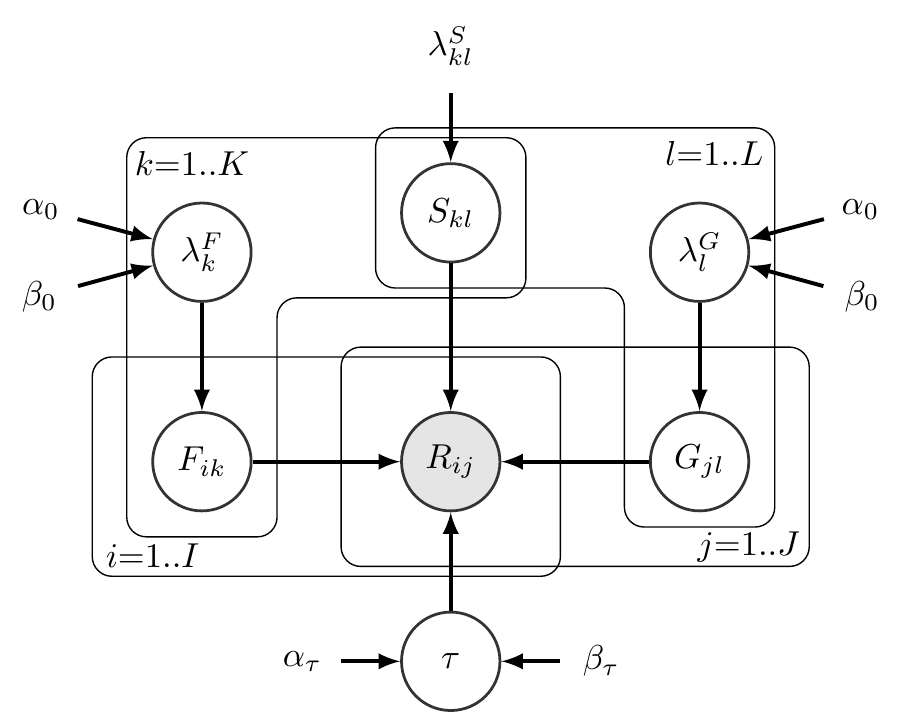}
		\captionsetup{width=1\columnwidth}
		\caption{Graphical model representation of Bayesian nonnegative matrix factorisation (left) and tri-factorisation (right), with ARD.}
		\label{gm_bnmtf}
	\end{figure}

\subsection{Automatic Relevance Determination}
	Automatic relevance determination (ARD) is a Bayesian prior which helps perform automatic model selection. It works by replacing the individual $\lambda$ parameters in the factor matrix priors by one that is shared by all entries in the same column (in other words, shared for each factor). We then place a further Gamma prior over all these $\lambda_k$ parameters. For the NMF model, the priors become
	\begin{alignat*}{2}
		U_{ik} \sim \mathcal{E} ( U_{ik} | \lambdak)		\quad\quad		V_{jk} \sim \mathcal{E} ( V_{jk} | \lambdak)	\quad\quad		\lambda_k \sim \mathcal{G} (\lambda_k | \alpha_0, \beta_0 ).
	\end{alignat*}
	Since this parameter is shared by all entries in the same column, the entire factor $k$ is either activated (if $\lambda_k^t$ has a low value) or ``turned off'' (if $\lambda_k^t$ has a high value), pushing factors that are active for only a few entities further to zero.
	This prior has been used for both real-valued \cite{Virtanen2011,Virtanen2012} and nonnegative matrix factorisation \cite{Tan2013}. Instead of having to choose the correct $K$, we give an upper bound and the model will automatically determine the number of factors to use. 
	A similar approach can be found in \cite{Figueiredo2002}, which incorporates the elimination of unused factors into their expectation-maximisation inference algorithm. ARD is implemented on a model level, and therefore works with all inference approaches.
	
	For NMTF we use two ARD's, one for $\F$ ($\lambdaFk$) and another for $\G$ ($\lambdaGl$),
	\begin{alignat*}{2}
		%&F_{ik} \sim \mathcal{E} ( F_{ik} | \lambdaFk)		\quad\quad\quad		S_{kl} \sim \mathcal{E}( S_{kl} | \lambdaSkl)			\quad\quad		&&G_{jl} \sim \mathcal{E}( G_{jl} | \lambdaGl) \\
		%&\lambdaFk \sim \mathcal{G} (\lambdaFk | \alpha_0, \beta_0 ) 		\quad\quad 		&&\lambdaGl \sim \mathcal{G} (\lambdaGl | \alpha_0, \beta_0 ).
		F_{ik} \sim \mathcal{E} ( F_{ik} | \lambdaFk)		\quad			\lambdaFk \sim \mathcal{G} (\lambdaFk | \alpha_0, \beta_0 )			\quad\quad		&&G_{jl} \sim \mathcal{E}( G_{jl} | \lambdaGl) 			\quad			\lambdaGl \sim \mathcal{G} (\lambdaGl | \alpha_0, \beta_0 ). 
	\end{alignat*}
	The graphical models for Bayesian NMF and NMTF are given in Figure \ref{gm_bnmtf}.
	
% % % % % % % % % % % % % % % % % % % % % % % % % % % % % % % % % % % % % % % % % % % % % % % % % 

\section{Inference}
In this section we give details for four different types of inference for nonnegative matrix factorisation (NMF) and tri-factorisation (NMTF) models. Non-probabilistic inference gives a point estimate solution. Gibbs sampling and variational Bayesian inference both give a full posterior estimate, whereas iterated conditional modes gives a maximum a posteriori (MAP) point estimate. 

\subsection{Non-Probabilistic Inference}
	A non-probabilistic (NP) approach for NMF can be found in Lee and Seung (2000) \cite{Lee2000}. Their algorithm relies on multiplicative updates, where at each iteration the values in the $\U$ and $\V$ matrices are updated using the following values:
	\begin{equation*}
		\displaystyle U_{ik} = U_{ik} \frac{\sum_{j \in \Omega_i} R_{ij} V_{jk} / (\U_i\V_j)}{\sum_{j \in \Omega_i} V_{jk}} 		\quad\quad\quad		 V_{jk} = V_{jk} \frac{\sum_{i \in \Omega_j} R_{ij} U_{ik} / (\U_i\V_j)}{\sum_{i \in \Omega_j} U_{ik}}
	\end{equation*}
	where $ \Omega_i = \left\{ j \text{ $ \vert $ } (i,j) \in \Omega \right\} $ and  $ \Omega_j = \left\{ i \text{ $ \vert $ } (i,j) \in \Omega \right\} $.
	These updates can be shown to minimise the I-divergence (generalised KL-divergence),
	\begin{equation*}
		D(\R||\U\V^T) = \sum_{(i,j) \in \Omega} \left( R_{ij} \log \frac{R_{ij}}{(\U\V^T)_{ij}} - R_{ij} + (\U\V^T)_{ij} \right).
	\end{equation*}
	Yoo and Choi (2009) \cite{Yoo2009} extended this approach to NMTF, giving the following multiplicative updates, with $\S_{\cdot l}$ denoting the $l$th column of $\S$:
	\begin{alignat*}{1}
		&F_{ik} = F_{ik} \frac{\sum_{j \in \Omega_i} R_{ij} (\S_k \G_j) / (\F_i \S \G_j)}{\sum_{j \in \Omega_i} (\S_k \G_j)} 		\quad		 G_{jl} = G_{jl} \frac{\sum_{i \in \Omega_j} R_{ij} (\F_i S_{\cdot l}) / (\F_i \S \G_j)}{\sum_{i \in \Omega_j} (\F_i S_{\cdot l})} \\
		& \quad\quad\quad\quad\quad\quad\quad\quad S_{kl} = S_{kl} \frac{\sum_{(i,j) \in \Omega} R_{ij} F_{ik} G_{jl} / (\F_i \S \G_j)}{\sum_{(i,j) \in \Omega} F_{ik} G_{jl}}.
	\end{alignat*}
	
\subsection{Gibbs Sampling}
	Schmidt et al. \cite{Schmidt2009} introduced a Gibbs sampling algorithm for approximating the posterior distribution---a similar NMF model that uses Gibbs sampling can be found in \cite{Zhong2009,Zhong2011}. Gibbs sampling works by sampling new values for each parameter $ \theta_i $ from its marginal distribution given the current values of the other parameters $ \btheta_{-i} $, and the observed data $ D $. If we sample new values in turn for each parameter $ \theta_i $ from $ p(\theta_i | \btheta_{-i}, D ) $, we will eventually converge to draws from the posterior, which can be used to approximate the posterior $ p(\btheta|D) $. We have to discard the first $n$ draws because it takes a while to converge (\textit{burn-in}), and since consecutive draws are correlated we only use every $i$th value (\textit{thinning}). 
	
	For NMF this means that we need to be able to draw from distributions
	\begin{alignat*}{2}
	&p(\tau| \U, \V, \boldsymbol \lambda, D)			\quad\quad 		&&p(U_{ik}|\tau, \U_{-ik}, \V, \boldsymbol \lambda, D)	\\				
	&p(\lambdak| \tau, \U, \V, D)		 \quad\quad		&&p(V_{jk}|\tau, \U, \V_{-jk}, \boldsymbol \lambda, D).
	\end{alignat*}
	where $\U_{-ik}$ denotes all elements in $\U$ except $U_{ik}$, and similarly for $\V_{-jk}$. $\boldsymbol \lambda$ is a vector including all $\lambda_k$ values. Using Bayes theorem we obtain the following posterior distributions:
	\begin{alignat*}{2}
	&p(\tau| \U, \V, \boldsymbol \lambda, D) = \mathcal{G} (\tau | \alpha^*_{\tau}, \beta^*_{\tau} ) 			\quad		&&p(U_{ik}|\tau, \U_{-ik}, \V, \boldsymbol \lambda, D) = \mathcal{TN} ( U_{ik} | \muUik, \tauUik )	 \\
	&p(\lambdak| \tau, \U, \V, D) = \mathcal{G} (\lambdak | \alpha_k^*, \beta_k^* ) 		\quad			&&p(V_{jk}|\tau, \U, \V_{-jk}, \boldsymbol \lambda, D) = \mathcal{TN} ( V_{jk} | \muVjk, \tauVjk )
	\end{alignat*}
	where 
	\begin{equation*}
	\mathcal{TN} ( x | \mu, \tau ) = \left\{
	\begin{array}{ll}
	\displaystyle \frac{ \sqrt{ \frac{\tau}{2\pi} } \exp \left\{ -\frac{\tau}{2} (x - \mu)^2 \right\} }{ 1 - \Phi ( - \mu \sqrt{\tau} )}  & \mbox{if } x \geq 0 \\
	0 & \mbox{if } x < 0
	\end{array}
	\right.
	\end{equation*}
	is a truncated normal: a normal distribution with zero density below $ x = 0 $ and renormalised to integrate to one. $ \Phi(\cdot) $ is the cumulative distribution function of $ \mathcal{N}(0,1) $.
	
	For NMTF we can derive a Gibbs sampling algorithm similarly, as done by Brouwer and Li\'{o} \cite{Brouwer2017}. The posteriors, together with the parameter values for both Gibbs samplers, are given in the supplementary materials.

\subsection{Iterated Conditional Modes}
	The iterated conditional models (ICM) algorithm for inference in the NMF model was given in Schmidt et al. \cite{Schmidt2009}. It works very similarly to the Gibbs sampler, but instead of randomly drawing a value from the conditional posteriors, we take the mode at each iteration. This gives a maximum a posteriori (MAP) point estimate $\btheta_{\text{MAP}} = \max_{\btheta} p(\btheta|D)$, rather than a full posterior distribution. We furthermore still need to use thinning and burn-in. For random variables $ X \sim \mathcal{G}(a,b) $, $ Y \sim \mathcal{TN}(\mu,\tau) $, the modes are $\frac{a-1}{b}$ and $\max{(0,\mu)}$, respectively.
	
	In practice ICM often converges to solutions where multiple columns in the matrices are all zeros, leading to poor approximations. We have addressed this issue by resetting zeros to a small positive value like $ 0.1 $ at each iteration.

\subsection{Variational Bayesian Inference}
	Variational Bayesian inference (VB) has been used for other matrix factorisation models before \cite{Gonen2012}, but not for the nonnegative model in this paper. We therefore now introduce a new VB algorithm for our model. Like Gibbs sampling, VB is a way to approximate the true posterior $ p(\btheta|D) $. The idea behind VB is to introduce an approximation $q(\btheta)$ to the true posterior that is easier to compute, and to make our variational distribution $q(\btheta)$ as similar to $ p(\btheta|D) $ as possible (as measured by the KL-divergence). 
	We assume the variational distribution $ q(\btheta) $ factorises completely, so all variables are independent in the posterior,
	\begin{equation*}
	q(\btheta) = \prod_{\theta_i \in \btheta} q(\theta_i).
	\end{equation*} 
	This is called the mean-field assumption. We use the same forms of $ q(\theta_i) $ as we used in Gibbs sampling,
	\begin{alignat*}{2}
	&q(\tau) = \mathcal{G} (\tau | \alpha^*_{\tau}, \beta^*_{\tau} ) \quad\quad &&q(\lambda_k) = \mathcal{G} (\lambda_k | \alpha_k^*, \beta_k^* ) \\
	&q(U_{ik}) = \mathcal{TN} ( U_{ik} | \muUik, \tauUik ) \quad\quad &&q(V_{jk}) = \mathcal{TN} ( V_{jk} | \muVjk, \tauVjk ).
	\end{alignat*}
	It can be shown \cite{J.M.Bernardo} that the optimal distribution for the $i$th parameter, $q^*(\theta_i)$, can be expressed as follows (for some constant $C$), allowing us to find the optimal updates for the variational parameters.
	\begin{equation*}
	\log q^*(\theta_i) = \mathbb{E}_{q(\btheta_{-i})} \left[ \log p(\btheta, D) \right] + C.
	\end{equation*}
	We now take the expectation with respect to the distribution $ q(\btheta_{-i}) $ over the parameters but excluding the $i$th one. This gives rise to an iterative algorithm: for each parameter $\theta_i$ we update its distribution to that of its optimal variational distribution, and then update the expectation and variance with respect to $ q $. We therefore need updates for the variational parameters, and to be able to compute the expectations and variances of the random variables. This algorithm is guaranteed to maximise the Evidence Lower Bound (ELBO) 
	%$\mathcal{L} =  \mathbb{E}_{q} \left[ \log p(\btheta, D) - \log q(\btheta) \right]$,
	%
	\begin{equation*}
	\mathcal{L} =  \mathbb{E}_{q} \left[ \log p(\btheta, D) - \log q(\btheta) \right],
	\end{equation*}
	which is equivalent to minimising the KL-divergence.
	
	We use $ \widetilde{f(X)} $ as a shorthand for $ \mathbb{E}_q \left[ f(X) \right] $, where $X$ is a random variable and $f$ is a function over $X$. For random variables $ X \sim \mathcal{G}(a,b) $ and $ Y \sim \mathcal{TN}(\mu,\tau) $ the variance and expectation are 
	\begin{alignat*}{1}
	\widetilde{X} = \frac{a}{b}		\quad\quad\quad 		\widetilde{Y} = \mu + \frac{1}{\sqrt{\tau}} \lambda \left( - \mu \sqrt{ \tau } \right)		\quad\quad\quad		\mathrm{Var} \left[ Y \right] = \frac{1}{\tau} \left[ 1 - \delta \left( - \mu \sqrt{ \tau } \right) \right],
	\end{alignat*}
	\noindent where $ \psi(x) = \frac{d}{dx} \log \Gamma(x) $ is the digamma function, $ \lambda(x) = \phi(x) / [ 1 - \Phi(x) ] $, and $ \delta(x) = \lambda(x) [ \lambda(x) - x ] $. $ \phi(x) = \frac{1}{\sqrt{2\pi}} \exp \lbrace - \frac{1}{2} x^2 \rbrace $ is the density function of $ \mathcal{N}(0,1) $. 
	
	The updates for NMF are given in the supplementary materials. Our VB algorithm for NMTF follows the same steps as before, but now has an added complexity due to the term $ \expdiffTRI $. Before, all covariance terms for $ k' \neq k $ were zero due to the factorisation in $ q $, but we now obtain some additional non-zero covariance terms:
	
	\begin{alignat}{1}
	\nonumber \expdiffTRI = &\diffexpTRI^2 \\
	&	\quad	+ \sumk \suml \mathrm{Var}_q \left[ F_{ik} S_{kl} G_{jl} \right] \\
	&	\quad	+ \sumk \suml \sumexclk \mathrm{Cov} \left[ F_{ik} S_{kl} G_{jl}, F_{ik'} S_{k'l} G_{jl} \right] \\
	&	\quad	+ \sumk \suml \sumexcll \mathrm{Cov} \left[ F_{ik} S_{kl} G_{jl}, F_{ik} S_{kl'} G_{jl'} \right].
	\end{alignat}
	The above variance and covariance terms are equal to the following, respectively, leading to the variational updates given in the supplementary materials.
	\setcounter{equation}{0}
	%\begin{alignat}{1}
	%	& \expFiksqr \expSklsqr \expGjlsqr - \expFik^2 \expSkl^2 \expGjl^2 \\
	%	& \varFik \expSkl \expGjl \expSklp \expGjlp \\	
	%	& \expFik \expSkl \varGjl \expFikp \expSkpl.
	%\end{alignat}
	\begin{alignat*}{1}
		\expFiksqr \expSklsqr \expGjlsqr - \expFik^2 \expSkl^2 \expGjl^2,  				\quad 				 \varFik \expSkl \expGjl \expSklp \expGjlp,  					\quad 					\expFik \expSkl \varGjl \expFikp \expSkpl.
	\end{alignat*}

%\section{Model discussion}
\subsection{Complexity}
	Each of the four approaches have the same time complexities, but vary in how efficiently the updates can be computed, and how quickly they converge. The time complexity per iteration for NMF is $ \mathcal{O}( I J K^2 ) $, and $ \mathcal{O}( I J (K^2 L + K L^2) ) $ for NMTF. However, the updates in each column of $ \U, \V, \F, \G $ are independent of each other and can therefore be updated in parallel. 
	For Gibbs and ICM this means we can draw these values in parallel, but for VB and NP we can jointly update the columns using a single matrix operation. Modern computer architectures can exploit this using vector processors, leading to a great speedup.
	
	Furthermore, after the VB algorithm converges we have our approximation to the posterior distributions immediately, whereas with Gibbs and ICM we need to obtain further draws after convergence and use a thinning rate to obtain an accurate MAP (ICM) or posterior (Gibbs) estimate. This deterministic behaviour of VB and NP makes them easier to use. Although additional variables need to be stored to represent the posteriors, this does not result in a worse space complexity, as the Gibbs sampler needs to store draws over time.

\subsection{Initialisation}
	Initialising the parameters of the models can vastly influence the quality of convergence. This can be done by using the hyperparameters $ \lambdaUik $, $\lambdaVjk$, $\lambdaFik$, $\lambdaSkl$, $\lambdaGjl$, $\alpha$, $\beta$, $\alpha_0$, $\beta_0$, $\alpha^F_0$, $\beta^F_0$, $\alpha^G_0$, $\beta^G_0$ to set the initial values to the mean of the priors of the model, or using random draws. We found that random draws tend to give faster and better convergence than the expectation, as it provides a better initial guess of the right patterns in the matrices.
	For matrix tri-factorisation we can initialise $ \F $ by running the K-means clustering algorithm on the rows as datapoints, and similarly $ \G $ for the columns, as suggested by Ding et al. (2006) \cite{Ding2006}. For the VB and NP algorithms we then set the $ \mu $ parameters to the cluster indicators, and for Gibbs and ICM we add $ 0.2 $ for smoothing. We found that this improved the convergence as well, with $ \S $ initialised using random draws.

\subsection{Software}
	Implementations of all methods, the datasets, and experiments described in the next section, are available at \url{https://github.com/ThomasBrouwer/BNMTF_ARD}.

% % % % % % % % % % % % % % % % % % % % % % % % % % % % % % % % % % % % % % % % % % % % % % % % % 

\section{Experiments}
	To demonstrate the trade-offs between the four inference methods presented, we conducted experiments on synthetic data and four real-world drug sensitivity datasets. We compare the convergence speed, robustness to noise, and robustness to sparsity. 

\subsection{Datasets}
	For the synthetic datasets we generated the latent matrices using unit mean exponential distributions, and adding zero mean unit variance Gaussian noise to the resulting product. For the matrix factorisation model we used $ I = 100, J = 80, K = 10 $, and for the matrix tri-factorisation $ I = 100, J = 80, K = 5, L = 5 $.
	
	We considered four drug sensitivity datasets, each detailing the effectiveness ($IC_{50}$ or $EC_{50}$ values) of a range of drugs on different cell lines for cancer and tissue types, where some of the entries are missing. We consider the Genomics of Drug Sensitivity in Cancer (GDSC v5.0 \cite{Yang2013}, $IC_{50}$), Cancer Therapeutics Response Portal (CTRP v2 \cite{Seashore-Ludlow2015}, $EC_{50}$), and Cancer Cell Line Encyclopedia (CCLE \cite{Barretina2012}, $IC_{50}$ and $EC_{50}$). The four datasets are summarised in Table \ref{summary_drug_sensitivity}, giving the number of cell lines, drugs, and the fraction of entries that are observed. 
	
	In some experiments we focused on a selection of the datasets, but results for all can be found in the supplementary materials, together with preprocessing details. For all models we used weak priors ($\lambda = 0.1, \alpha_{\tau} = \beta_{\tau} = \alpha_0 = \beta_0 = 1$).
	
	\begin{table*}[t]
		\captionsetup{width=\columnwidth}
		\caption{Overview of the four drug sensitivity datasets, giving the number of cell lines (rows), drugs (columns), and the fraction of entries that are observed.} \label{summary_drug_sensitivity}
		\centering
		\begin{tabular}{lccc}
			\toprule
			Dataset \hspace{25pt} & Cell lines & \hspace{5pt} Drugs \hspace{5pt} & Fraction observed \\
			\midrule
			GDSC $IC_{50}$ & 707 & 139 & 0.806 \\
			CTRP $EC_{50}$ & 887 &  545 & 0.801 \\
			CCLE $IC_{50}$ & 504 & 24 & 0.965 \\
			CCLE $EC_{50}$ & 504 & 24 & 0.630 \\
			\bottomrule
		\end{tabular}
	\end{table*}
	
\subsection{Convergence Speed}
\begin{figure*}[t]
		\centering
		\begin{subfigure}[t]{\columnwidth}
			\hspace{33pt}
			\includegraphics[width=0.8\columnwidth]{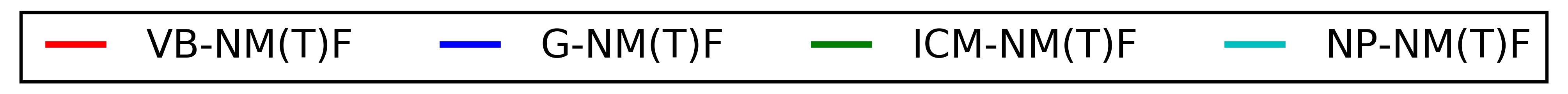}
			\vspace{3pt}
		\end{subfigure}
		\begin{subfigure}[t]{0.19 \columnwidth}
			\includegraphics[width=\columnwidth]{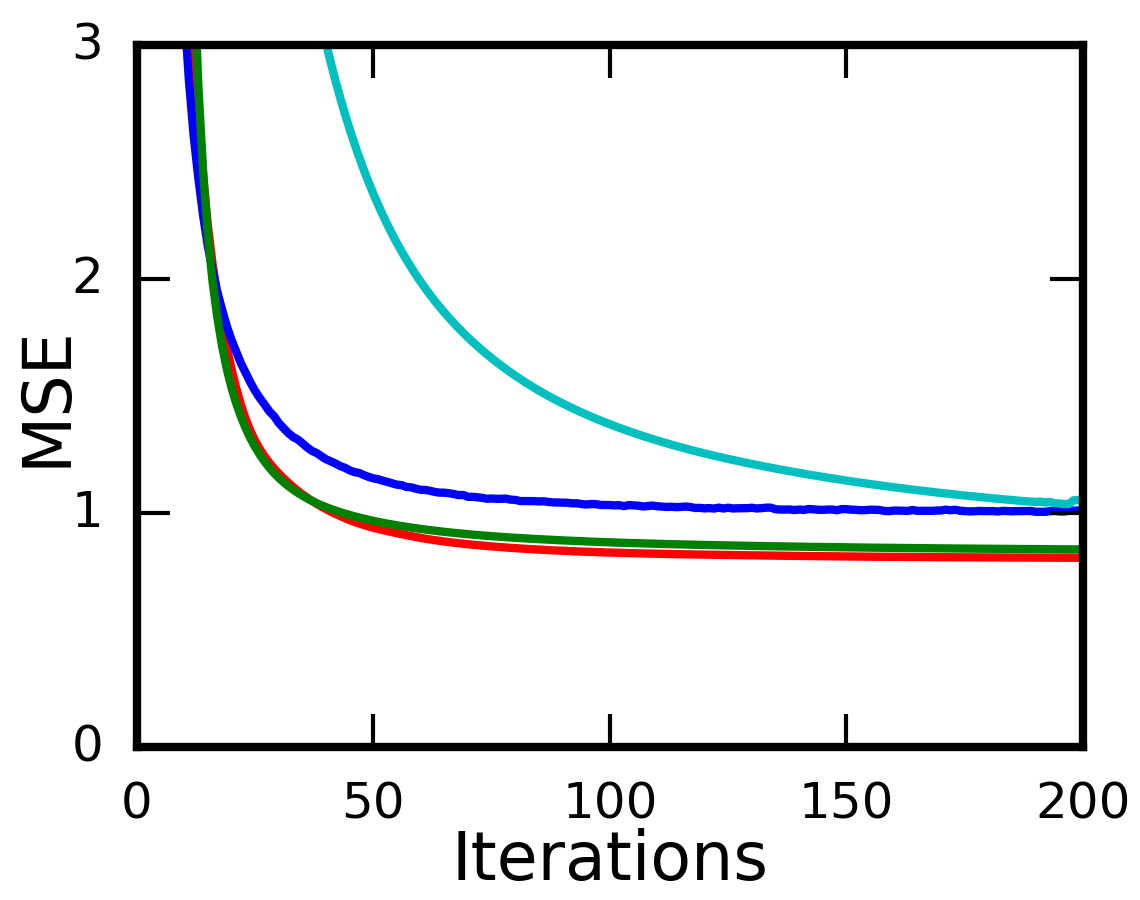}
		\end{subfigure} %
		\begin{subfigure}[t]{0.19 \columnwidth}
			\includegraphics[width=\columnwidth]{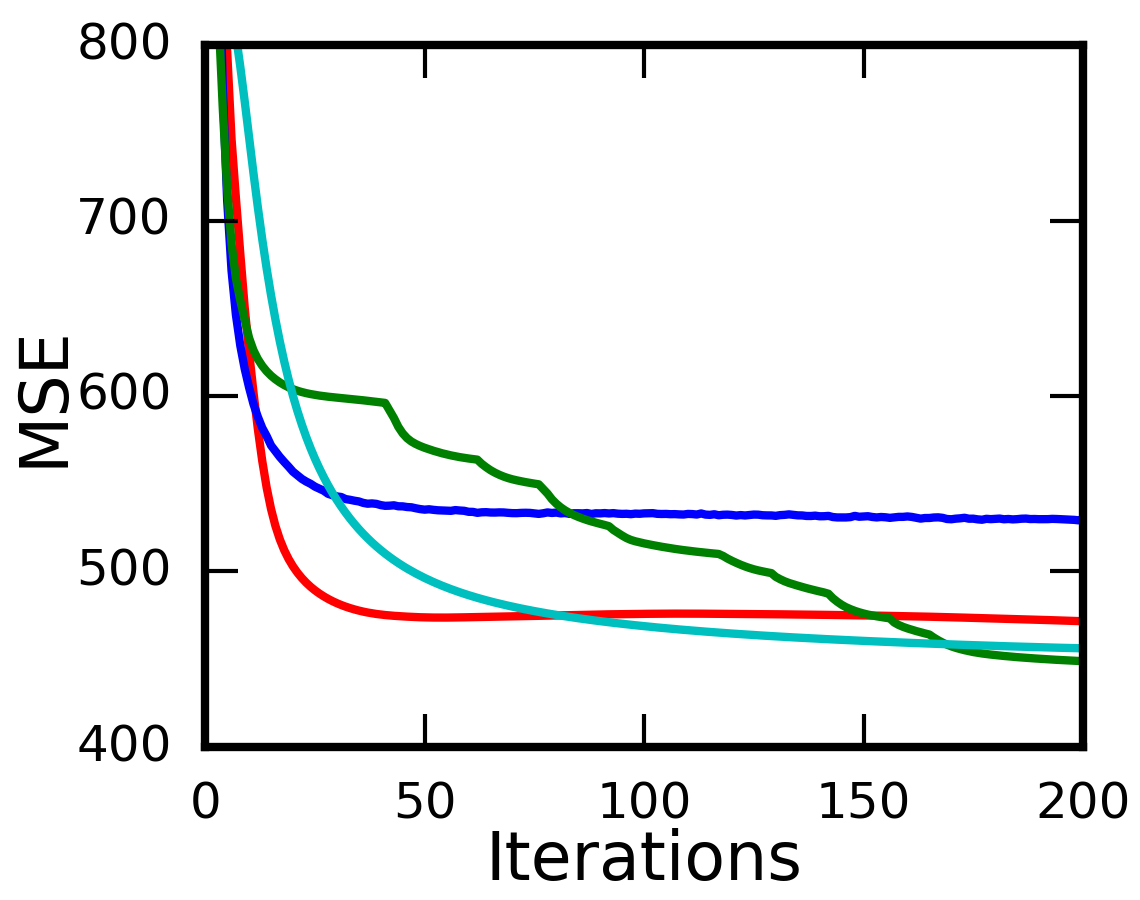}
		\end{subfigure} %
		\begin{subfigure}[t]{0.19 \columnwidth}
			\includegraphics[width=\columnwidth]{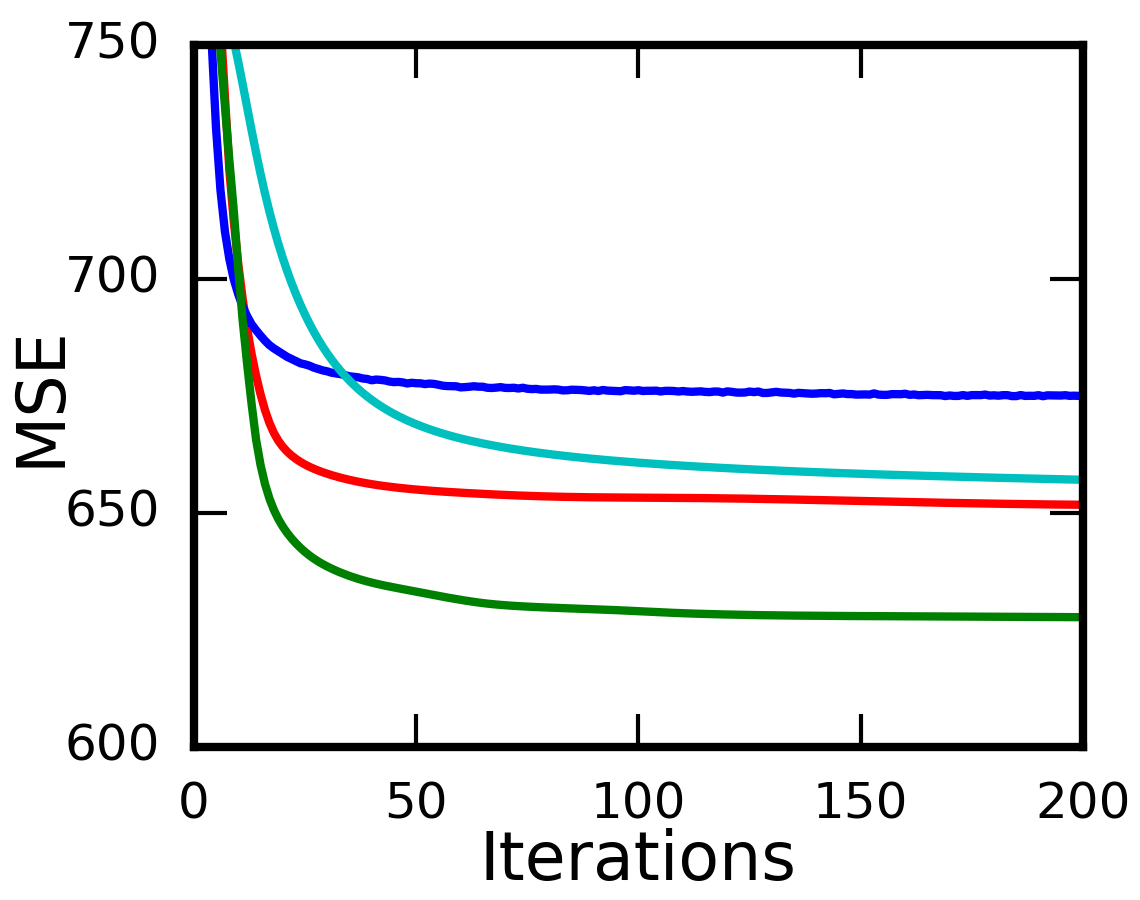}
		\end{subfigure} %
		\begin{subfigure}[t]{0.19 \columnwidth}
			\includegraphics[width=\columnwidth]{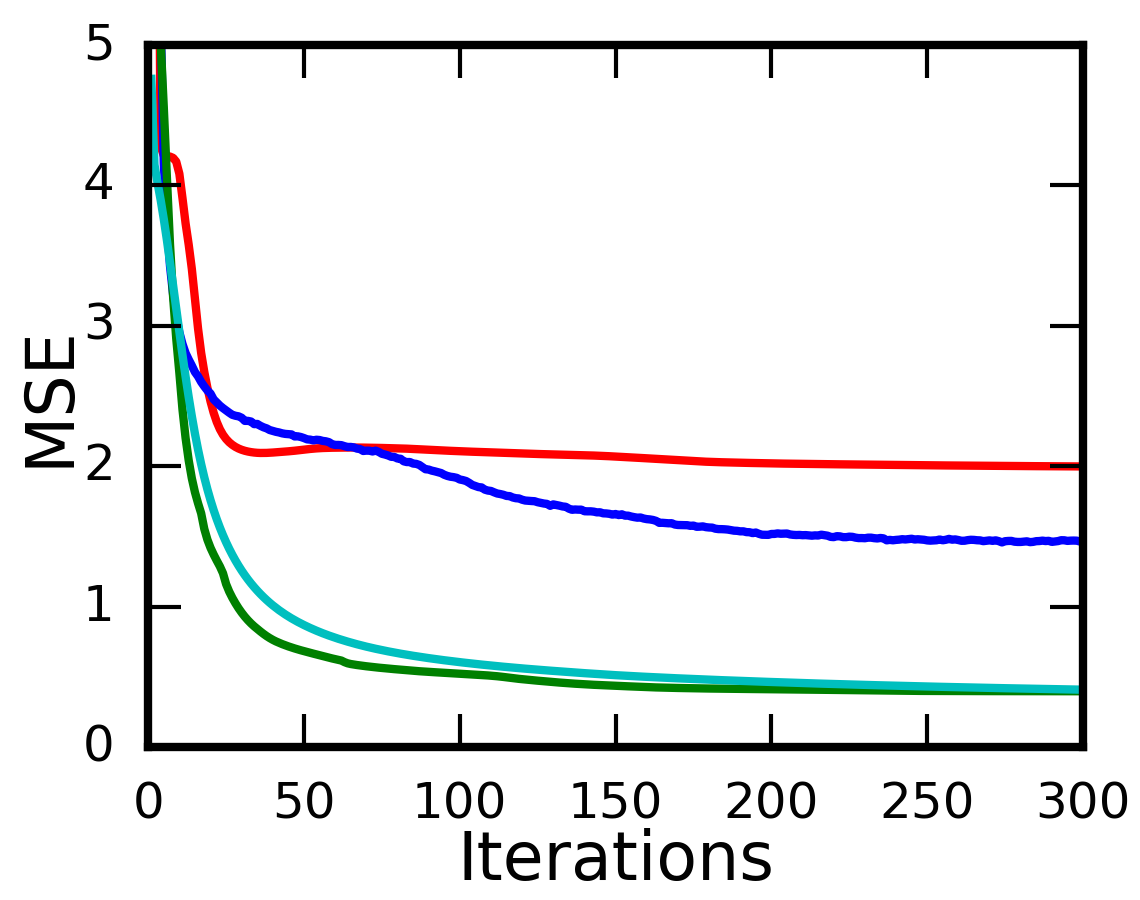}
		\end{subfigure} %
		\begin{subfigure}[t]{0.19 \columnwidth}
			\includegraphics[width=\columnwidth]{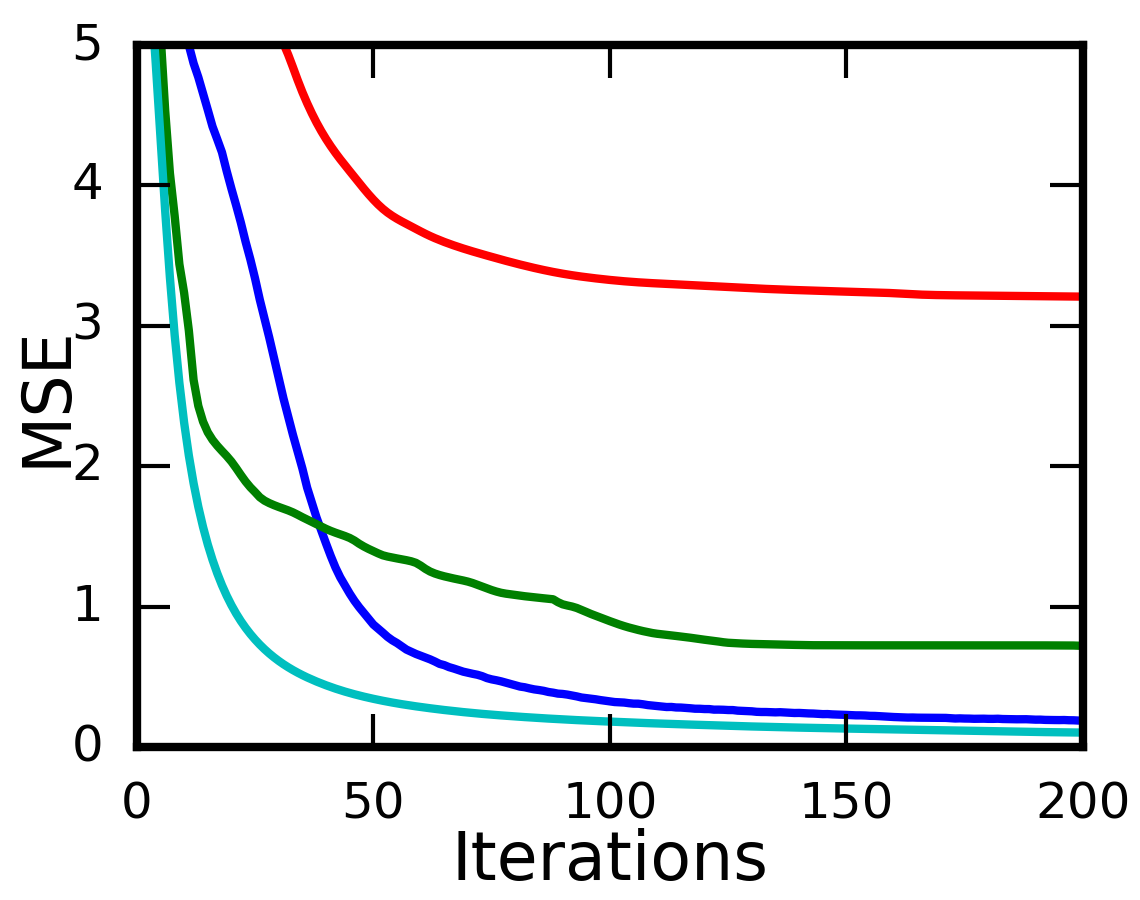}
		\end{subfigure} %
		\begin{subfigure}[t]{0.19 \columnwidth}
			\includegraphics[width=\columnwidth]{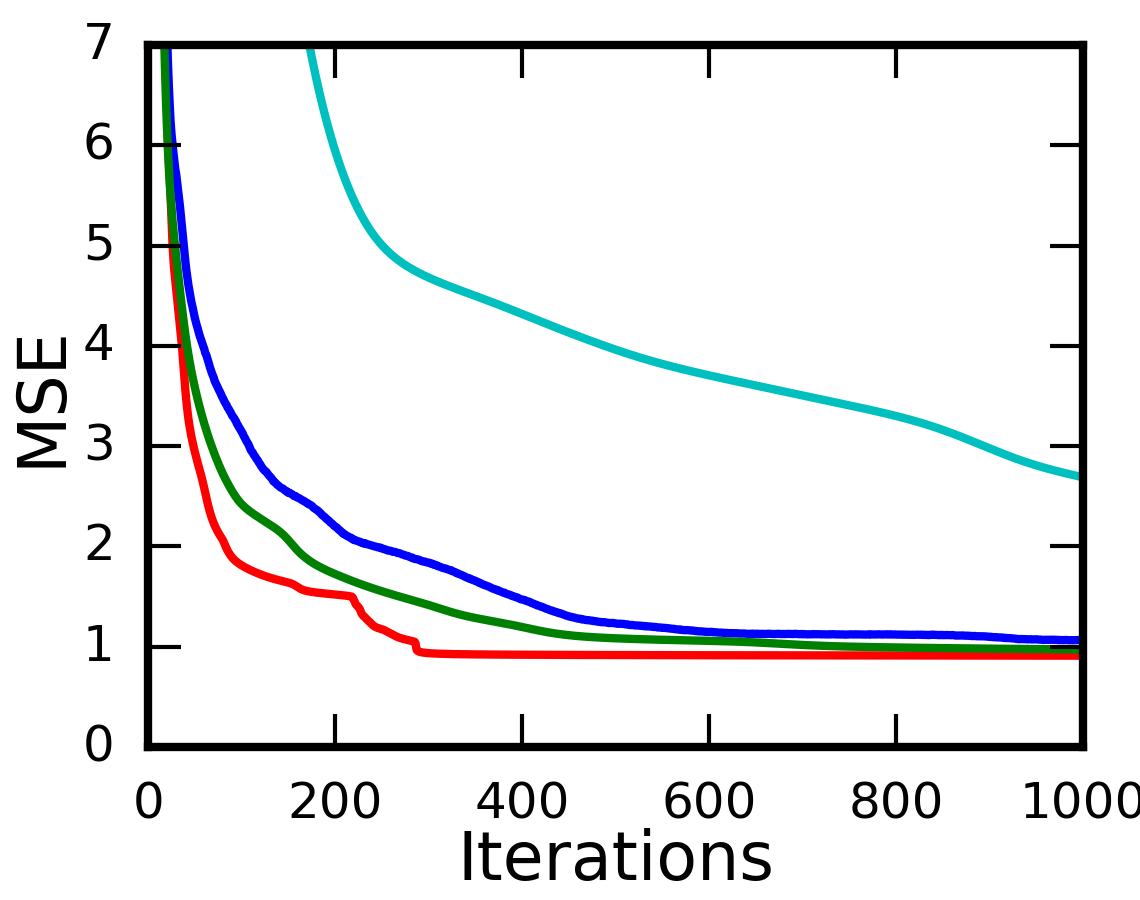}
			\captionsetup{width=\columnwidth}
			\caption{Synthetic} 
			\label{convergence_toy}
		\end{subfigure} %
		\begin{subfigure}[t]{0.19 \columnwidth}
			\includegraphics[width=\columnwidth]{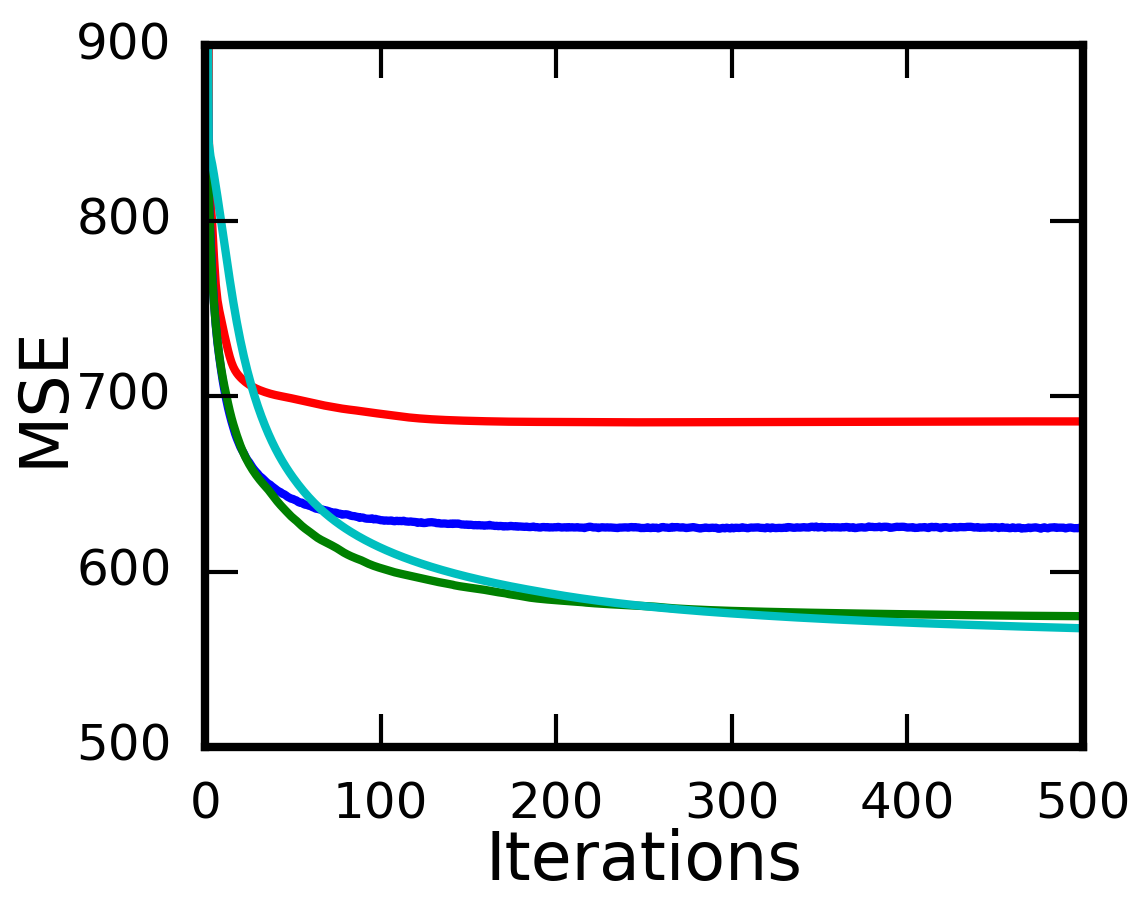}
			\captionsetup{width=\columnwidth}
			\caption{GDSC} 
			\label{convergence_gdsc}
		\end{subfigure} %
		\begin{subfigure}[t]{0.19 \columnwidth}
			\includegraphics[width=\columnwidth]{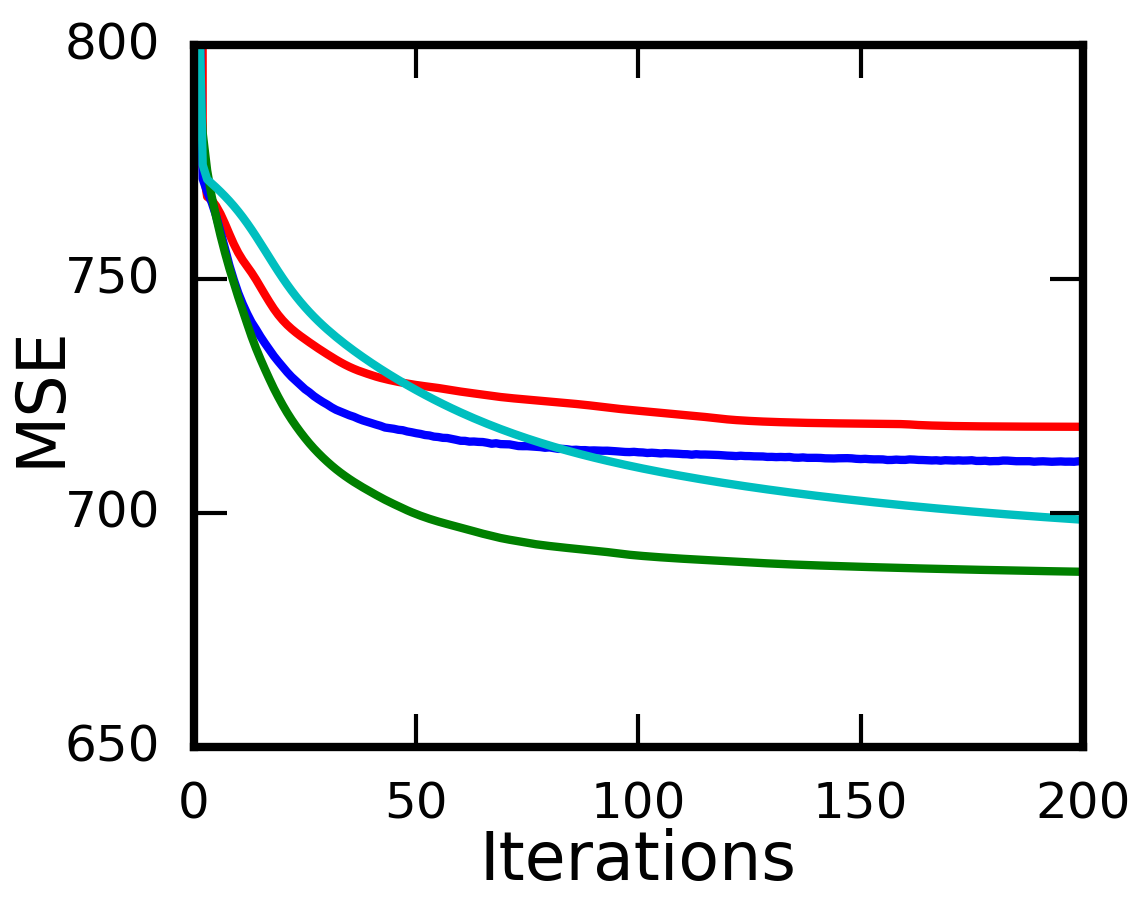}
			\captionsetup{width=\columnwidth}
			\caption{CTRP} 
			\label{convergence_ctrp}
		\end{subfigure} %
		\begin{subfigure}[t]{0.19 \columnwidth}
			\includegraphics[width=\columnwidth]{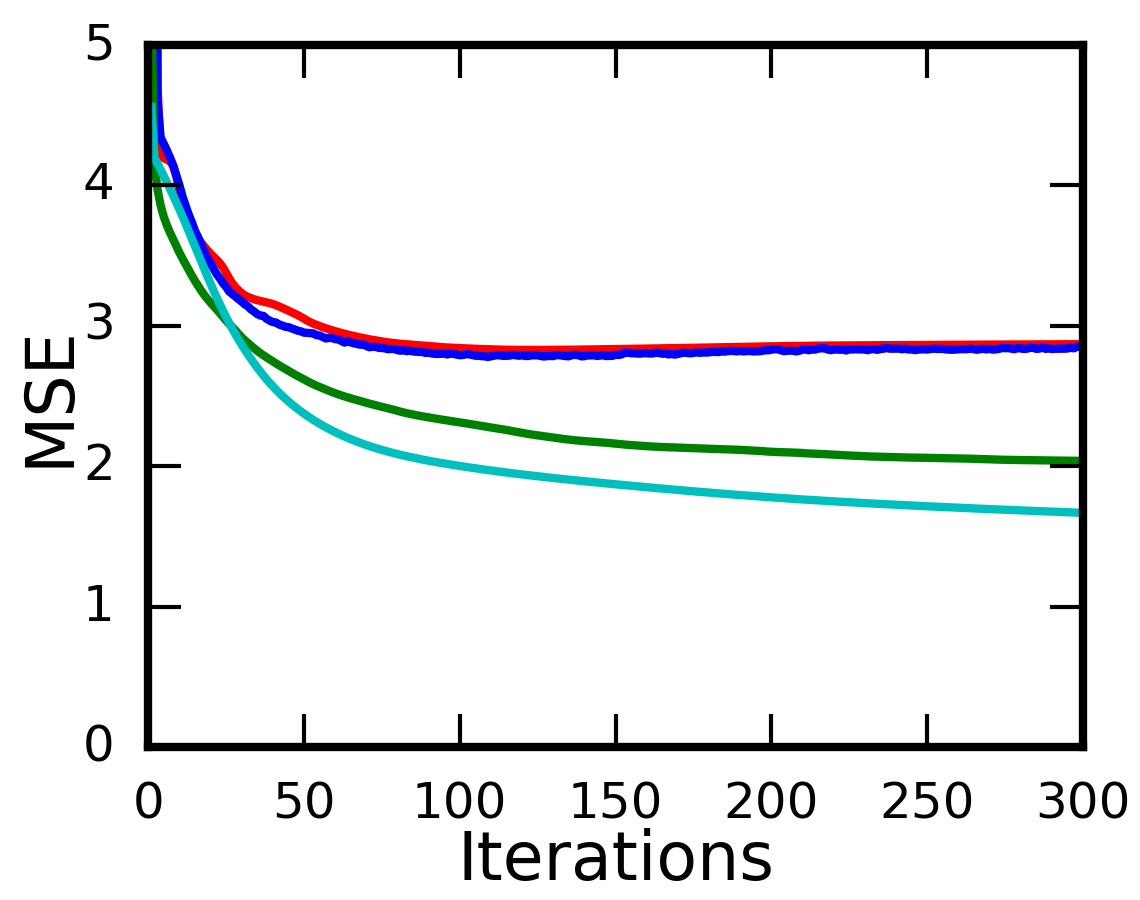}
			\captionsetup{width=\columnwidth}
			\caption{CCLE $IC_{50}$} 
			\label{convergence_ccle_ic}
		\end{subfigure} %
		\begin{subfigure}[t]{0.19 \columnwidth}
			\includegraphics[width=\columnwidth]{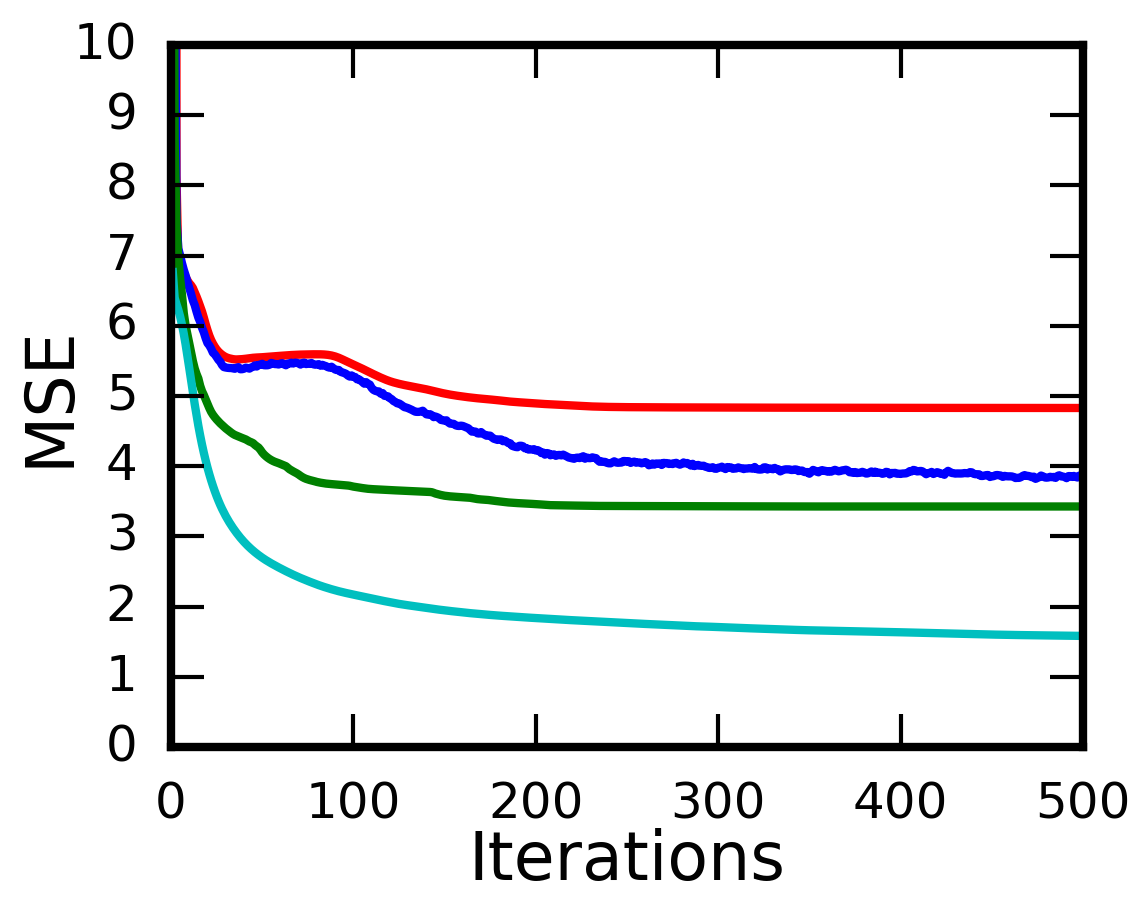}
			\captionsetup{width=\columnwidth}
			\caption{CCLE $EC_{50}$} 
			\label{convergence_ccle_ec}
		\end{subfigure} %
		\captionsetup{width=\columnwidth}
		\caption{Convergence of algorithms on the synthetic and drug sensitivity datasets, measuring the training data fit (mean square error) across iterations, for each of the inference approaches for NMF (top row) and NMTF (bottom row).}
		\label{convergence_results}
	\end{figure*}

	We firstly measured the convergence speeds of the different inference methods on the datasets, using the versions of NMF and NMTF without ARD. Convergence plots on all datasets are given in Figure \ref{convergence_results}, plotting the mean squared error on the training data against the number of iterations, for NMF (top row) and NMTF (bottom row). For the synthetic data we used the correct number of factors, and for the drug sensitivity datasets we used $K=20$ for NMF and $K=L=10$ for NMTF. We ran each method 20 times, taking the average training errors.
	
	Although the results are empirical, they show that the inference approaches have different convergence speeds and depths (final training error reached). On the synthetic data VB is the fastest, followed by ICM and Gibbs, and finally NP. All methods reach the optimal MSE of 1 (which is the level of noise added).
	On the real-world drug sensitivity datasets, all methods reach their lowest depth at roughly the same number of iterations. However, ICM and NP generally converge much deeper than VB and Gibbs. Although this initially seems good, this is a sign of overfitting to the training data, and can lead to poor predictions for unseen data. We will see this later in the noise and sparsity experiments (Sections \ref{section_noise} and \ref{section_sparsity}), where VB and Gibbs are more robust than ICM and NP.
	
	In the supplementary materials we also give the convergence speed against time taken, which shows that the NP approach takes the least amount of time per iteration, followed by ICM, VB, and then Gibbs. 
	In summary, ICM and NP give the fastest convergence, followed by VB, and then Gibbs.
	
\subsection{Cross-Validation} \label{cross_validation}
\begin{figure}[t]
		\centering
		\includegraphics[width=0.95\columnwidth]{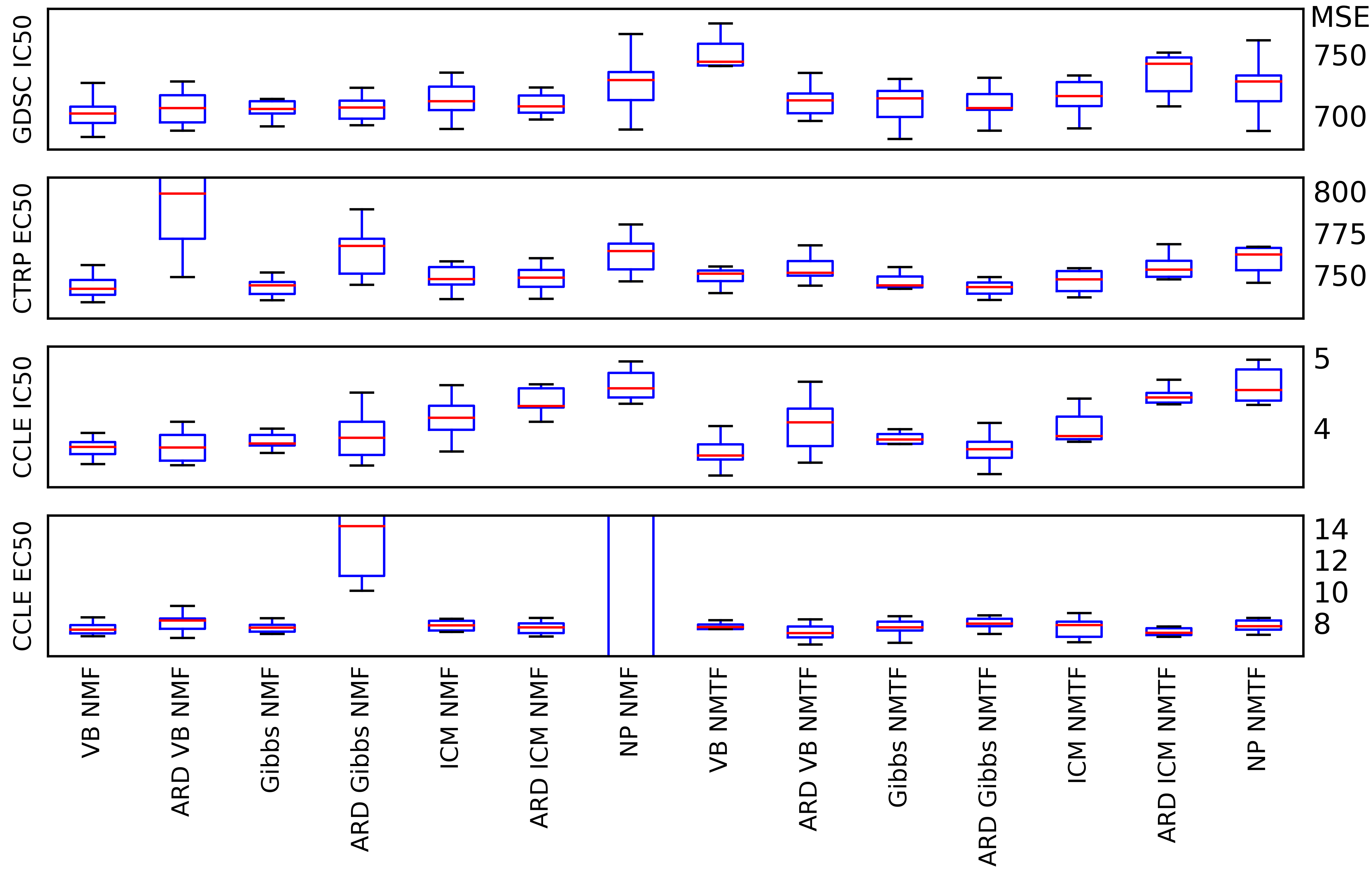}
		\captionsetup{width=1\columnwidth}
		\caption{10-fold cross-validation results (mean squared error) for drug sensitivity predictions on each of the four datasets. Each boxplot gives the median (red line), standard deviation (blue box), and upper quartiles (black lines).}
		\label{cross_validation_results}
	\end{figure}

	Next we measured the cross-validation performances of the methods on the four drug sensitivity datasets. For each method we performed 10-fold nested cross-validation (nested to pick the dimensionality $K$---for simplicity we used $L=K$ for the NMTF models), giving the average performance in Figure \ref{cross_validation_results}. For the ARD models we did not need to pick the dimensionality, instead using $K=20$ for NMF, and $K=10, L=10$ for NMTF. 
	
	We can see that most models perform very similarly, with little to no difference between the matrix factorisation and tri-factorisation versions. Using the ARD models often works equally well as without ARD, but with the added benefit of not having to run nested cross-validation to choose the dimensionality, reducing the running time from hours to minutes. 
	However, sometimes ARD fails to prevent overfitting, such as for VB NMF on CTRP $EC_{50}$, and Gibbs NMF on CCLE $EC_{50}$). This is unsurprising as the ARD models are given dimensionalities that are way too high.
	We will see in Section \ref{section_model_selection} that the ARD is actually very efficient at turning off unnecessary factors and reducing overfitting.
	
	We can also see that the VB and Gibbs models often do a bit better than the NP and ICM versions. This is especially obvious on the CCLE $IC_{50}$ dataset, and also on GDSC $IC_{50}$. On the CCLE $EC_{50}$ dataset the NP NMF model completely overfits on one of the folds, leading to extremely high predictive errors.
	
\subsection{Noise Test} \label{section_noise}
\begin{figure*}
		\centering
		\begin{subfigure}[t]{\columnwidth}
			\hspace{35pt}
			\includegraphics[width=0.80\columnwidth]{legend.png}
		\end{subfigure}
		\begin{subfigure}{0.4 \columnwidth}
			\includegraphics[width=\columnwidth]{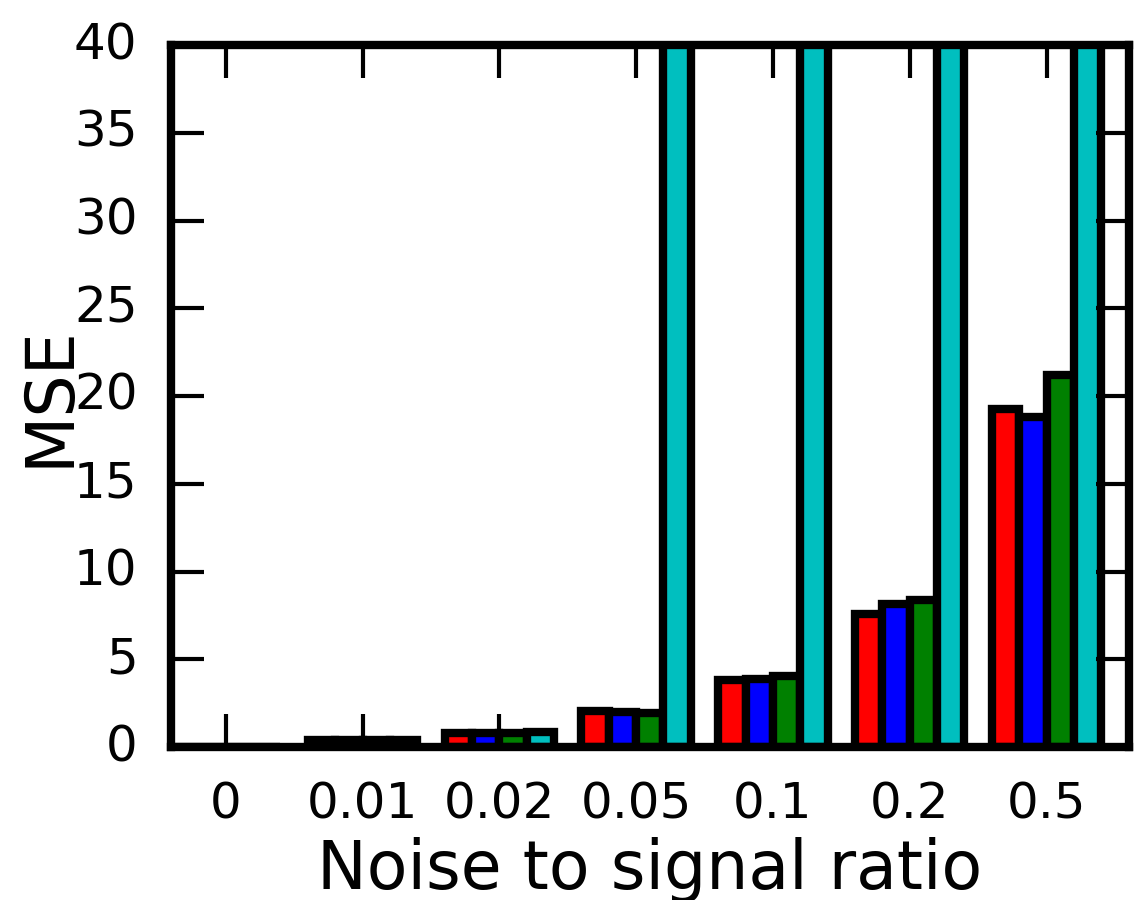}
			\captionsetup{width=0.9\columnwidth}
			\caption{NMF} 
			\label{mse_nmf_noise_test}
		\end{subfigure} %
		\begin{subfigure}{0.4 \columnwidth}
			\includegraphics[width=\columnwidth]{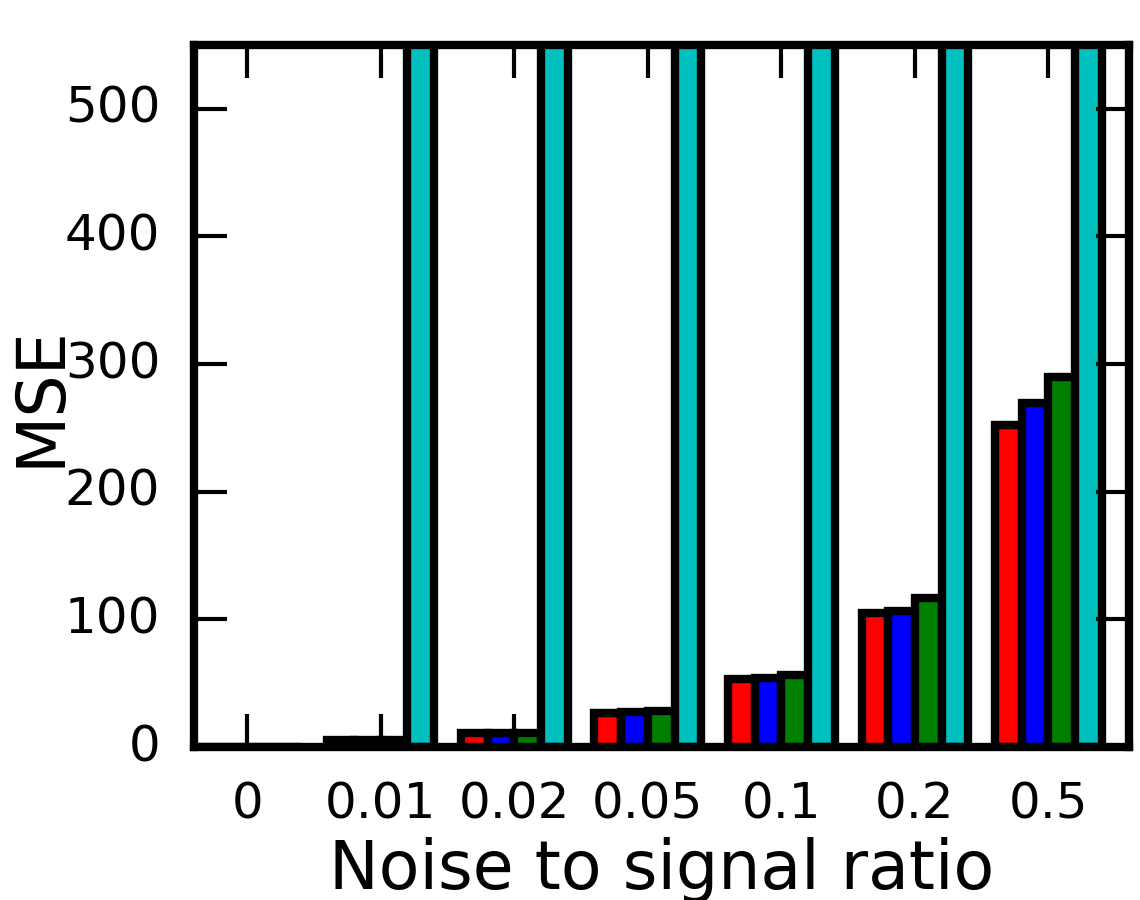}
			\captionsetup{width=0.9\columnwidth}
			\caption{NMTF} 
			\label{mse_nmtf_noise_test}
		\end{subfigure}
		\captionsetup{width=1\columnwidth}
		\caption{Noise test performances, measured by average predictive performance on test set (mean square error) for different noise-to-signal ratios.}
		\vspace{20pt}
		\begin{subfigure}[t]{\columnwidth}
			\hspace{35pt}
			\includegraphics[width=0.80\columnwidth]{legend.png}
		\end{subfigure}
		\begin{subfigure}{0.32 \columnwidth}
			\includegraphics[width=\columnwidth]{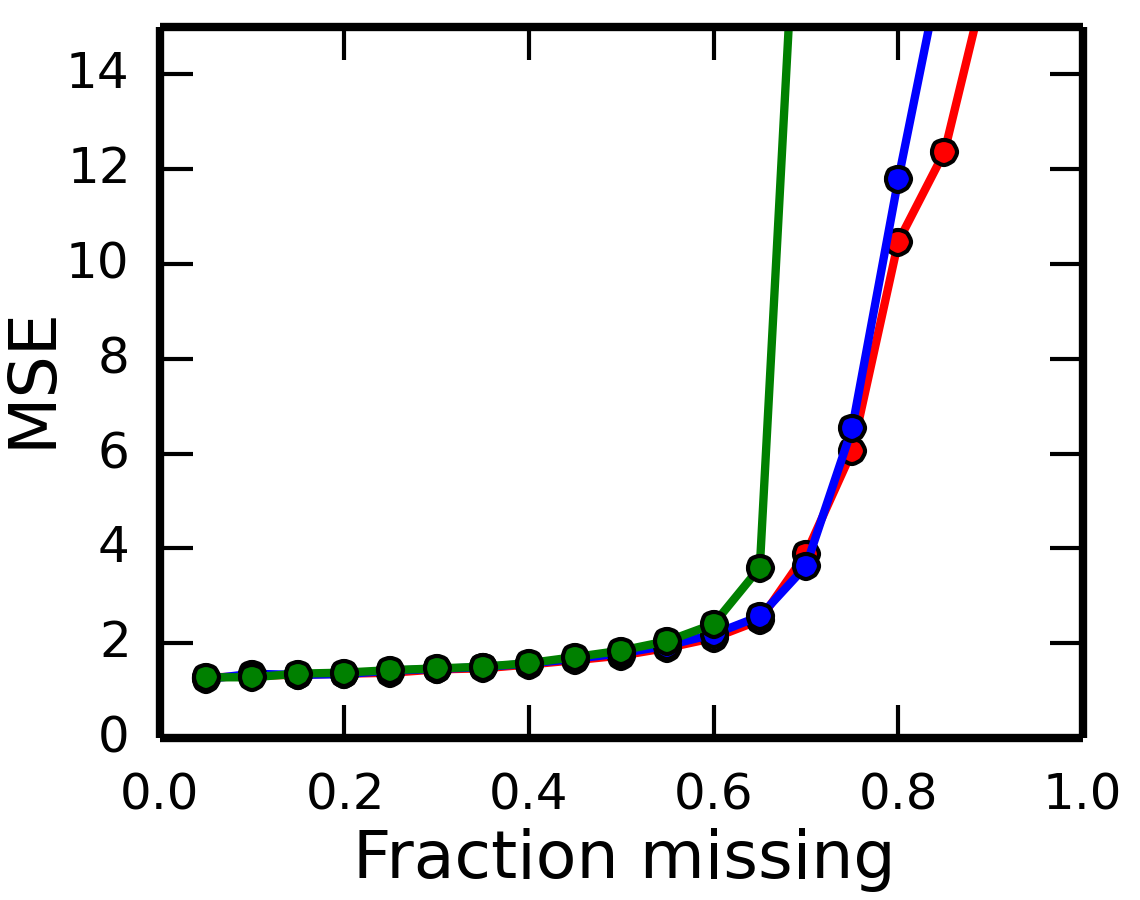}
			\captionsetup{width=0.9\columnwidth}
			\caption{Synthetic, NMF} 
			\label{mse_nmf_sparsity_test_toy}
		\end{subfigure} %
		\begin{subfigure}{0.32 \columnwidth}
			\includegraphics[width=\columnwidth]{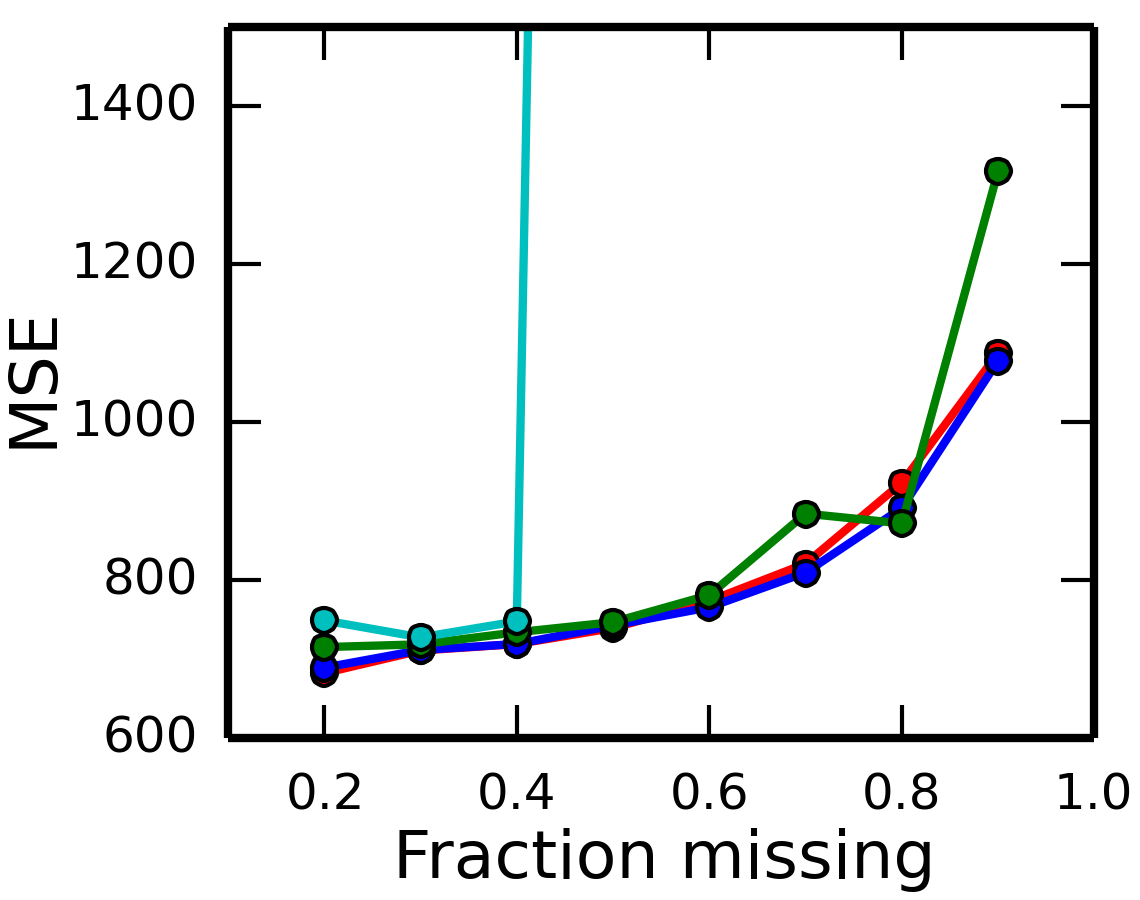}
			\captionsetup{width=0.9\columnwidth}
			\caption{GDSC, NMF} 
			\label{mse_nmf_sparsity_test_gdsc}
		\end{subfigure} %
		\begin{subfigure}{0.32 \columnwidth}
			\includegraphics[width=\columnwidth]{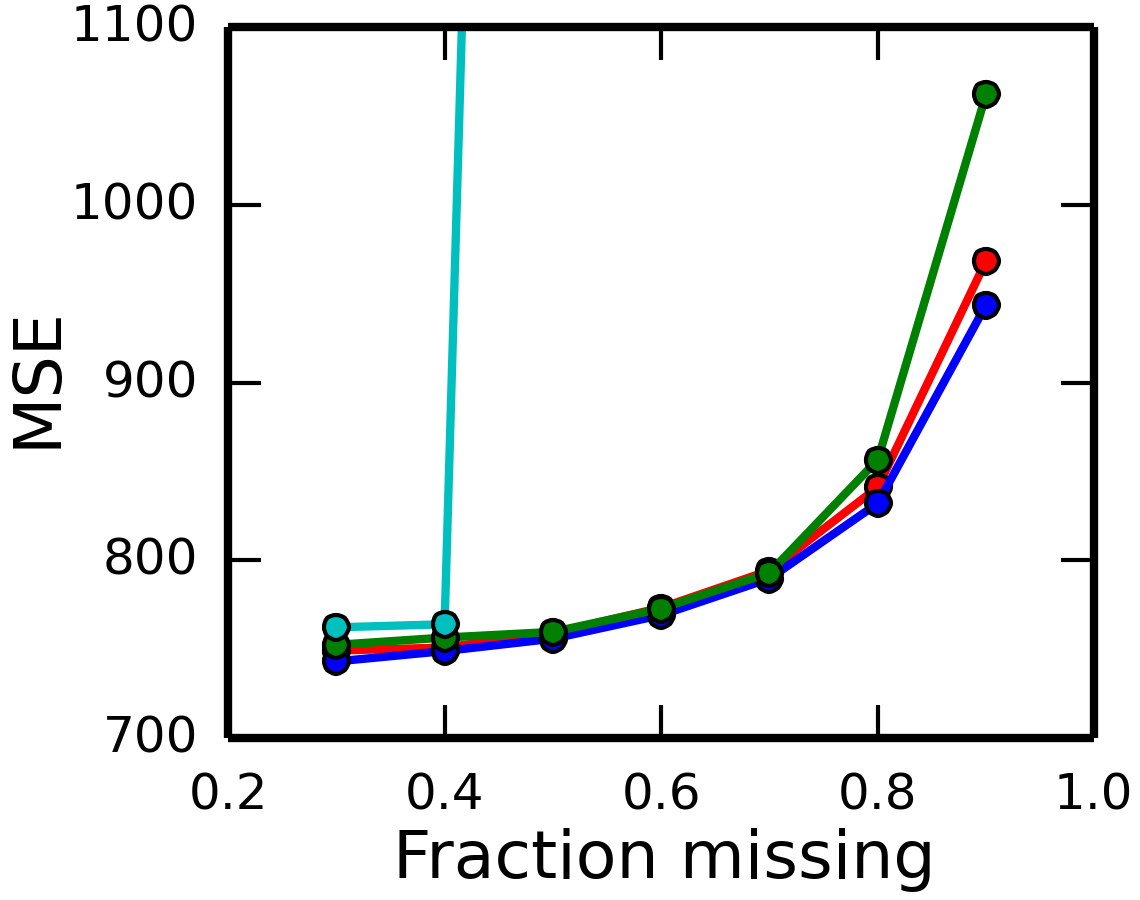}
			\captionsetup{width=0.9\columnwidth}
			\caption{CTRP, NMF} 
			\label{mse_nmf_sparsity_test_ctrp}
			\vspace{5pt}
		\end{subfigure} %
		\begin{subfigure}{0.32 \columnwidth}
			\includegraphics[width=\columnwidth]{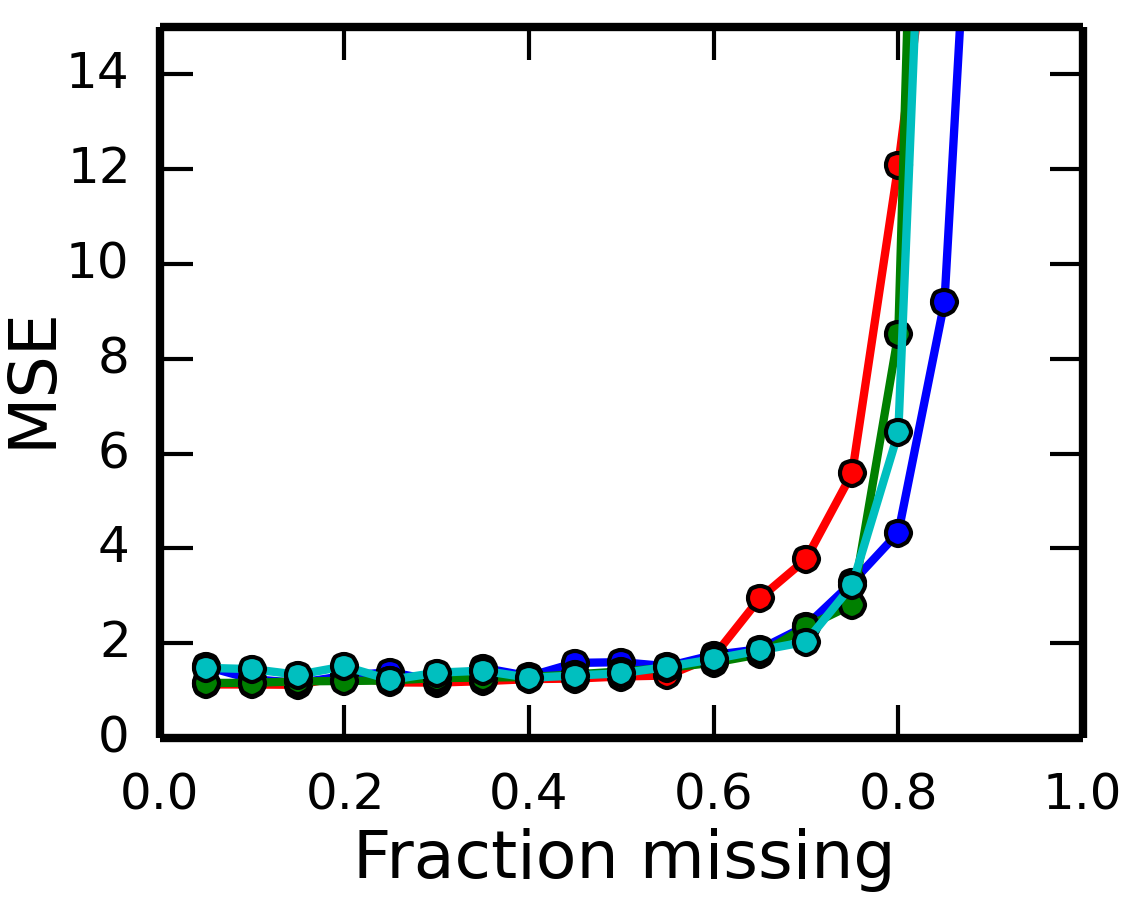}
			\captionsetup{width=0.9\columnwidth}
			\caption{Synthetic, NMTF} 
			\label{mse_nmtf_sparsity_test_toy}
		\end{subfigure}
		\begin{subfigure}{0.32 \columnwidth}
			\includegraphics[width=\columnwidth]{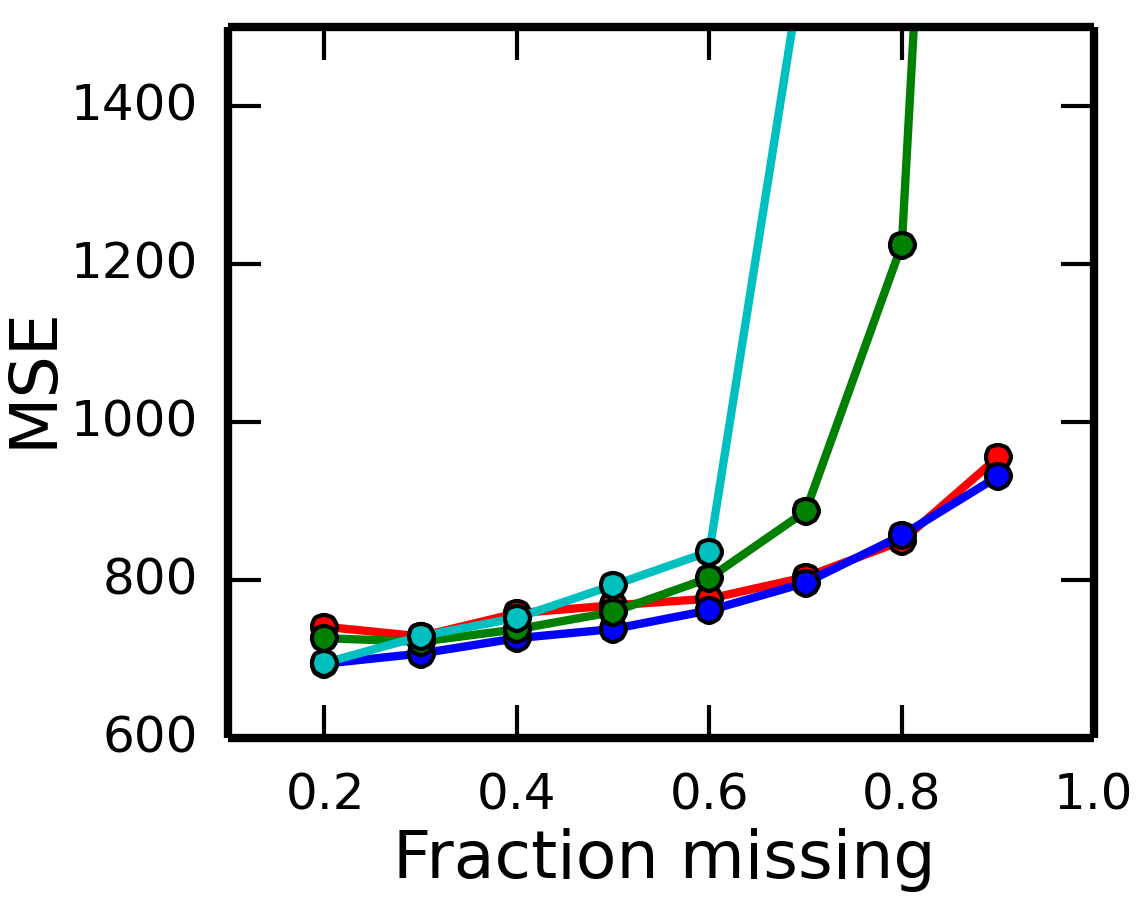}
			\captionsetup{width=0.9\columnwidth}
			\caption{GDSC, NMTF} 
			\label{mse_nmtf_sparsity_test_gdsc}
		\end{subfigure}
		\begin{subfigure}{0.32 \columnwidth}
			\includegraphics[width=\columnwidth]{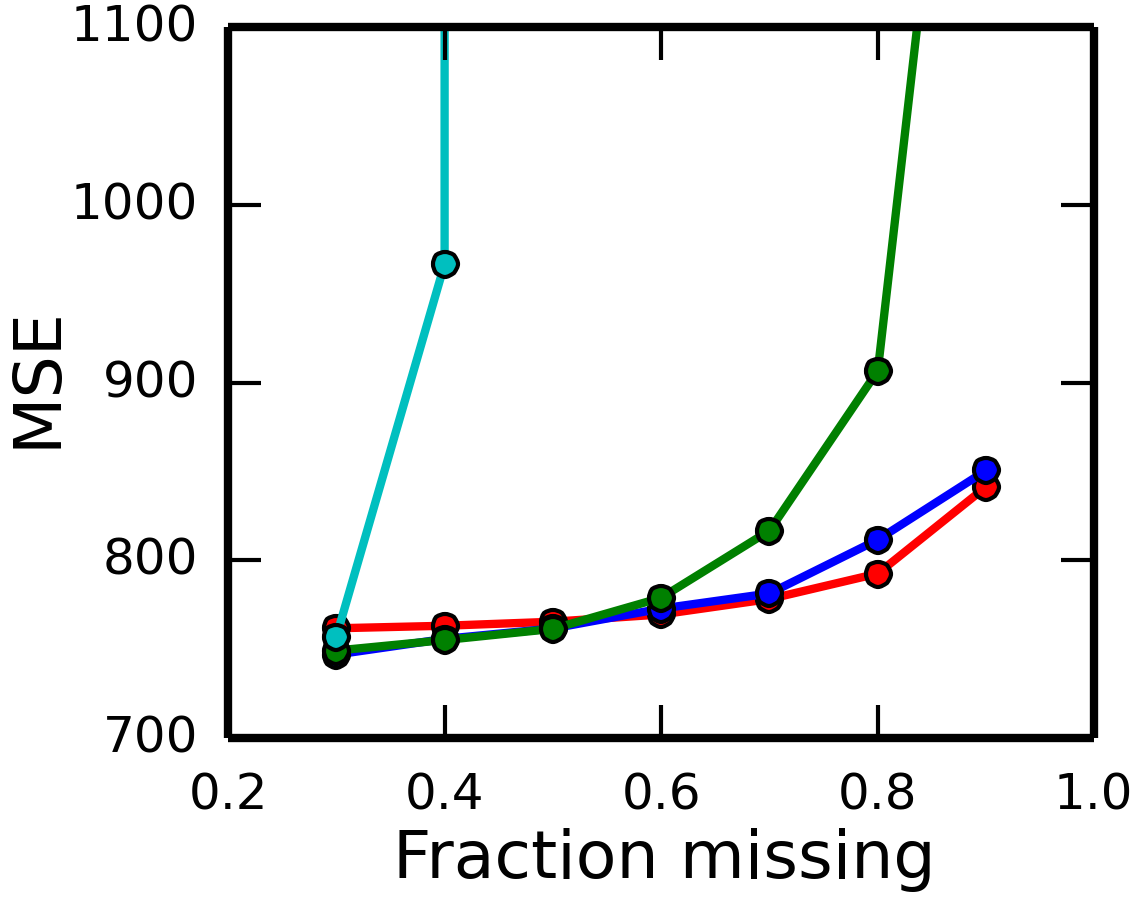}
			\captionsetup{width=0.9\columnwidth}
			\caption{CTRP, NMTF} 
			\label{mse_nmtf_sparsity_test_ctrp}
		\end{subfigure}
		\captionsetup{width=1\columnwidth}
		\caption{Sparsity test performances, measured by average predictive performance on test set (mean square error) for different sparsity levels. The top row gives the performances for NMF, and the bottom for NMTF, for the synthetic data (left), GDSC dataset (middle), and CTRP dataset (right).}
	\end{figure*}

	We conducted a noise test on the synthetic data to measure the robustness of the methods. We add different levels of Gaussian noise to the data, with the noise-to-signal ratio being given by the ratio of the standard deviation of the Gaussian noise we add, to the standard deviation of the generated data. For each noise level we split the datapoints randomly into ten folds, and measure the predictive performance of the models on one held-out set. 
	The results are given in Figures \ref{mse_nmf_noise_test} (NMF) and \ref{mse_nmtf_noise_test} (NMTF), where we can see that the non-probabilistic approach starts overfitting heavily at low levels of noise, whereas the Bayesian approaches achieve the best possible predictive powers even at high levels of noise. %ICM only does slightly worse in the very noisy cases.
	In the supplementary materials we also show that adding ARD did not make a difference for the robustness of the Bayesian models.
	
\subsection{Sparsity Test} \label{section_sparsity}
	We furthermore measured the robustness of each inference technique to sparsity of the data. For different fractions of missing values we randomly split the data ten times into train and test sets using those proportions, and measured the average predictive error.
	We conducted this experiment on the synthetic data, using the true dimensionality $K$ (and $L$) for each model. We also performed it on the GDSC and CTRP datasets, using the most common dimensionalities in the cross-validation from Section \ref{cross_validation} (given in supplementary materials).
	
	The results are given in Figures \ref{mse_nmf_sparsity_test_toy} (NMF) and \ref{mse_nmtf_sparsity_test_toy} (NMTF) for the synthetic data, \ref{mse_nmf_sparsity_test_gdsc} and \ref{mse_nmtf_sparsity_test_gdsc} for GDSC, and \ref{mse_nmf_sparsity_test_ctrp} and \ref{mse_nmtf_sparsity_test_ctrp} for CTRP.
	We can see that the non-probabilistic models start overfitting even on very low sparsity levels (with the exception of \ref{mse_nmtf_sparsity_test_toy})---in Figure \ref{mse_nmf_sparsity_test_toy} we cannot even see the line. The ICM models are also less robust when the sparsity is high. In contrast, the Gibbs sampling model achieves very good predictive performance even under extreme sparsity. The VB models are similar, but for sparser data it can sometimes not find the best solution, as can be seen in Figure \ref{mse_nmtf_sparsity_test_toy}.
	We conducted this experiment for the models with ARD as well (results given in supplementary materials), where we show that ARD makes no difference to the robustness of Gibbs and VB (which are already very robust), but for ICM it can sometimes improve results.

\subsection{Model Selection} \label{section_model_selection}
	Finally, we conducted an experiment to see the extent of overfitting if the model is given a high dimensionality $K$, and whether this is remedied through the use of ARD. If we give a model a higher dimensionality, it can fit more to the data, but this can lead to overfitting and a higher predictive error. ARD can remedy this by turning off scarsely used factors, hopefully leading to less overfitting.
	
\begin{figure*}[t]
		\centering
		\begin{subfigure}{0.3 \columnwidth}
			\includegraphics[width=\columnwidth]{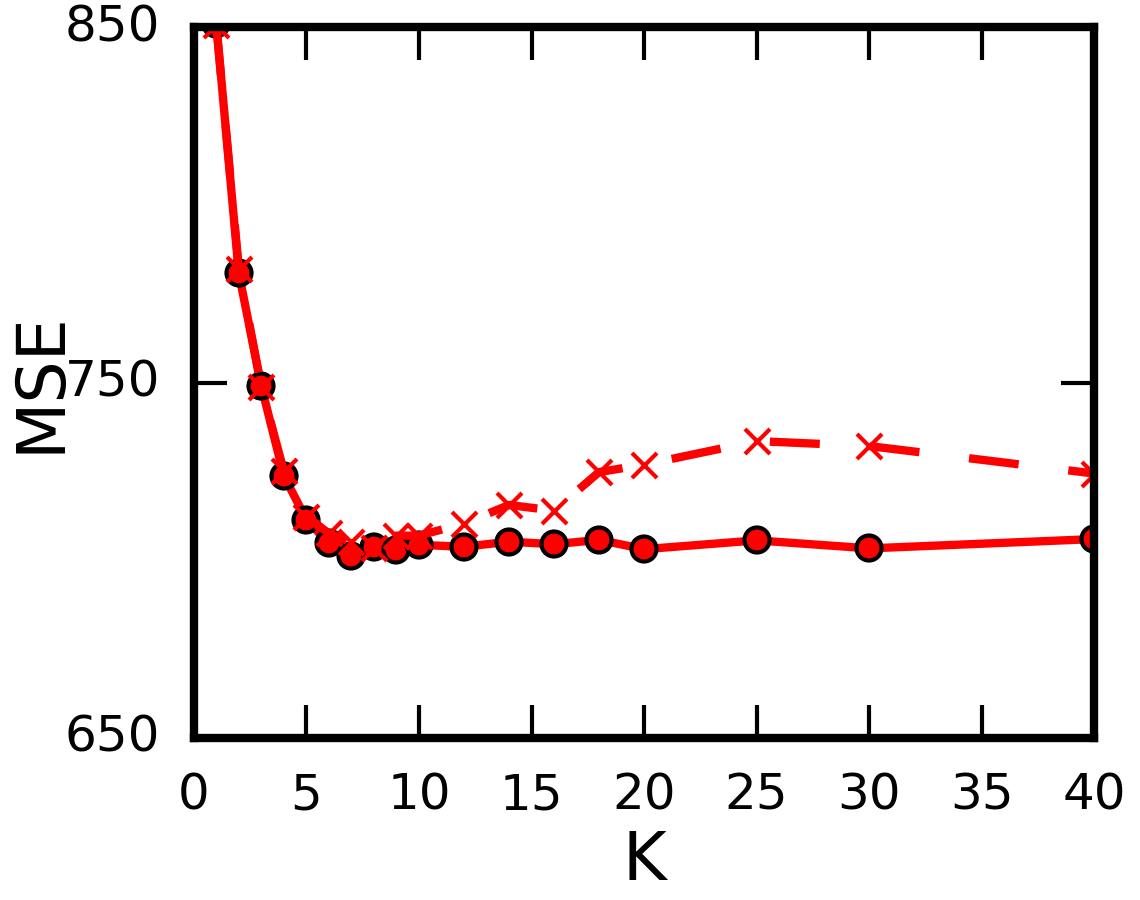}
			\captionsetup{width=0.9\columnwidth}
			\caption{NMF VB} 
			\label{nmf_vb_model_selection}
		\end{subfigure}
		\begin{subfigure}{0.3 \columnwidth}
			\includegraphics[width=\columnwidth]{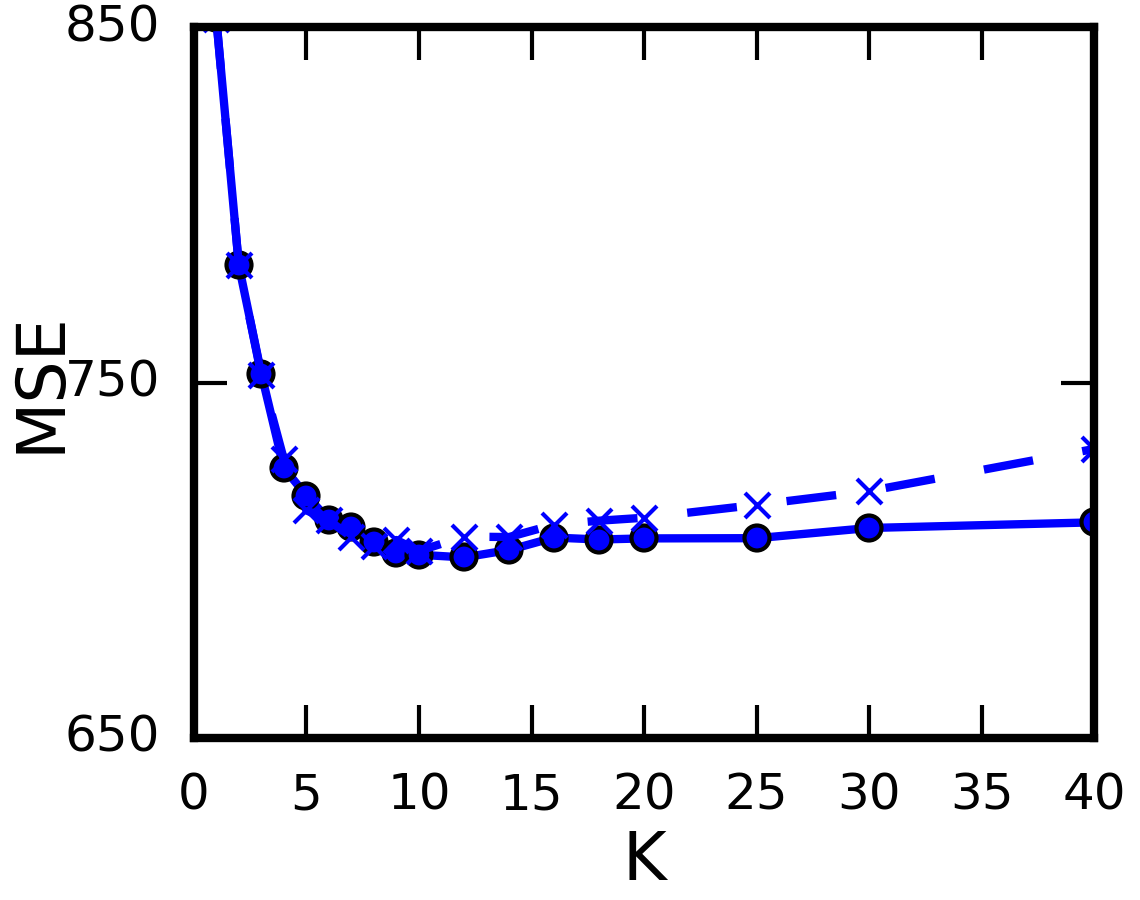}
			\captionsetup{width=0.9\columnwidth}
			\caption{NMF Gibbs} 
			\label{nmf_gibbs_model_selection}
		\end{subfigure}
		\begin{subfigure}{0.3 \columnwidth}
			\includegraphics[width=\columnwidth]{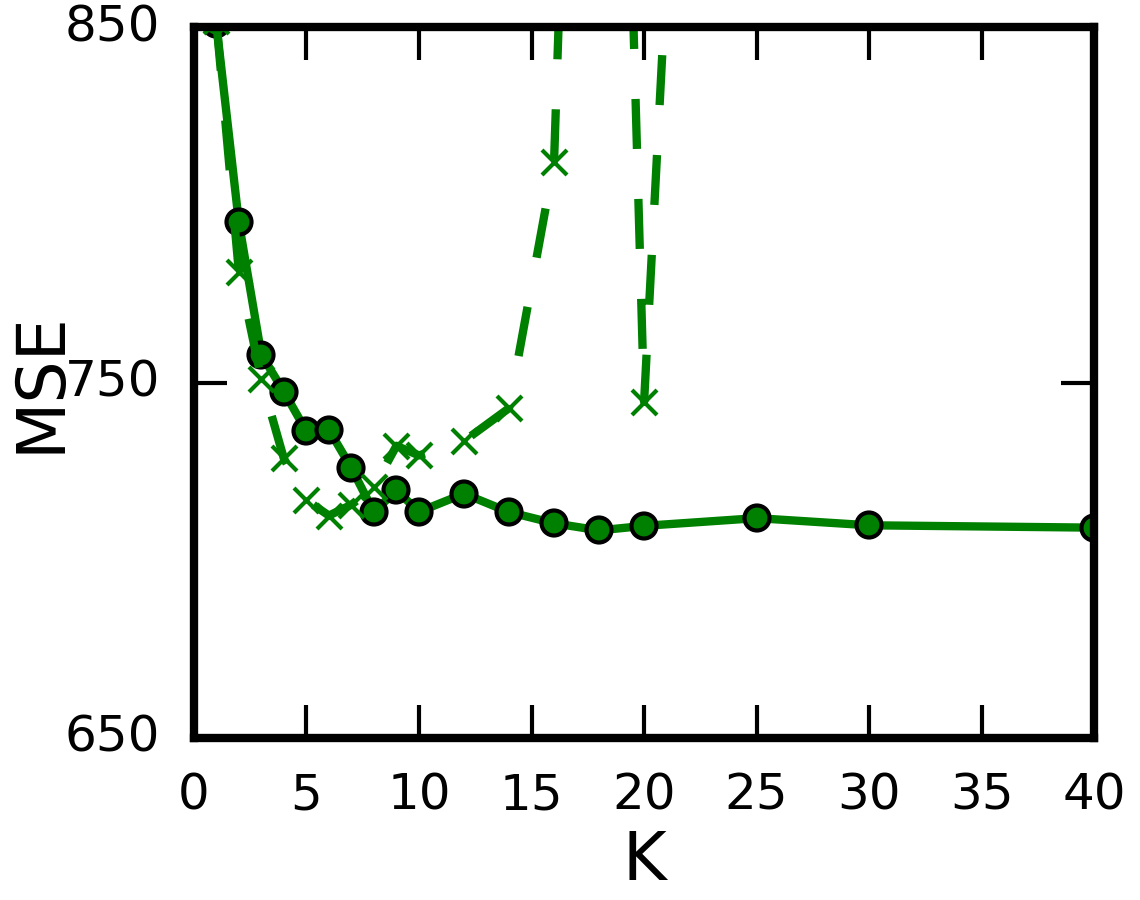}
			\captionsetup{width=0.9\columnwidth}
			\caption{NMF ICM} 
			\label{nmf_icm_model_selection}
		\end{subfigure} 
		\begin{subfigure}{0.3 \columnwidth}
			\includegraphics[width=\columnwidth]{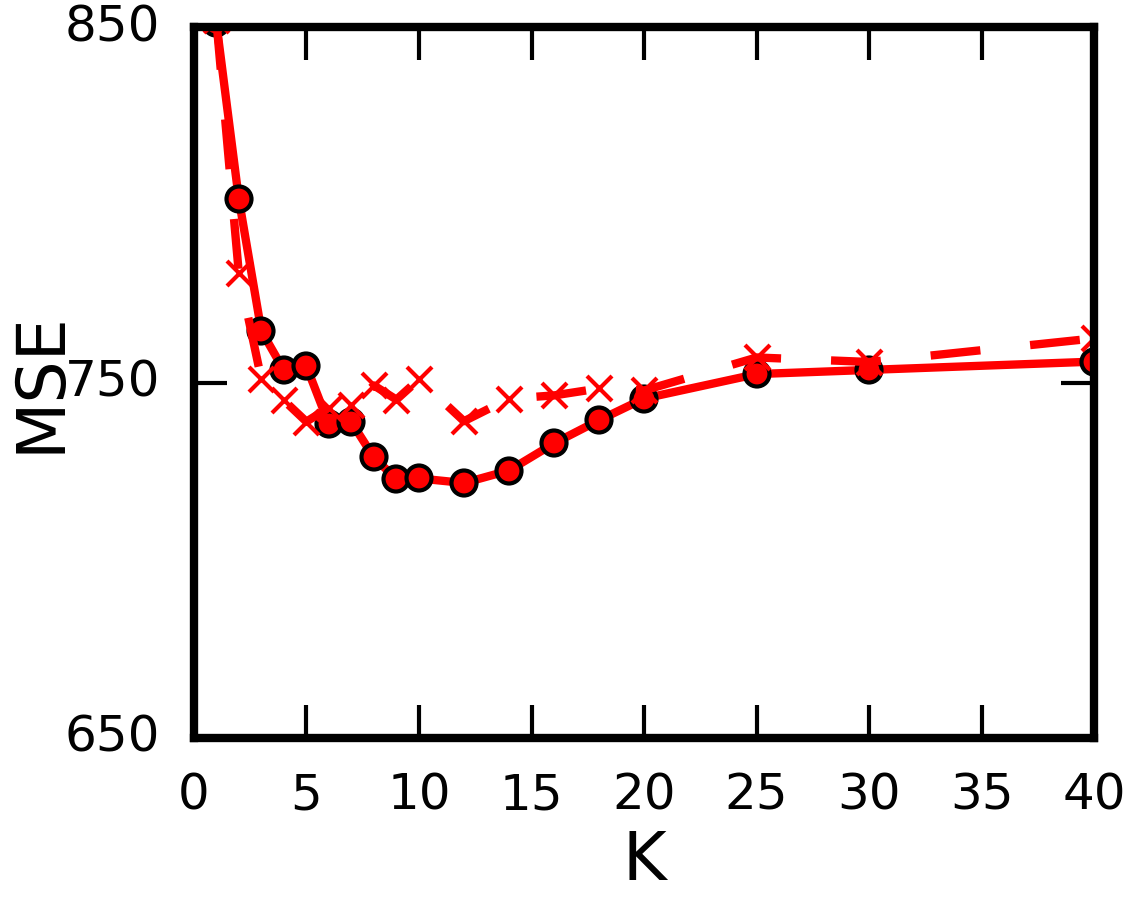}
			\captionsetup{width=\columnwidth}
			\caption{NMTF VB} 
			\label{nmtf_vb_model_selection}
		\end{subfigure}
		\begin{subfigure}{0.3 \columnwidth}
			\includegraphics[width=\columnwidth]{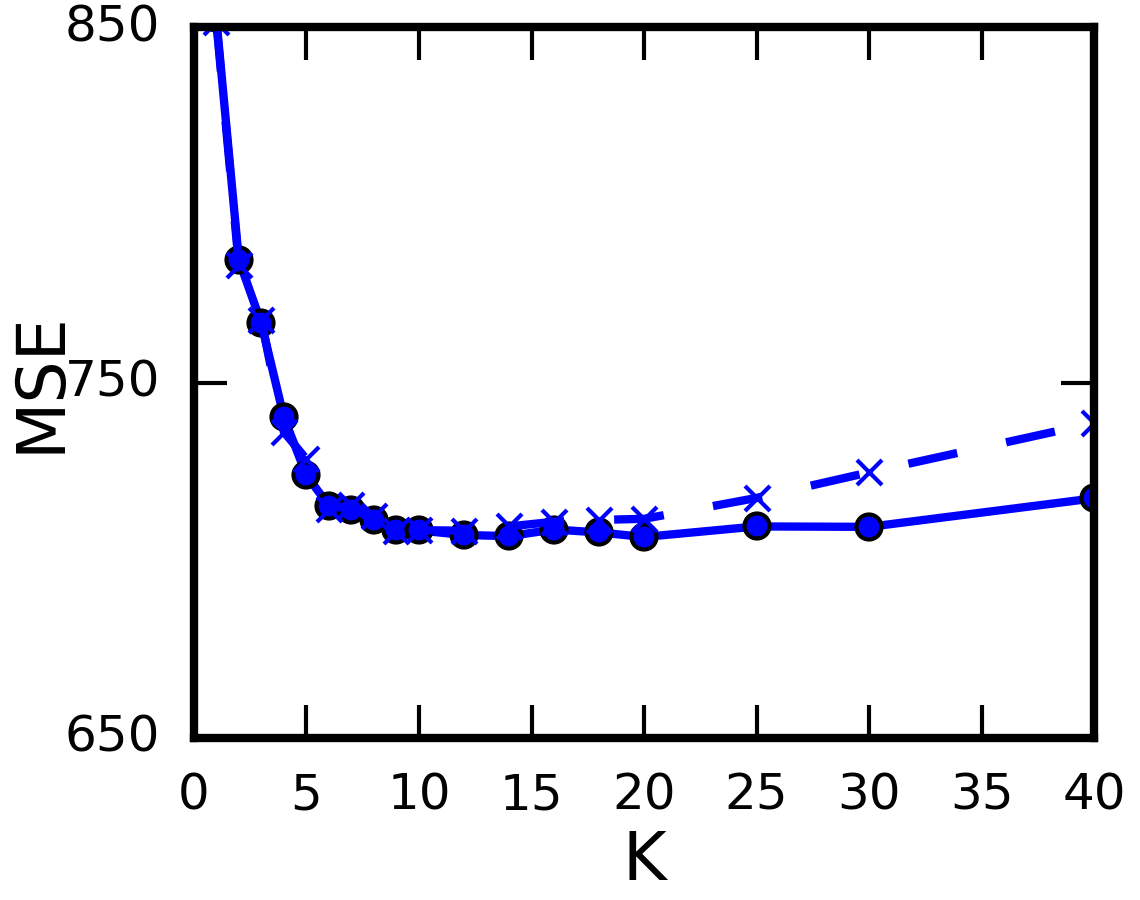}
			\captionsetup{width=\columnwidth}
			\caption{NMTF Gibbs}
			\label{nmtf_gibbs_model_selection}
		\end{subfigure}
		\begin{subfigure}{0.3 \columnwidth}
			\includegraphics[width=\columnwidth]{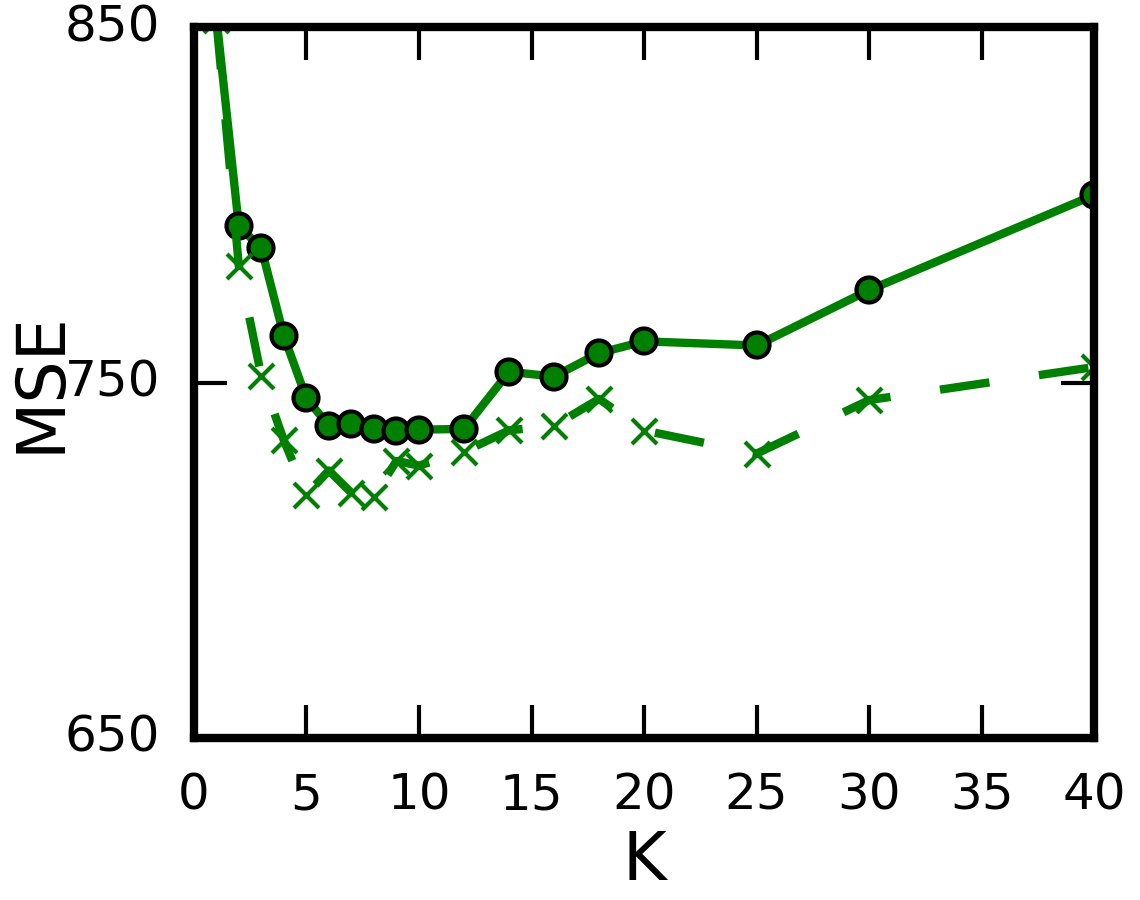}
			\captionsetup{width=\columnwidth}
			\caption{NMTF ICM} 
			\label{nmtf_icm_model_selection}
		\end{subfigure} 
		\captionsetup{width=1\columnwidth}
		\caption{10-fold cross-validation performances of the Bayesian models on the GDSC dataset, where we vary the dimensionality $K$ (using $L=K$ for NMTF). The top row gives the performances for NMF, the bottom row for NMTF. Performances for models without ARD are given by dotted lines and crosses (x), with ARD by circles (o).}
	\end{figure*}

	On the GDSC dataset, we performed 10-fold cross-validation for different values of $K$ (and $L$ for NMTF, using $K=L$) for Gibbs, VB, and ICM. We show these results in Figures \ref{nmf_vb_model_selection} to \ref{nmtf_icm_model_selection}, where the results for models without ARD are given by crosses (x) and with ARD by circles (o). We can see that in most graphs, the models with ARD have a much flatter line as the dimensionality increases, hence reducing overfitting. This effect is more apparent for the NMF models than for the NTMF ones. The only exception is NMTF ICM, where the ARD is preventing the model from fitting as much to the data, hence leading to poor predictive results. Results for this experiment on the other three drug sensitivity datasets is given in the supplementary materials, which show that this problem only occurred for NMTF ICM on the GDSC dataset.
	
% % % % % % % % % % % % % % % % % % % % % % % % % % % % % % % % % % % % % % % % % % % % % % % % %

\section{Conclusion}
We have studied the trade-offs between different inference approaches for Bayesian nonnegative matrix factorisation and tri-factorisation models. We considered three methods, namely Gibbs sampling, iterated conditional modes, and non-probabilistic inference, and introduced a fourth one based on variational Bayesian inference. We furthermore extended these models with the Bayesian automatic relevance determination prior, to perform automatic model selection.
Through experiments on both synthetic data, and real-world drug sensitivity datasets, we explored the trade-offs in convergence, robustness to noise, and robustness to sparsity.

A qualitative summary based on our quantitative findings can be found in Table \ref{comparison_methods}. We found that the non-probabilistic methods are not very robust to noise and sparsity. Gibbs sampling is the most robust of the methods, especially for sparse datasets, and gives a full Bayesian posterior estimate. However, it converges slowly, and requires additional samples to estimate the posterior. Iterated conditional modes offers a much faster convergence and run-time speed, but sacrifices some robustness, still requires sampling, and no longer returns a full posterior (giving a MAP estimate instead). Our variational Bayesian inference gives good convergence speeds while maintaining more robustness properties.

Finally, we have shown that ARD is an effective way of reducing overfitting when using the wrong dimensionality in matrix factorisation models. This can eliminate the use for performing model selection, or nested cross-validation---although it is not perfect. We also discovered that adding ARD has little impact on performance, or on the robustness of the models to sparsity and noise (except for iterated conditional modes, where ARD increases its robustness to sparsity).

Our experiments were conducted for a specific version of Bayesian matrix factorisation and tri-factorisation, but we believe they offer insights into the trade-offs between different inference techniques in other matrix factorisation models, as well as tensor and Tucker decomposition methods.

\begin{table*}[t]
	\caption{Qualitative comparison of inference methods.} \label{comparison_methods}
	\begin{tabular}{lllll}
		\toprule
		& &{\bf Requires} & {\bf Speed of} & \\
		{\bf Method} \hspace{72pt} &{\bf Estimate} \hspace{17pt} &{\bf sampling} \hspace{2pt} &{\bf convergence} \hspace{2pt} &{\bf Robustness} \\
		\midrule
		Non-probabilistic & Point & No & High & Low \\
		Iterated conditional modes & Point (MAP) & Yes & High & Medium \\
		Gibbs sampling & Full posterior & Yes & Low & High \\
		Variational Bayes & Full posterior & No & Medium & Fairly high \\
		\bottomrule
	\end{tabular}
\end{table*}

% % % % % % % % % % % % % % % % % % % % % % % % % % % % % % % % % % % % % % % % % % % % % % % % %

\subsubsection{Acknowledgements.} 
This work was supported by the UK Engineering and Physical Sciences Research Council (EPSRC), grant reference EP/M506485/1. JF acknowledge funding from the Danish Council for Independent Research 0602-02909B.

% % % % % % % % % % % % % % % % % % % % % % % % % % % % % % % % % % % % % % % % % % % % % % % % % % % % % % %

\bibliography{bibliography}
\bibliographystyle{splncs03}

\end{document}

\title{Comparative Study of Inference Methods for \\ Bayesian Nonnegative Matrix Factorisation \\ \vspace{10pt} Supplementary Materials}
	\author{Thomas Brouwer, Jes Frellsen, Pietro Li\'{o}}
	\date{}
	\maketitle{}
	
	\noindent \large{European Conference on Machine Learning and Principles and Practice of Knowledge Discovery in Databases (ECML PKDD 2017).} \\
	
	\tableofcontents
	\clearpage
		
%%%%%%%%%%%%%%%%%%%%%%%%%%%%%%%%%%%%%%%%%%%%%%%%%%%%%%%%%%%%%%%%
		
	\section{Details of inference algorithm updates}
		\subsection{NMF Gibbs sampling parameter values}
			The parameter values for the NMF Gibbs sampling algorithm are given in Table \ref{bnmf_updates}. These can be derived using Bayes' theorem as follows, using the notation defined in the main paper,
			\begin{alignat*}{1}
				p(U_{ik}|\tau,\U_{-ik},\V,\boldsymbol \lambda) 
				&\propto p(\R|\tau,\U,\V) \times p(U_{ik}|\lambda_k) \\
				&\propto \prod_{j \in \Omega^1_i} \mathcal{N} (R_{ij} | \U_i \cdot \V_j, \tau^{-1} ) \times \mathcal{E} ( U_{ik} | \lambda_k) \\
				&\propto \exp \left\{ - \frac{\tau}{2} \sum_{j \in \Omega^1_i} (R_{ij} - \U_i \V_j)^2 \right\} \times \exp \left\{ - \lambda_k U_{ik} \right\} \times u(x) \\
				&\propto \exp \left\{ - \frac{U_{ik}^2}{2} \left[ \displaystyle \tau \sum_{j \in \Omega^1_i} V_{jk}^2 \right] \right. \\
				& \left. \hspace{46pt} + U_{ik} \left[ - \lambda_k + \tau \sum_{j \in \Omega^1_i} \diffexclk V_{jk} \right] \right\} \times u(x) \\
				&\propto \exp \left\{ - \frac{\tauUik}{2} ( U_{ik} - \muUik )^2 \right\} \times u(x) \\
				&\propto \mathcal{TN} ( U_{ik} | \muUik, \tauUik ).
			\end{alignat*}
	
		\subsection{NMF Gibbs sampling and Variational Bayes parameter values}
			The parameter values for the NMF Variational Bayesian inference algorithm are given in Table \ref{bnmf_updates}. These can be derived using the following optimal expression for the variational posterior distribution (\cite{J.M.Bernardo}), 
			\begin{equation*}
			\log q^*(\theta_i) = \expqi \left[ \log p(\btheta, D) \right] + C,
			\end{equation*}
			allowing us to find the optimal updates for the variational parameters. We now take the expectation with respect to the distribution $ q(\btheta_{-i}) $ over the parameters but excluding the $i$th one. The derivation for $U_{ik}$ is shown below.
			We use $ \widetilde{f(X)} $ as a shorthand for $ \mathbb{E}_q \left[ f(X) \right] $, where $X$ is a random variable and $f$ is a function over $X$. In the code implementation of these updates we make use of the identity $ \widetilde{X^2} = \widetilde{X}^2 + \mathrm{Var}_q \left[ X \right] $.
			
			\begin{alignat*}{1}
				q^*(U_{ik})
				&\propto \exp \left\{ \expqUik \left[ \log p(D|\btheta) + \log p(\btheta) \right] \right\} \\
				&\propto \exp \left\{ \expqUik \left[ \sumOmegai \log p(R_{ij}|\U,\V) + \log p(U_{ik}|\lambda_k) \right] \right\} \\
				&\propto \exp \left\{ \expqUik \left[ \sumOmegai \log \left[ \sqrt{\frac{\tau}{2 \pi}} \exp \left\{ - \frac{\tau}{2} \left( R_{ij} - \U_i \V_j \right)^2 \right\} \right] + \log \left[ \lambda_k \exp \left\{ - \lambda_k U_{ik} \right\} \right] \right] \right\} \times u(x) \\
				&\propto \exp \left\{ \expqUik \left[ \sumOmegai - \frac{\tau}{2} \left( R_{ij} - \U_i \V_j \right)^2 - \lambda_k U_{ik} \right] \right\} \times u(x) \\
				&\propto \exp \left\{ \expqUik \left[ - \frac{\tau}{2} \sumOmegai \left[ U_{ik}^2 V_{jk}^2 - 2 U_{ik} V_{jk} \diffexclk \right] \right] - U_{ik} \explambdak \right\} \times u(x) \\
				&\propto \exp \left\{ - \frac{U_{ik}^2}{2} \left[ \exptau \sumOmegai \expVjk^2 \right] + U_{ik} \left[ - \explambdak + \exptau \sumOmegai \diffexpexclk \expVjk \right] \right\} \times u(x) \\
				&\propto \exp \left\{ - \frac{\tauUik}{2} ( U_{ik} - \muUik )^2 \right\} \times u(x) \\
				&\propto \mathcal{TN} ( U_{ik} | \muUik, \tauUik ).
			\end{alignat*}

		\begin{table*}[b]
			\caption{NMF variable update rules} \label{bnmf_updates}
			\begin{center}
				%{\renewcommand{\arraystretch}{1.25}
				\begin{tabular}{c|l|l}%{c|l|l}
					\toprule
					{\bf}  &{\bf \hspace{0.5pt} GIBBS SAMPLING}  &{\bf \hspace{0.5pt} VARIATIONAL BAYES} \\
					%\hline & \\
					%\midrule
					\toprule
					$\alpha_{\tau}^*$ & \hspace{1pt} $ \displaystyle \alpha_{\tau} + \frac{|\Omega|}{2} $		& \hspace{1pt} $ \displaystyle \alpha_{\tau} + \frac{|\Omega|}{2}	$	\\
					%\rule{0pt}{3ex}  
					$ \beta_{\tau}^* $			& \hspace{1pt} $ \displaystyle \beta_{\tau} + \frac{1}{2} \sumOmega (R_{ij} - \U_i \V_j)^2 $ 		& \hspace{1pt} $ \displaystyle \beta_{\tau} + \frac{1}{2} \sumOmega \expdiff $ \\
					\midrule %\rule{0pt}{3ex}  
					$\tauUik $			& \hspace{1pt} $ \displaystyle \displaystyle \tau \sumOmegai V_{jk}^2$ 		& \hspace{1pt} $ \displaystyle \exptau \sumOmegai \expVjksqr $		\\	
					%\rule{0pt}{3ex}
					$ \muUik $			& \hspace{1pt} $ \displaystyle \frac{1}{\tauUik} \Big( - \lambdak + $ 		& \hspace{1pt} $ \displaystyle \frac{1}{\tauUik} \Big( - \explambdak + $ \\
					& \hspace{25pt} $ \displaystyle \tau \sumOmegai \diffexclk V_{jk} \Big) $		& \hspace{25pt} $ \displaystyle \exptau \sumOmegai \diffexpexclk \expVjk \Big) $ \\
					\midrule %\rule{0pt}{3ex}
					$ \tauVjk $			& \hspace{1pt} $ \displaystyle \tau \sumOmegaj U_{ik}^2 $ 		& \hspace{1pt} $ \displaystyle \exptau \sumOmegaj \expUiksqr $ 	\\
					%\rule{0pt}{3ex}
					$ \muVjk $			& \hspace{1pt} $ \displaystyle \frac{1}{\tauVjk} \Big( - \lambdak + $ 			& \hspace{1pt} $ \displaystyle \frac{1}{\tauVjk} \Big( - \explambdak + $ \\
					& \hspace{25pt} $ \displaystyle \tau \sumOmegaj \diffexclk U_{ik} \Big) $		& \hspace{25pt} $ \displaystyle \exptau \sumOmegaj \diffexpexclk \expUik \Big) $ \\
					\midrule %\rule{0pt}{3ex}  
					$ \alpha^*_k $ 	& 	\hspace{1pt} $ \displaystyle \alpha_0 + I + J $	&	\hspace{1pt} $ \displaystyle \alpha_0 + I + J $	\\
					$ \beta^*_k $ 	& 	\hspace{1pt} $ \displaystyle \beta_0 + \sum_{i=1}^I U_{ik} + \sum_{j=1}^J V_{jk} $	&	\hspace{1pt} $ \displaystyle \beta_0 + \sum_{i=1}^I \expUik + \sum_{j=1}^J \expVjk $	\\
					%& & \\
					%\hline 
					%\multicolumn{3}{c}{} \\
					\midrule
					\multicolumn{3}{c}{ $ \displaystyle \expdiff = \diffexp^2 + \sum_{k=1}^{K} \left( \expUiksqr \expVjksqr - \expUik^2 \expVjk^2 \right) $ } \\ 
					%\multicolumn{3}{c}{} \\
					%\hline 
					\bottomrule
				\end{tabular}
			\end{center}
		\end{table*}

%%%%%%%%%%%%%%%%%%%%%%%%%%%%%%%%%%%%%%%%%%%%%%%%%%%%%%%%%%%%%%%%
		
		\subsection{NMTF Gibbs sampling parameter values}
		For the NMTF Gibbs sampling algorithm we need to sample from the following posteriors:
		\begin{alignat*}{2}
			p(\tau | \F, \S, \G, \boldsymbol \lambda^F, \boldsymbol \lambda^G, D) &= \mathcal{G} (\tau | \alpha_{\tau}^*, \beta_{\tau}^* ) \\
			p(F_{ik} | \tau, \F_{-ik}, \S, \G, \boldsymbol \lambda^F, \boldsymbol \lambda^G, D) &= \mathcal{TN} ( F_{ik} | \muFik, \tauFik ) \\
			p(S_{kl} | \tau, \F, \S_{-kl}, \G, \boldsymbol \lambda^F, \boldsymbol \lambda^G, D) &= \mathcal{TN} ( S_{kl} | \muSkl, \tauSkl ) \\
			p(G_{jl} | \tau, \F, \S, \G_{-jl}, \boldsymbol \lambda^F, \boldsymbol \lambda^G, D) &= \mathcal{TN} ( G_{jl} | \muGjl, \tauGjl )  \\
			p(\lambda^F_k | \F, \S, \G, \boldsymbol \lambda^F_{-k}, \boldsymbol \lambda^G, D) &= \mathcal{G} (\lambda^F_k | \alpha^{F*}_k, \beta^{F*}_k ) \\
			p(\lambda^G_l | \F, \S, \G, \boldsymbol \lambda^F, \boldsymbol \lambda^G_{-l}, D) &= \mathcal{G} (\lambda^G_l | \alpha^{G*}_l, \beta^{G*}_l ).
		\end{alignat*}
		In the above, $\boldsymbol \lambda^F$ is a vector including all $\lambda^F_k$ values, $\boldsymbol \lambda^F_{-k}$ excludes $\lambda^F_k$, and similarly for $\boldsymbol \lambda^G$.
		The updates are given in Table \ref{bnmtf_gibbs_updates}.
		
		\begin{table*}[b]
			\caption{NMTF Gibbs update rules.} \label{bnmtf_gibbs_updates}
			\begin{center}
				\begin{tabular}{c|l}
					{\bf}  &{\bf GIBBS SAMPLING} \\
					\toprule
					$ \alpha_{\tau}^* $ 		& $ \displaystyle \alpha_{\tau} + \frac{|\Omega|}{2} $	\\
					$ \beta_{\tau}^* $			& $ \displaystyle \beta_{\tau} + \frac{1}{2} \sumOmega\diffTRI^2 $	 \\
					\midrule
					$ \tauFik $			& $ \displaystyle \tau \sumOmegai \left( \S_k \cdot \G_j \right)^2 $		\\
					$ \muFik $			& $ \displaystyle \frac{1}{\tauFik} \left(  - \lambdaFk + \tau \sumOmegai \diffTRIexclK \left( \S_k \cdot \G_j \right) \right) $ \\	
					\midrule
					$ \tauSkl $			& $ \displaystyle \tau \sumOmega F_{ik}^2 G_{jl}^2 $		\\
					$ \muSkl $			& $ \displaystyle \frac{1}{\tauSkl} \left( - \lambdaSkl + \tau \sumOmega \diffTRIexclKL F_{ik} G_{jl} \right) $ \\
					\midrule
					$ \tauGjl $			& $ \displaystyle \tau \sumOmegaj \left( \F_i \cdot \S_{\cdot,l} \right)^2 $ \\
					$ \muGjl $			& $ \displaystyle \frac{1}{\tauGjl} \left( - \lambdaGl + \tau \sumOmegaj \diffTRIexclL \left( \F_i \cdot \S_{\cdot,l} \right) \right) $ \\
					\midrule
					$ \alpha_k^{F*} $ 		& $ \displaystyle \alpha_0 + I $	\\
					$ \beta_k^{F*} $		& $ \displaystyle \beta_0 + \sum_{i=1}^I F_{ik} $	 \\
					\midrule
					$ \alpha_l^{G*} $ 		& $ \displaystyle \alpha_0 + J $	\\
					$ \beta_l^{G*} $		& $ \displaystyle \beta_0 + \sum_{j=1}^J G_{jl} $	 \\
					\bottomrule
				\end{tabular}
			\end{center}
		\end{table*}
		
%%%%%%%%%%%%%%%%%%%%%%%%%%%%%%%%%%%%%%%%%%%%%%%%%%%%%%%%%%%%%%%%
	
		\subsection{NMTF Variational Bayes parameter updates}
		We have the following approximations to the posteriors for the NMTF Variational Bayes algorithm:
		\begin{alignat*}{2}
			q(\tau) &= \mathcal{G} (\tau | \alpha_{\tau}^*, \beta_{\tau}^* ) \\
			q(F_{ik}) &= \mathcal{TN} ( F_{ik} | \muFik, \tauFik ) \\
			q(S_{kl}) &= \mathcal{TN} ( S_{kl} | \muSkl, \tauSkl ) \\
			q(G_{jl}) &= \mathcal{TN} ( G_{jl} | \muGjl, \tauGjl )  \\
			q(\lambda^F_k) &= \mathcal{G} (\lambda^F_k | \alpha^{F*}_k, \beta^{F*}_k ) \\
			q(\lambda^G_l) &= \mathcal{G} (\lambda^G_l | \alpha^{G*}_l, \beta^{G*}_l ).
		\end{alignat*}
		The updates are given in Table \ref{bnmtf_vb_updates}. The expression for $\expdiffTRI$ can be found in the main paper.
		
		\begin{table*}[b]
			\caption{NMTF VB update rules.} \label{bnmtf_vb_updates}
			\begin{center}
				\begin{tabular}{c|l}
					\toprule
					{\bf}  &{\bf VARIATIONAL BAYES} \\
					\midrule
					$ \alpha_{\tau}^* $ 		& $ \displaystyle \alpha_{\tau} + \frac{|\Omega|}{2} $	\\
					$ \beta_{\tau}^* $			& $ \displaystyle \beta_{\tau} + \frac{1}{2} \sumOmega \expdiffTRI $	  \\
					\midrule
					$ \tauFik $			& $ \displaystyle \exptau \sumOmegai \left( \left( \sum_{l=1}^L \expSkl \expGjl \right)^2 + \sum_{l=1}^L \left( \expSklsqr \expGjlsqr - \expSkl^2 \expGjl^2 \right) \right) $		\\
					$ \muFik $			& $ \displaystyle \frac{1}{\tauFik} \left( - \explambdaFk + \exptau \sumOmegai \left( \diffexpTRIexclK \sum_{l=1}^L \expSkl \expGjl - \suml \expSkl \varGjl \sumexclk \expFikp \expSkpl \right) \right) $ \\	
					\midrule
					$ \tauSkl $			& $ \displaystyle \exptau \sumOmega \expFiksqr \expGjlsqr $		\\
					$ \muSkl $			& $ \displaystyle \frac{1}{\tauSkl} \left( - \lambdaSkl + \exptau \sumOmega \left( \diffexpTRIexclKL \expFik \expGjl \right. \right. $ \\
										& \hspace{117pt} $ \displaystyle \left. \left. - \expFik \varGjl \sumexclk \expFikp \expSkpl  - \varFik \expGjl \sumexcll \expSklp \expGjlp \right) \right) $ \\
					\midrule
					$ \tauGjl $			& $ \displaystyle \exptau \sumOmegaj \left( \left( \sum_{k=1}^K \expFik \expSkl \right)^2 + \sum_{k=1}^K \left( \expFiksqr \expSklsqr - \expFik^2 \expSkl^2 \right) \right) $ \\
					$ \muGjl $			& $ \displaystyle \frac{1}{\tauGjl} \left( - \explambdaGl + \exptau \sumOmegaj \left( \diffexpTRIexclL \sum_{k=1}^K \expFik \expSkl - \sumk \varFik \expSkl \sumexcll \expSklp \expGjlp \right) \right) $ \\ 
					\midrule
					$ \alpha_k^{F*} $ 		& $ \displaystyle \alpha_0 + I $	\\
					$ \beta_k^{F*} $		& $ \displaystyle \beta_0 + \sum_{i=1}^I \expFik $	 \\
					\midrule
					$ \alpha_l^{G*} $ 		& $ \displaystyle \alpha_0 + J $	\\
					$ \beta_l^{G*} $		& $ \displaystyle \beta_0 + \sum_{j=1}^J \expGjl $	 \\
					\bottomrule
				\end{tabular}
			\end{center}
		\end{table*}
		
%%%%%%%%%%%%%%%%%%%%%%%%%%%%%%%%%%%%%%%%%%%%%%%%%%%%%%%%%%%%%%%%

	\clearpage
	\section{Implementation details}
		All algorithms mentioned were implemented using the Python language. The \textit{numpy} package was used for fast matrix operations, and for random draws of the truncated normal distribution we used the Python package \textit{rtnorm} by C. Lassner (\url{http://miv.u-strasbg.fr/mazet/rtnorm/}), giving more efficient draws than the standard libraries and dealing with rounding errors. \\
		
		\noindent The mean and variance of the truncated normal involve operations prone to numerical errors when $ \mu $ takes high negative values. To deal with this we observe that when $ \mu \tau \ll 0 $ the truncated normal distribution approximates an exponential one with rate $ \vert \mu \tau \vert $, and therefore has mean $ 1 / \vert \mu \tau \vert $ and variance $ 1 / \vert \mu \tau \vert^2 $. \\
		
		\noindent All experiments were run on a MacBook Pro laptop, with 2.2 GHz Intel Core i7 processor, 16 GB 1600 MHz DDR3 memory, and an Intel Iris Pro 1536 MB Graphics card.	
		
%%%%%%%%%%%%%%%%%%%%%%%%%%%%%%%%%%%%%%%%%%%%%%%%%%%%%%%%%%%%%%%%

	\section{Data preprocessing}
		We will now describe the preprocessing steps undertaken for the drug sensitivity datasets used in the paper. We used four different datasets:
		\begin{itemize}
			\item Genomics of Drug Sensitivity in Cancer (GDSC v5.0, \citet{Yang2013})---giving the natural log of $IC_{50}$ values for 139 drugs across 707 cell lines, with 80\% observed entries.
			\item Cancer Therapeutics Response Portal (CTRP v2, \cite{Seashore-Ludlow2015})---giving $EC_{50}$ values for 545 drugs across 887 cell lines, with 80\% observed entries.
			\item Cancer Cell Line Encyclopedia (CCLE, \cite{Barretina2012})---giving both $IC_{50}$ and $EC_{50}$ values for 24 drugs across 504 cell lines, with 96\% and 63\% observed entries respectively.
		\end{itemize}
		$IC_{50}$ values indicate the required drug concentration needed to reduce the activity of a given cell line (cancer type in a tissue) by half. We thus measure when an undesired effect has been inhibited by half. With $EC_{50}$ values we measure the maximal (desired) effect a drug can have on a cell line, and then measure the concentration of the drug where we achieve half of this value. In both cases, a lower value is better. \\
		
		\noindent The values in the CCLE datasets were in the range [0,8] for $IC_{50}$, and [0,10] for $EC_{50}$. The CTRP values were all nonnegative, with some very high values, so we capped them at 100. For the GDSC dataset we undid the natural log transform by taking the exponent, making all values nonnegative, and then also capped high values at 100. For this dataset we also had two cell lines with only one and two observed entries, so we filtered them out. \\
		
\begin{figure}
			\centering
			\captionsetup{width=\columnwidth}
			\begin{subfigure}{0.24 \columnwidth}
				\includegraphics[width=\columnwidth]{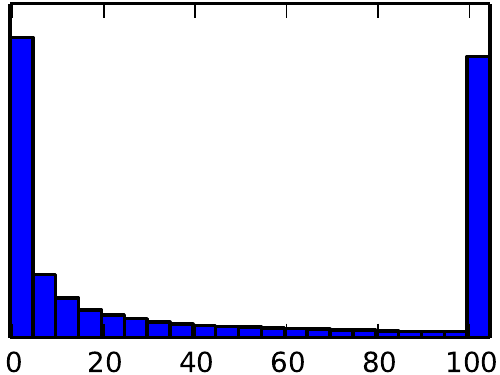}
				\captionsetup{width=0.9\columnwidth}
				\caption{GDSC $IC_{50}$} 
			\end{subfigure} %
			\begin{subfigure}{0.24 \columnwidth}
				\includegraphics[width=\columnwidth]{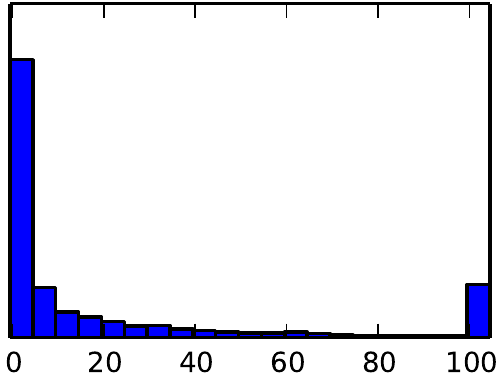}
				\captionsetup{width=0.9\columnwidth}
				\caption{CTRP $EC_{50}$} 
			\end{subfigure}
			\begin{subfigure}{0.24 \columnwidth}
				\includegraphics[width=\columnwidth]{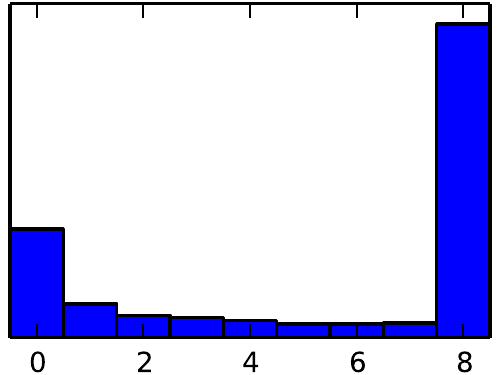}
				\captionsetup{width=0.9\columnwidth}
				\caption{CCLE $IC_{50}$} 
			\end{subfigure} %
			\begin{subfigure}{0.24 \columnwidth}
				\includegraphics[width=\columnwidth]{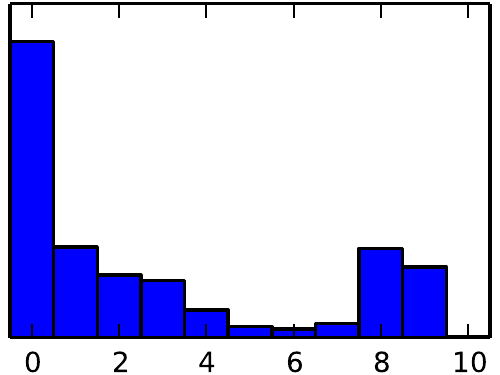}
				\captionsetup{width=0.9\columnwidth}
				\caption{CCLE $EC_{50}$} 
			\end{subfigure}
			\caption{Plots of the distribution of values in the drug sensitivity datasets, after capping the extremely high values in the CTRP $EC_{50}$ and GDSC $IC_{50}$ datasets to 100.}
			\label{distribution_drug_sensitivity}
		\end{figure}

		\noindent Distributions of the values are plotted in Figure \ref{distribution_drug_sensitivity}. A summary of the datasets can also be found in Table \ref{summary_drug_sensitivity}.
		
		\begin{table*}
			\captionsetup{width=0.8\columnwidth}
			\caption{Overview of the four drug sensitivity datasets, giving the number of cell lines (rows), drugs (columns), observed entries, and the fraction of entries that are observed.} \label{summary_drug_sensitivity}
			\centering
			\begin{tabular}{lcccc}
				\toprule
				Dataset & Cell lines & Drugs & Entries observed & Fraction observed \\
				\midrule
				GDSC $IC_{50}$ & 707 & 139 & 79262 & 0.806 \\
				CTRP $EC_{50}$ & 887 &  545 & 387130 & 0.801 \\
				CCLE $IC_{50}$ & 504 & 24 & 11670 & 0.965 \\
				CCLE $EC_{50}$ & 504 & 24 & 7626 & 0.630 \\
				\bottomrule
			\end{tabular}
		\end{table*}
			
%%%%%%%%%%%%%%%%%%%%%%%%%%%%%%%%%%%%%%%%%%%%%%%%%%%%%%%%%%%%%%%%
		
	\section{Additional results}
		\subsection{Convergence speed against time}
			In the main paper we plotted the convergence speeds of the inference algorithms against number of iterations taken. Here, we also give the convergence speed against time taken. The results are given in Figure \ref{convergence_speed}, with the average time per iteration in Table \ref{table_time_per_iteration} below. We can see that the ICM and NP methods can be implemented much more efficiently than the fully Bayesian models, leading to even faster convergence. However, as discussed in the main paper, as a result of fitting more and faster to the data, the ICM and NP approaches are also less robust to noise and sparsity. Finally, note the weird fitting behaviour of NP-NMF on the synthetic data, which occasionally happens.
			
			\begin{table*}[h]
				\captionsetup{width=0.835\columnwidth}
				\caption{Average time (in seconds) taken per iteration of the four inference approaches for NMF and NMTF, on the synthetic and four drug sensitivity datasets.} \label{table_time_per_iteration}
				\centering
				\begin{tabular}{lccccc}
					\toprule
					Method \hspace{25pt} & Synthetic & GDSC $IC_{50}$ & CTRP $EC_{50}$ & CCLE $IC{50}$ & CCLE $EC{50}$ \\
					\midrule
					NMF VB & 0.015 & 0.125 & 0.387 & 0.067 & 0.064 \\
					NMF Gibbs & 0.024 & 0.251 & 0.655 & 0.175 & 0.143 \\
					NMF ICM & 0.003 & 0.047 & 0.279 & 0.012 & 0.012 \\
					NMF NP & 0.002 & 0.042 & 0.268 & 0.010 & 0.013 \\
					NMTF VB & 0.019 & 0.298 & 1.703 & 0.114 & 0.111 \\
					NMTF Gibbs & 0.014 & 0.264 & 1.557 & 0.107 & 0.107 \\
					NMTF ICM & 0.005 & 0.173 & 1.259 & 0.035 & 0.034 \\
					NMTF NP & 0.004 & 0.124 & 0.697 & 0.026 & 0.030 \\
					\bottomrule
				\end{tabular}
			\end{table*}
			
			\begin{figure*}[b]
				\centering
				\begin{subfigure}[t]{\columnwidth}
					\hspace{90pt}
					\includegraphics[width=0.6\columnwidth]{legend.png}
					\vspace{3pt}
				\end{subfigure}
				\begin{subfigure}[t]{0.19 \columnwidth}
					\includegraphics[width=\columnwidth]{toy_mse_nmf_convergences.png}
				\end{subfigure} %
				\begin{subfigure}[t]{0.19 \columnwidth}
					\includegraphics[width=\columnwidth]{gdsc_mse_nmf_convergences_gdsc.png}
				\end{subfigure} %
				\begin{subfigure}[t]{0.19 \columnwidth}
					\includegraphics[width=\columnwidth]{ctrp_mse_nmf_convergences_ctrp.png}
				\end{subfigure} %
				\begin{subfigure}[t]{0.19 \columnwidth}
					\includegraphics[width=\columnwidth]{ccle_ic_mse_nmf_convergences_ccle_ic.png}
				\end{subfigure} %
				\begin{subfigure}[t]{0.19 \columnwidth}
					\includegraphics[width=\columnwidth]{ccle_ec_mse_nmf_convergences_ccle_ec.png}
				\end{subfigure} %
				\begin{subfigure}[t]{0.19 \columnwidth}
					\includegraphics[width=\columnwidth]{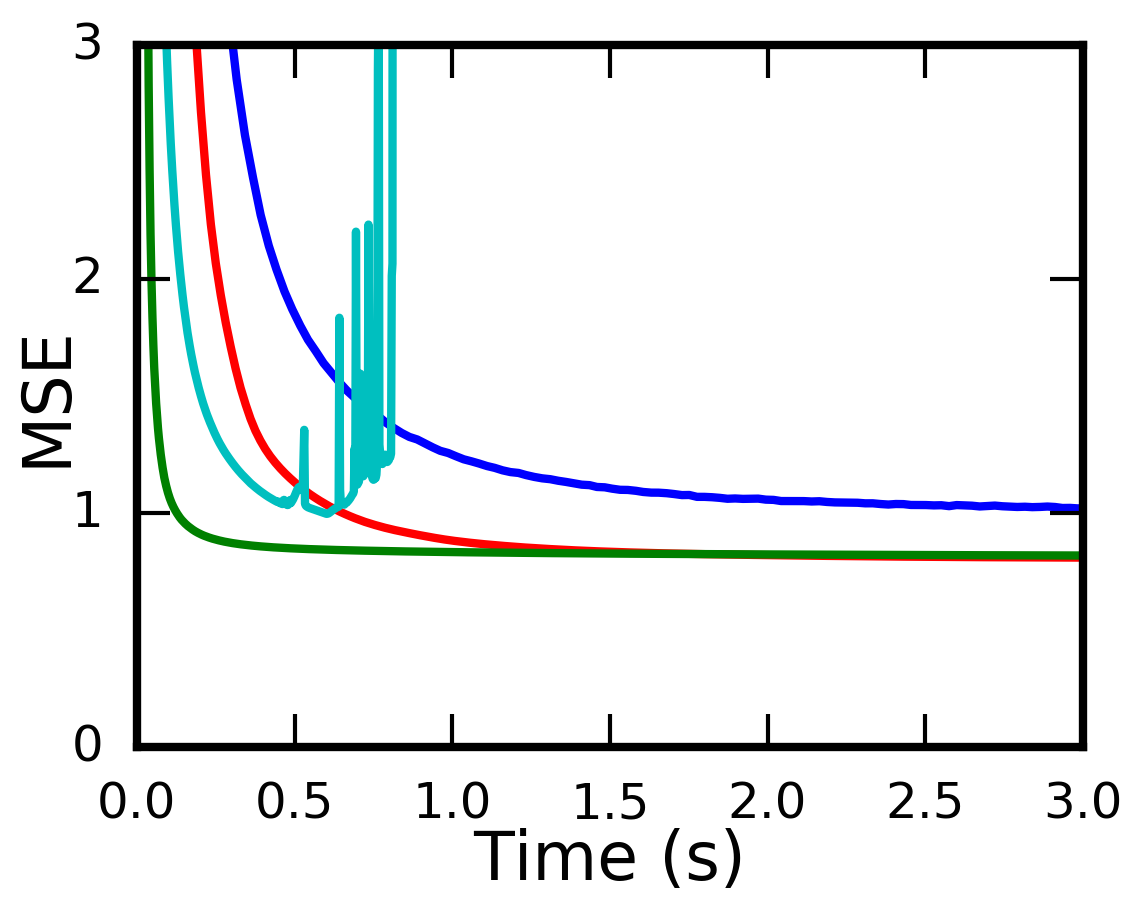}
					\captionsetup{width=\columnwidth}
					\caption{NMF, synthetic} 
					\label{nmf_convergence_toy}
				\end{subfigure} %
				\begin{subfigure}[t]{0.19 \columnwidth}
					\includegraphics[width=\columnwidth]{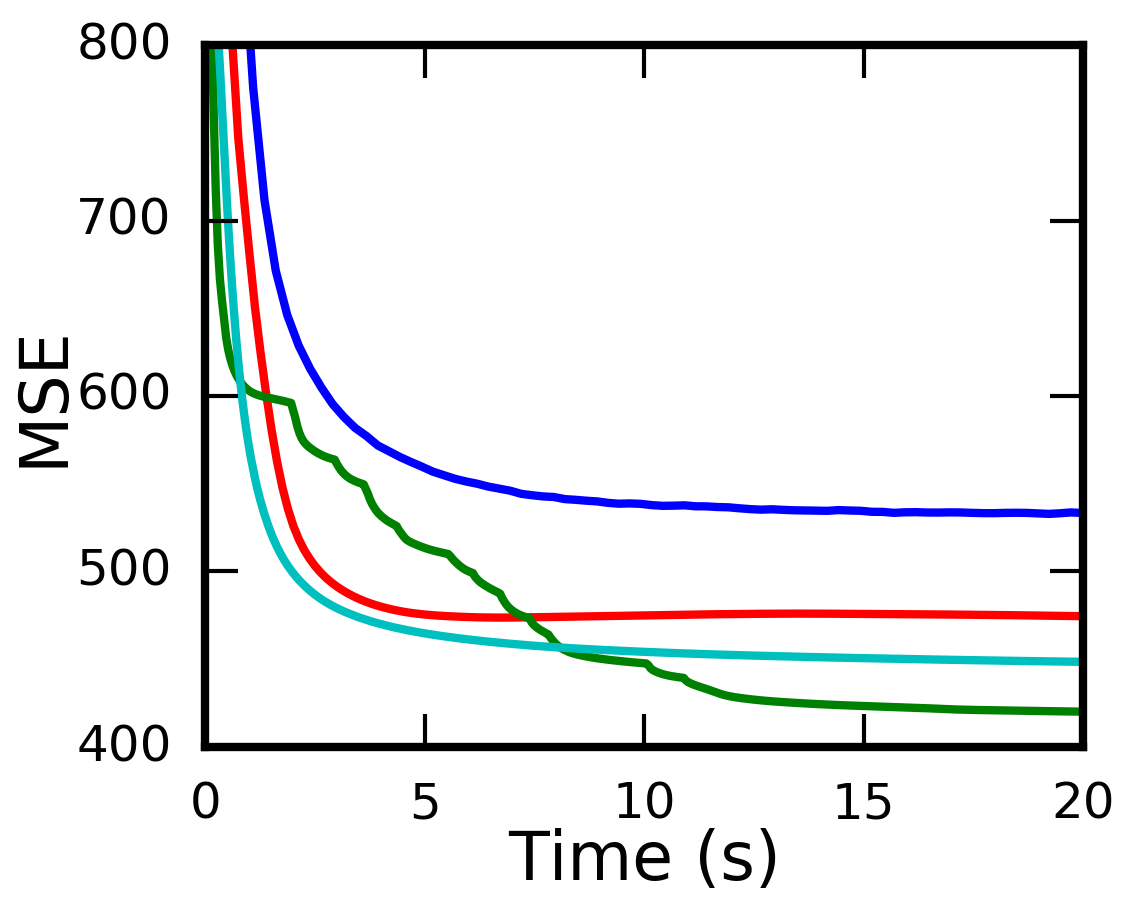}
					\captionsetup{width=\columnwidth}
					\caption{NMF, GDSC} 
					\label{nmf_convergence_gdsc}
				\end{subfigure} %
				\begin{subfigure}[t]{0.19 \columnwidth}
					\includegraphics[width=\columnwidth]{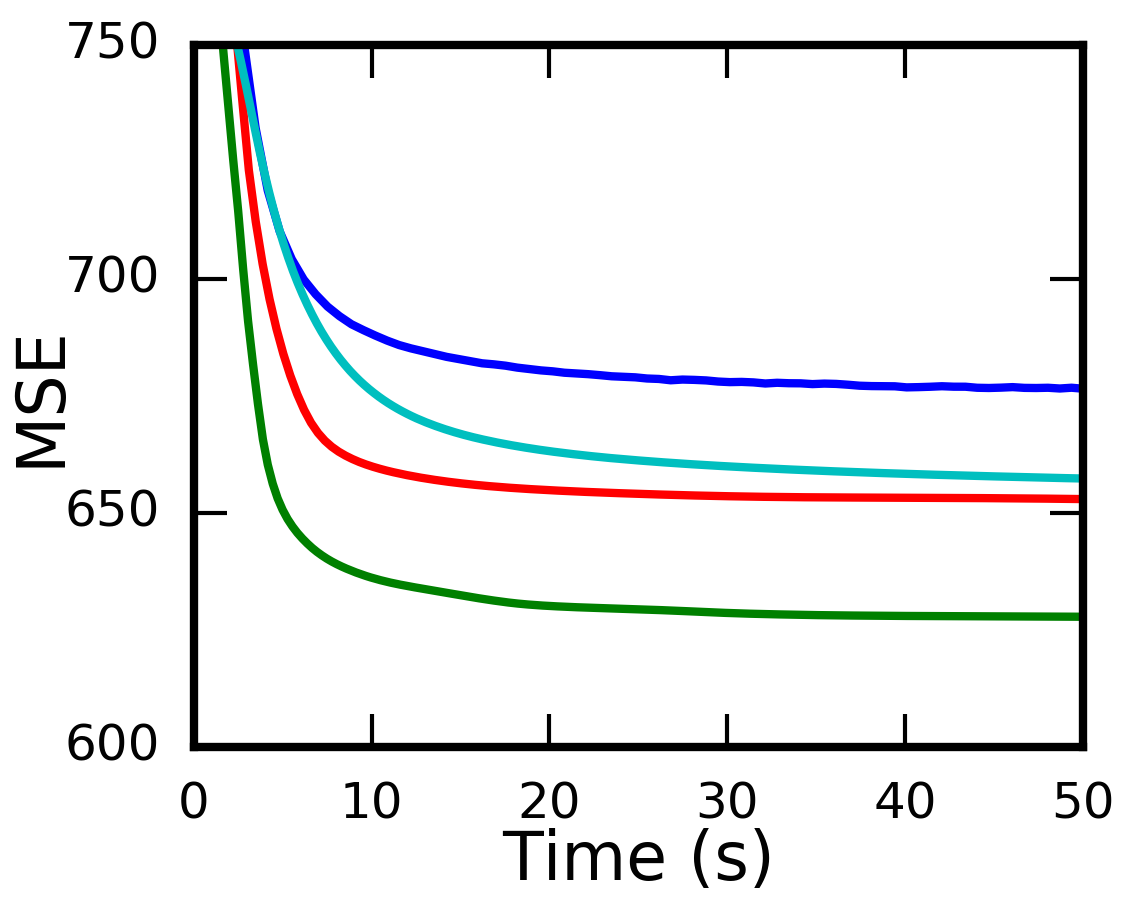}
					\captionsetup{width=\columnwidth}
					\caption{NMF, CTRP} 
					\label{nmf_convergence_ctrp}
				\end{subfigure} %
				\begin{subfigure}[t]{0.19 \columnwidth}
					\includegraphics[width=\columnwidth]{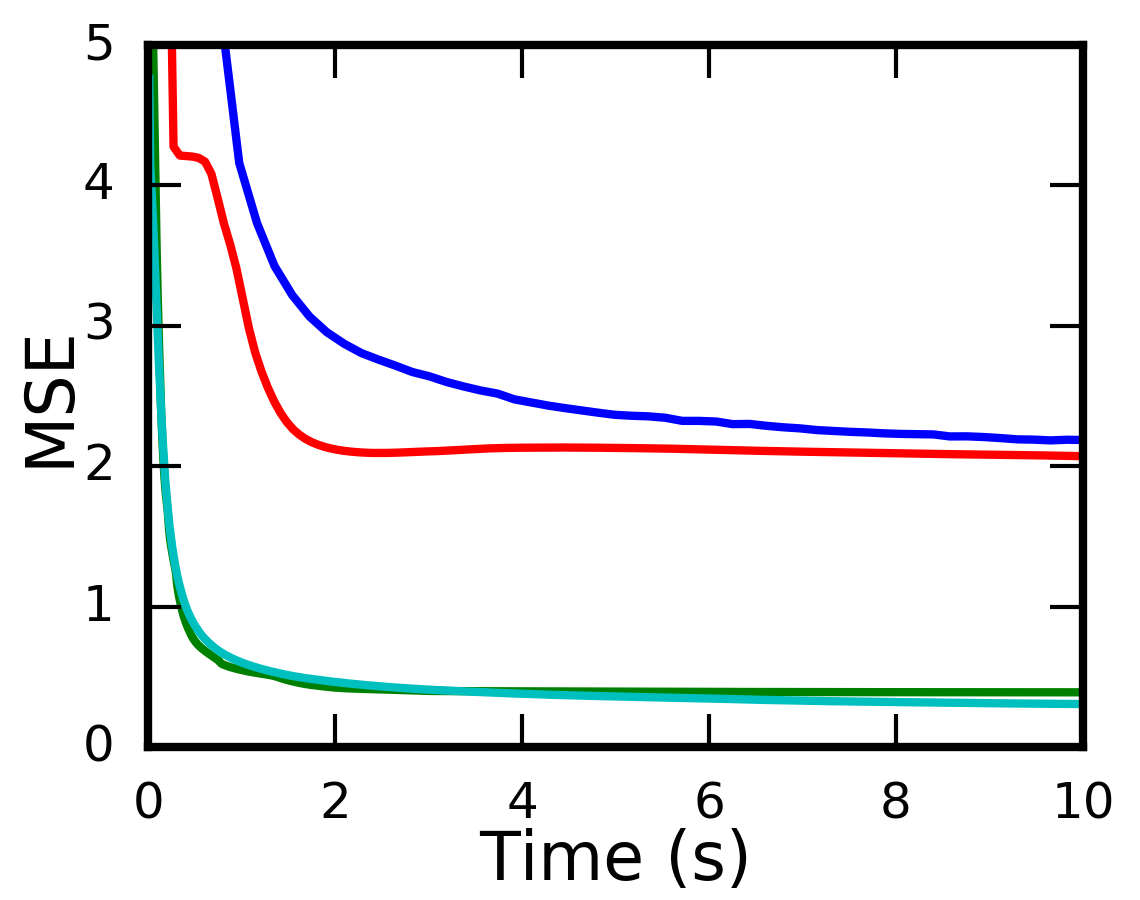}
					\captionsetup{width=\columnwidth}
					\caption{NMF, CCLE $IC_{50}$} 
					\label{nmf_convergence_ccle_ic}
				\end{subfigure} %
				\begin{subfigure}[t]{0.19 \columnwidth}
					\includegraphics[width=\columnwidth]{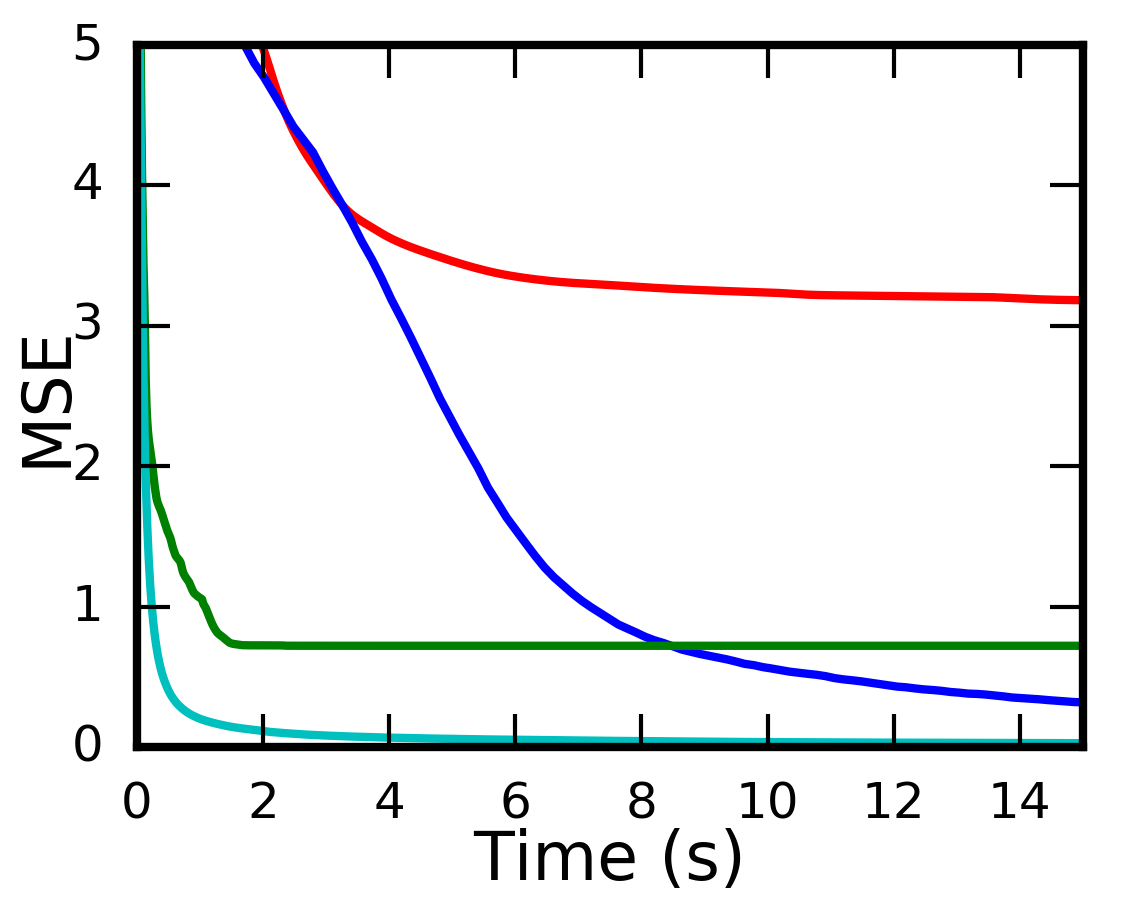}
					\captionsetup{width=\columnwidth}
					\caption{NMF, CCLE $EC_{50}$} 
					\label{nmf_convergence_ccle_ec}
					\vspace{5pt}
				\end{subfigure} %
				\begin{subfigure}[t]{0.19 \columnwidth}
					\includegraphics[width=\columnwidth]{toy_mse_nmtf_convergences.png}
				\end{subfigure} %
				\begin{subfigure}[t]{0.19 \columnwidth}
					\includegraphics[width=\columnwidth]{gdsc_mse_nmtf_convergences_gdsc.png}
				\end{subfigure} %
				\begin{subfigure}[t]{0.19 \columnwidth}
					\includegraphics[width=\columnwidth]{ctrp_mse_nmtf_convergences_ctrp.png}
				\end{subfigure} %
				\begin{subfigure}[t]{0.19 \columnwidth}
					\includegraphics[width=\columnwidth]{ccle_ic_mse_nmtf_convergences_ccle_ic.png}
				\end{subfigure} %
				\begin{subfigure}[t]{0.19 \columnwidth}
					\includegraphics[width=\columnwidth]{ccle_ec_mse_nmtf_convergences_ccle_ec.png}
				\end{subfigure} %
				\begin{subfigure}[t]{0.19 \columnwidth}
					\includegraphics[width=\columnwidth]{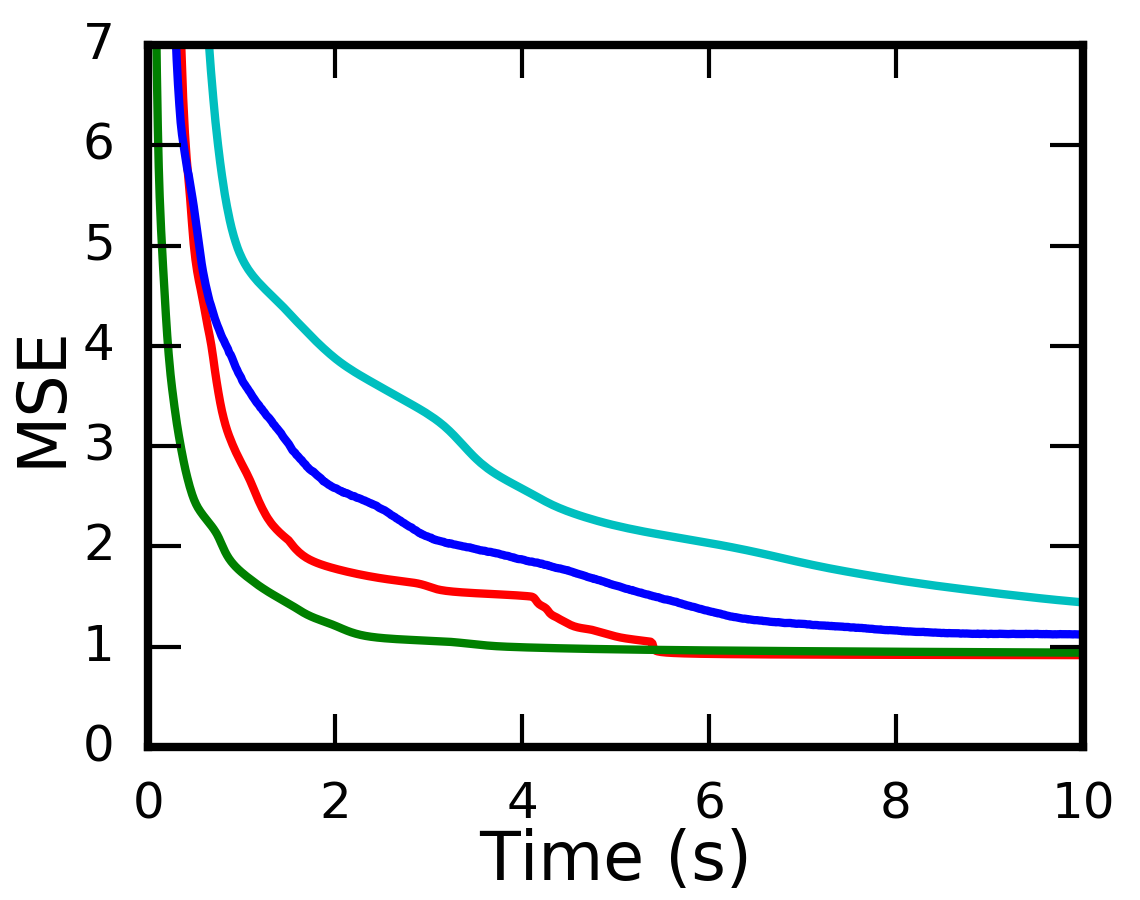}
					\captionsetup{width=\columnwidth}
					\caption{NMTF, synthetic} 
					\label{nmtf_convergence_toy}
				\end{subfigure} %
				\begin{subfigure}[t]{0.19 \columnwidth}
					\includegraphics[width=\columnwidth]{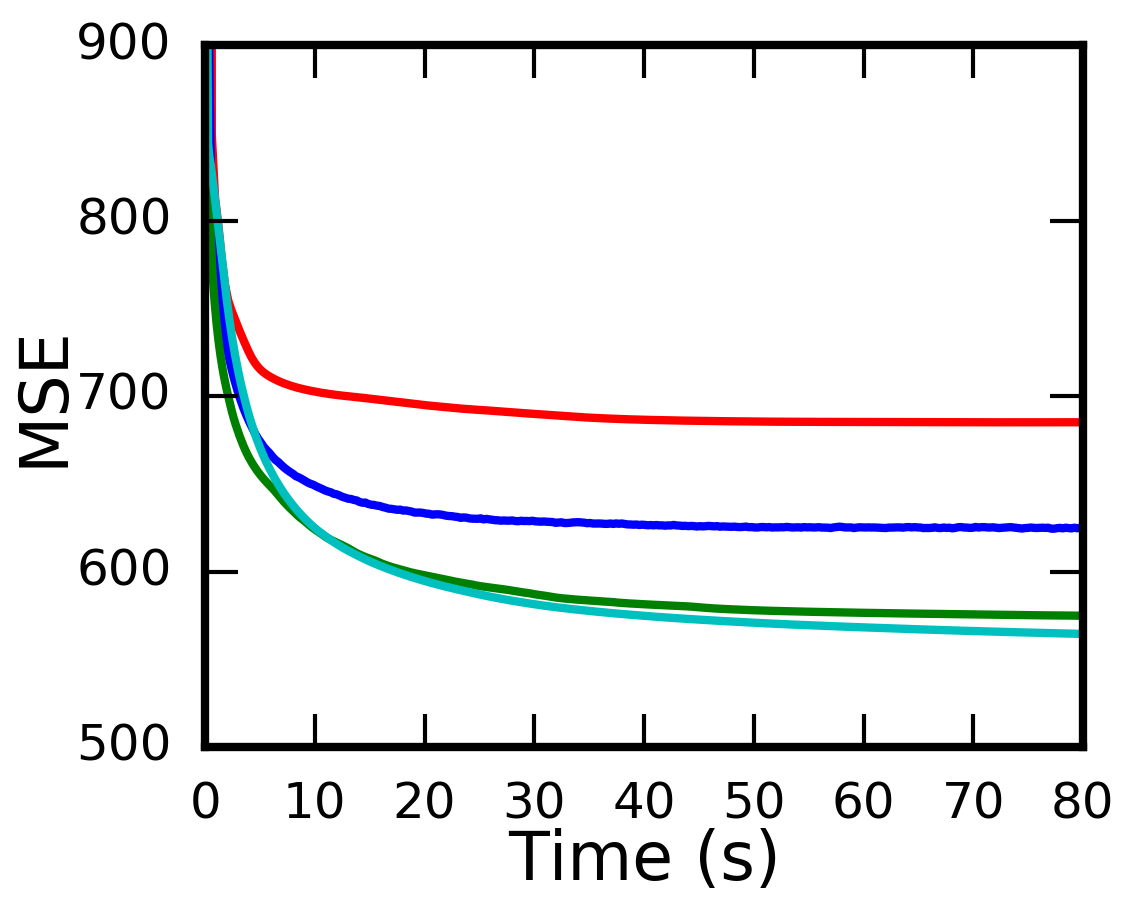}
					\captionsetup{width=\columnwidth}
					\caption{NMTF, GDSC} 
					\label{nmtf_convergence_gdsc}
				\end{subfigure} %
				\begin{subfigure}[t]{0.19 \columnwidth}
					\includegraphics[width=\columnwidth]{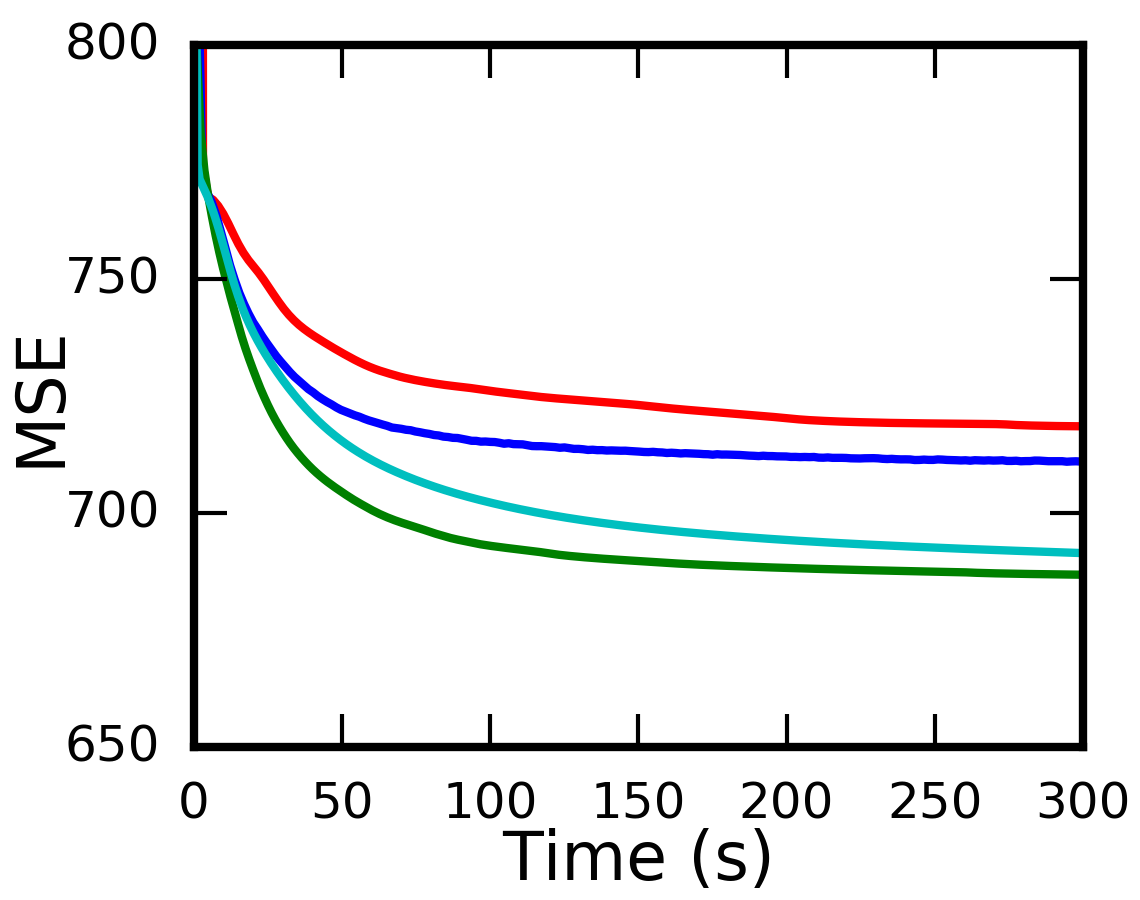}
					\captionsetup{width=\columnwidth}
					\caption{NMTF, CTRP} 
					\label{nmtf_convergence_ctrp}
				\end{subfigure} %
				\begin{subfigure}[t]{0.19 \columnwidth}
					\includegraphics[width=\columnwidth]{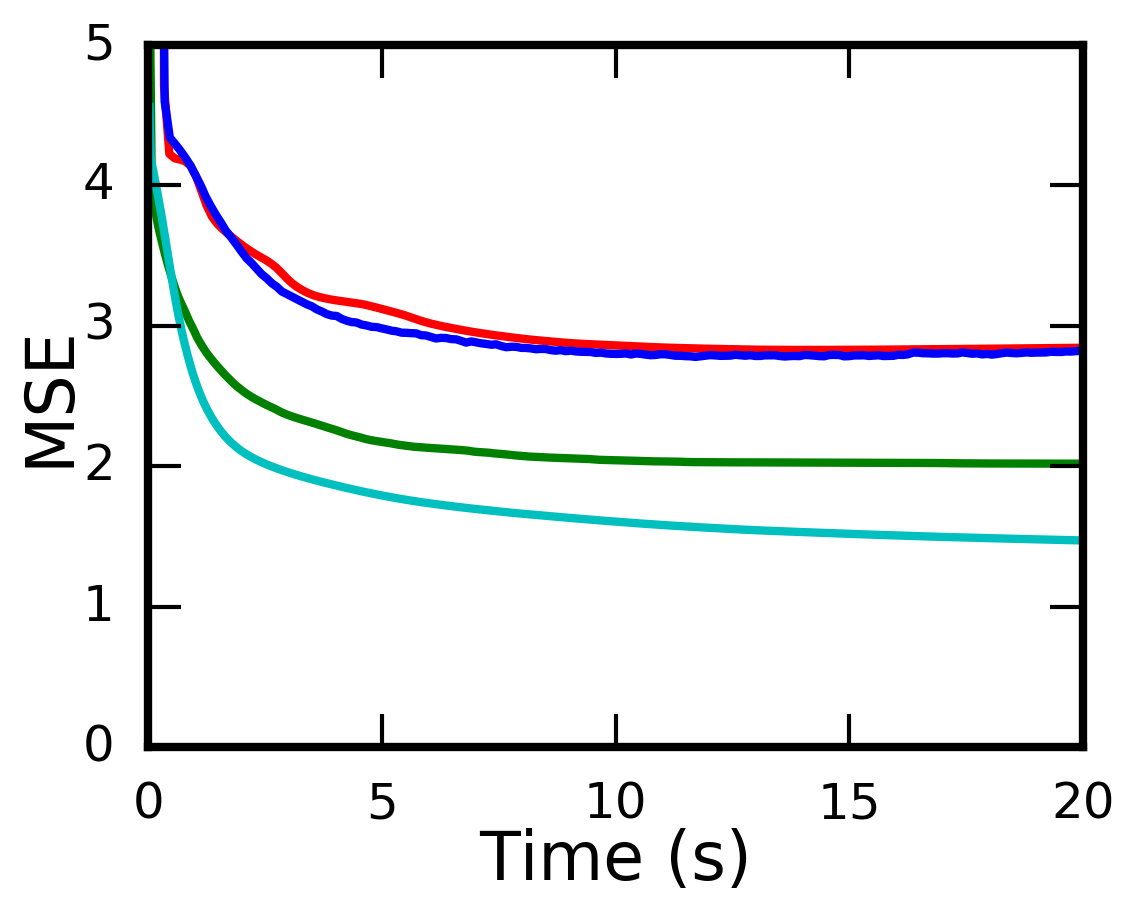}
					\captionsetup{width=\columnwidth}
					\caption{NMTF, CCLE $IC_{50}$} 
					\label{nmtf_convergence_ccle_ic}
				\end{subfigure} %
				\begin{subfigure}[t]{0.19 \columnwidth}
					\includegraphics[width=\columnwidth]{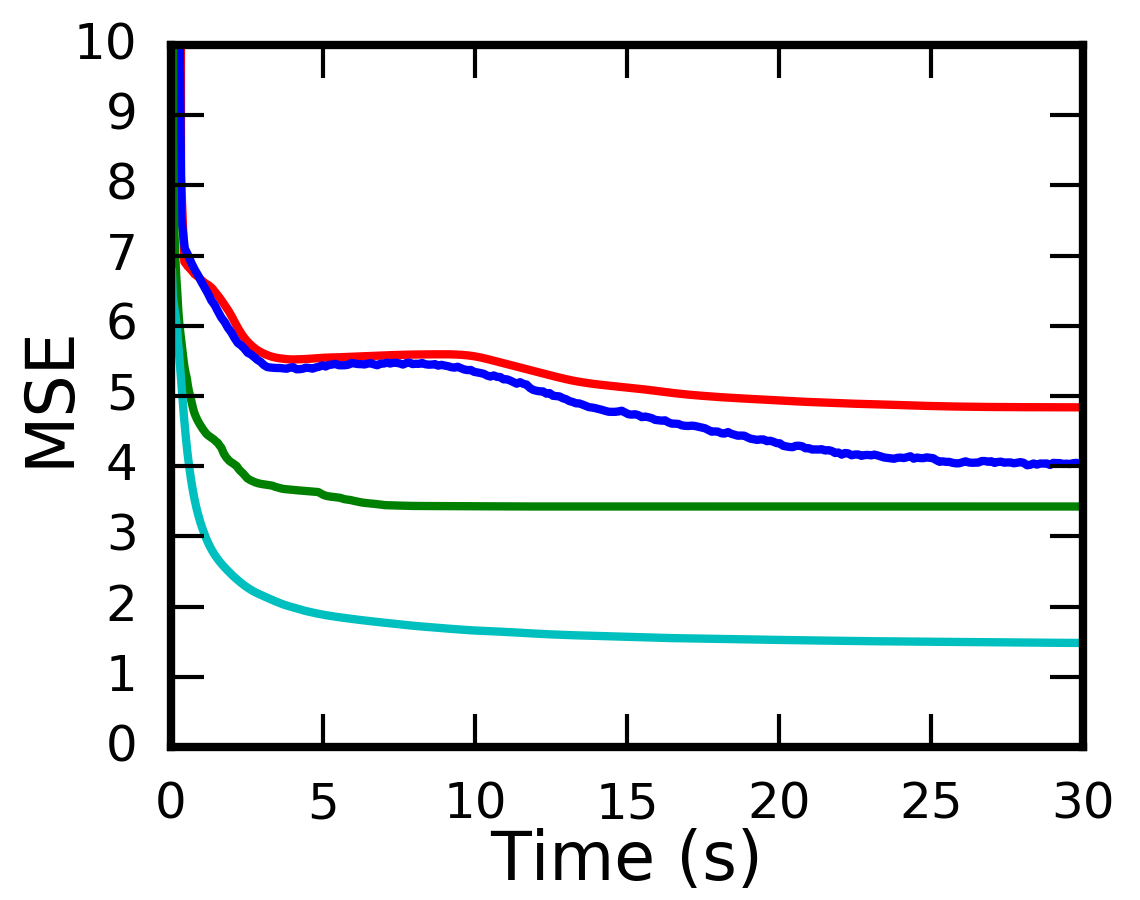}
					\captionsetup{width=\columnwidth}
					\caption{\small{NMTF, CCLE $EC_{50}$}}
					\label{nmtf_convergence_ccle_ec}
				\end{subfigure} %
				\captionsetup{width=\columnwidth}
				\caption{Convergence of algorithms on the synthetic and drug sensitivity datasets, measuring the training data fit (mean square error) across iterations and time taken, for each of the inference approaches for NMF (top two rows) and NMTF (bottom two rows).}
				\label{convergence_speed}
			\end{figure*}

		\subsection{Noise test with ARD}
			In the main paper we presented the results of the noise test for both NMF and NMTF on the synthetic data. Here, we demonstrate that the addition of ARD does not have an impact on the robustness to noise for each of the three probabilistic models (ICM, VB and Gibbs). The results are given in Figure \ref{noise_test_ard} below, where we see no difference in predictive performance for any of the models.
			
			\begin{figure*}
				\centering
				\begin{subfigure}[t]{\columnwidth}
					\hspace{75pt}
					\includegraphics[width=0.7\columnwidth]{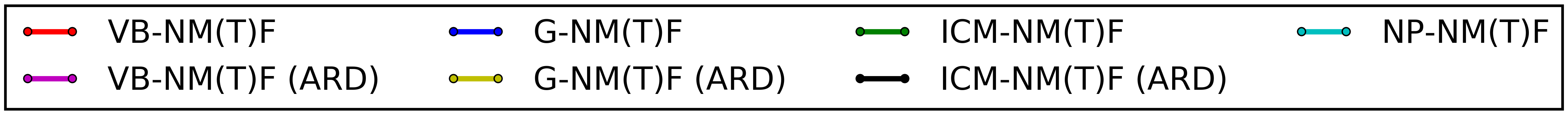}
				\end{subfigure}
				\begin{subfigure}{0.45 \columnwidth}
					\includegraphics[width=\columnwidth]{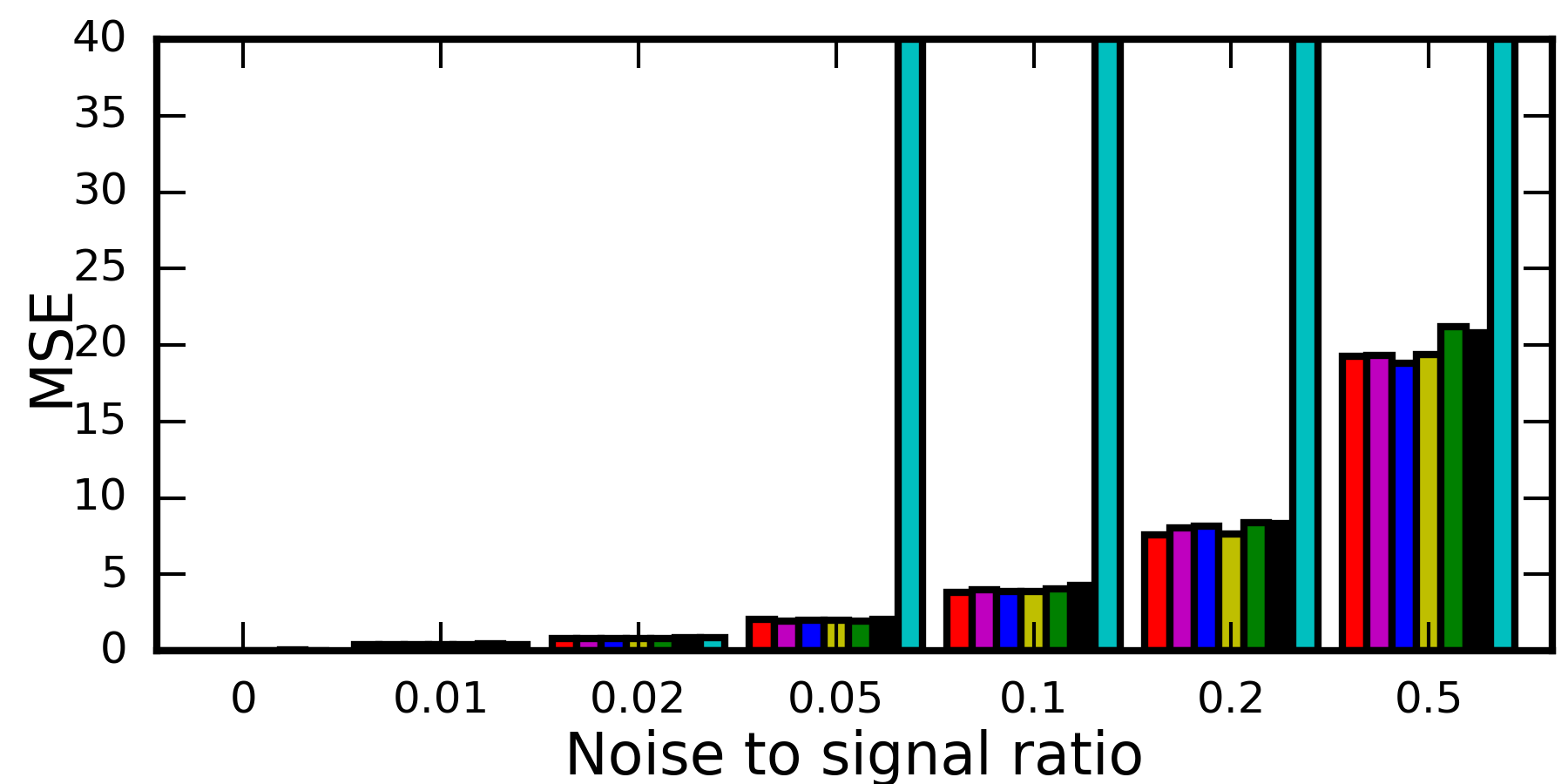}
					\captionsetup{width=0.9\columnwidth}
					\caption{NMF} 
					\label{mse_nmf_noise_test}
				\end{subfigure} %
				\begin{subfigure}{0.45 \columnwidth}
					\includegraphics[width=\columnwidth]{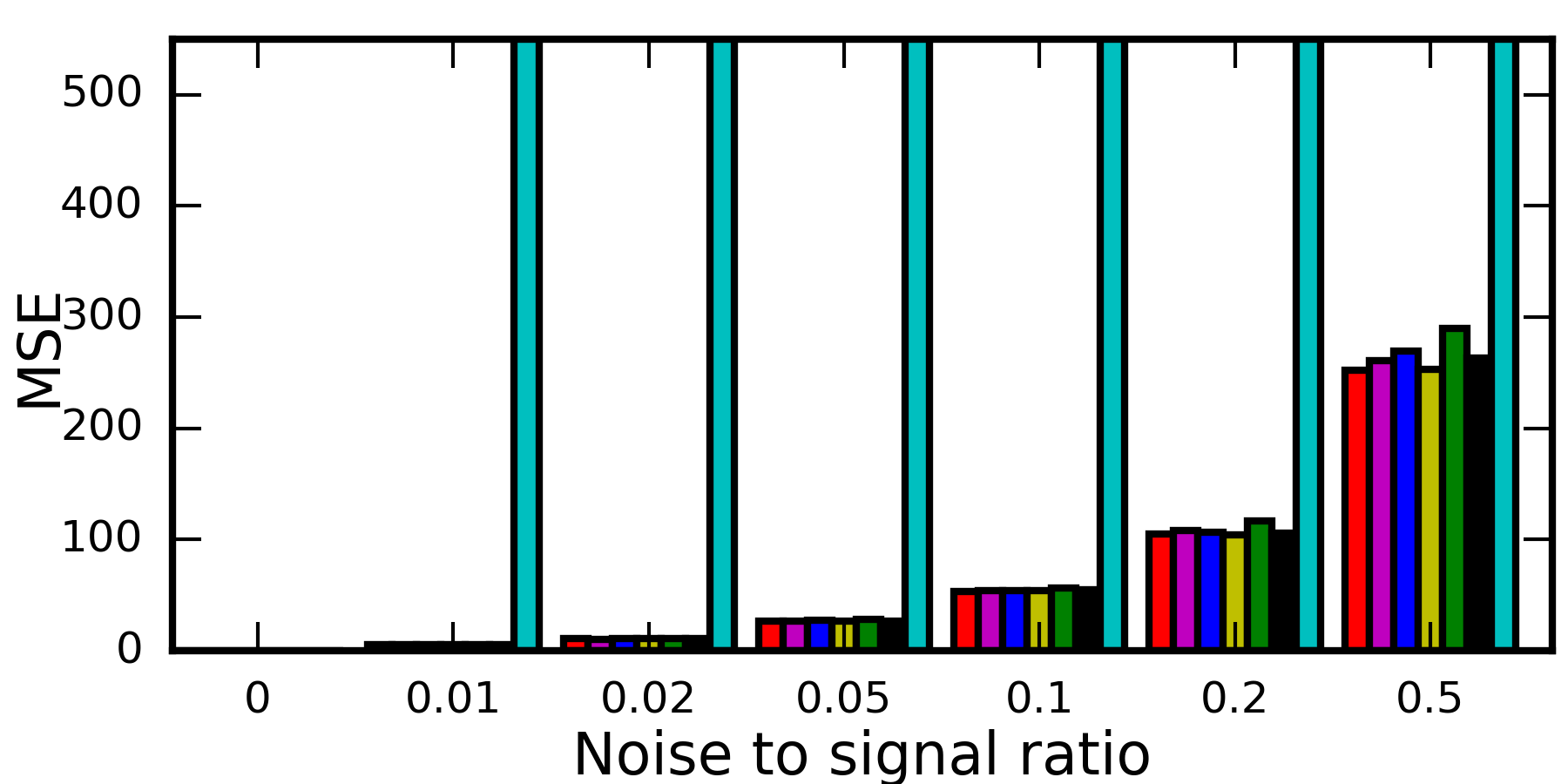}
					\captionsetup{width=0.9\columnwidth}
					\caption{NMTF} 
					\label{mse_nmtf_noise_test}
				\end{subfigure}
				\captionsetup{width=1\columnwidth}
				\caption{Noise test performances, measured by average predictive performance on test set (mean square error) for different noise-to-signal ratios.}
				\label{noise_test_ard}
			\end{figure*}

		\subsection{Most common dimensionalities cross-validation}
			We used nested cross-validation in the main paper to choose the dimensionality $K$ for NMF, and $K,L$ for NMTF. We give the most common dimensionalities for each of the four drug sensitivity datasets from this procedure in Table \ref{dimensionalities} below. This is used for the sparsity tests.
			Note that the best dimensionality is roughly the same for NMF and NMTF. VB and Gibbs have the highest values, because they overfit less when given more factors and can therefore leverage more of them. The CCLE $IC_{50}$ dataset has dimensionality 1 for all methods, indicating that no sensible predictions can be made other than a weighted row and column average. 
		
			\begin{table*}[h]
				\captionsetup{width=0.725\columnwidth}
				\caption{Most common dimensionalities ($K$ for NMF, $K,L$ for NMTF) of the inference methods on the four drug sensitivity datasets.} \label{dimensionalities}
				\centering
				\begin{tabular}{lcccc}
					\toprule
					Method \hspace{25pt} & GDSC $IC_{50}$ & CTRP $EC_{50}$ & CCLE $IC{50}$ & CCLE $EC{50}$ \\
					\midrule
					NMF VB & 7 & 6 & 5 & 1 \\
					NMF Gibbs & 8 & 7 & 5 & 1 \\
					NMF ICM & 5 & 4 & 4 & 1 \\
					NMF NP & 6 & 3 & 1 & 1 \\
					NMTF VB & 5,5 & 9,9 & 7,7 & 1,1 \\
					NMTF Gibbs & 10,10 & 8,8 & 7,7 & 1,1 \\
					NMTF ICM & 6,6 & 6,6 & 4,4 & 1,1 \\
					NMTF NP & 6,6 & 4,4 & 1,1 & 1,1 \\
					\bottomrule
				\end{tabular}
			\end{table*}
		
		\subsection{Sparsity test on other datasets}
\begin{figure*}
				\centering
				\begin{subfigure}[t]{\columnwidth}
					\hspace{90pt}
					\includegraphics[width=0.6\columnwidth]{legend.png}
					\vspace{3pt}
				\end{subfigure}
				\begin{subfigure}{0.24 \columnwidth}
					\includegraphics[width=\columnwidth]{gdsc_mse_nmf_sparsity.png}
					\captionsetup{width=0.9\columnwidth}
					\caption{GDSC, NMF} 
					\label{mse_nmf_sparsity_test_gdsc}
				\end{subfigure} %
				\begin{subfigure}{0.24 \columnwidth}
					\includegraphics[width=\columnwidth]{ctrp_mse_nmf_sparsity.png}
					\captionsetup{width=0.9\columnwidth}
					\caption{CTRP, NMF} 
					\label{mse_nmf_sparsity_test_ctrp}
					\vspace{5pt}
				\end{subfigure} %
				\begin{subfigure}{0.24 \columnwidth}
					\includegraphics[width=\columnwidth]{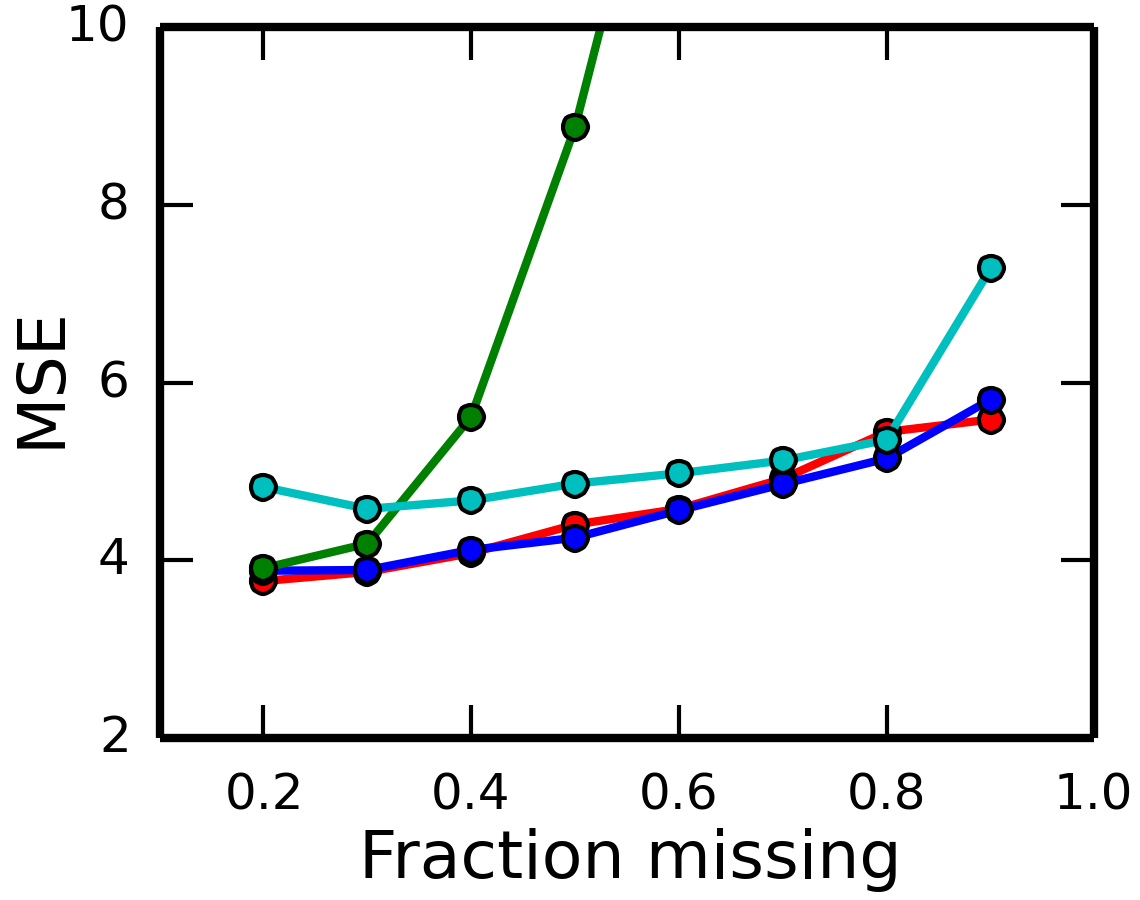}
					\captionsetup{width=0.9\columnwidth}
					\caption{CCLE $IC_{50}$, NMF} 
					\label{mse_nmf_sparsity_test_ccle_ic}
					\vspace{5pt}
				\end{subfigure} %
				\begin{subfigure}{0.24 \columnwidth}
					\includegraphics[width=\columnwidth]{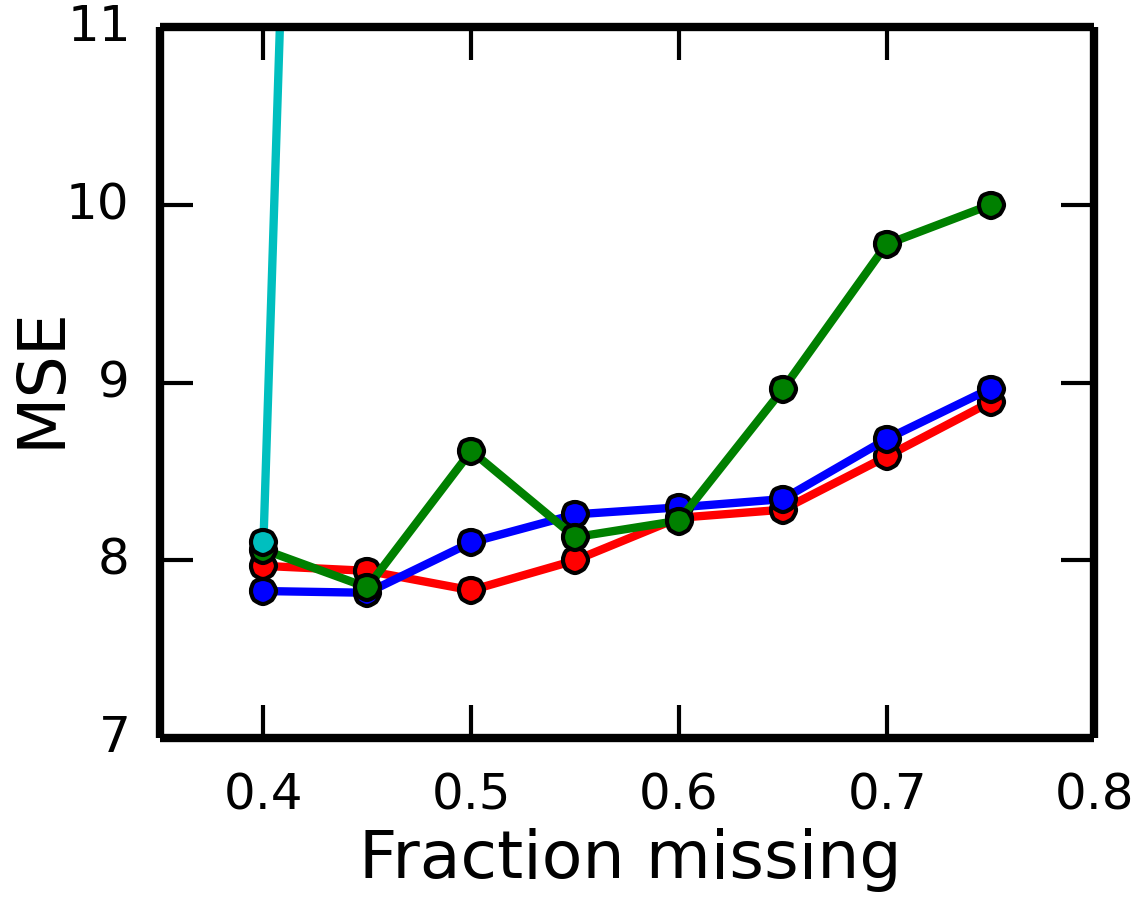}
					\captionsetup{width=0.9\columnwidth}
					\caption{CCLE $EC_{50}$, NMF} 
					\label{mse_nmf_sparsity_test_ccle_ec}
					\vspace{5pt}
				\end{subfigure} %
				\begin{subfigure}{0.24 \columnwidth}
					\includegraphics[width=\columnwidth]{gdsc_mse_nmtf_sparsity.png}
					\captionsetup{width=0.9\columnwidth}
					\caption{GDSC, NMTF} 
					\label{mse_nmtf_sparsity_test_gdsc}
				\end{subfigure}
				\begin{subfigure}{0.24 \columnwidth}
					\includegraphics[width=\columnwidth]{ctrp_mse_nmtf_sparsity.png}
					\captionsetup{width=0.9\columnwidth}
					\caption{CTRP, NMTF} 
					\label{mse_nmtf_sparsity_test_ctrp}
				\end{subfigure}
				\begin{subfigure}{0.24 \columnwidth}
					\includegraphics[width=\columnwidth]{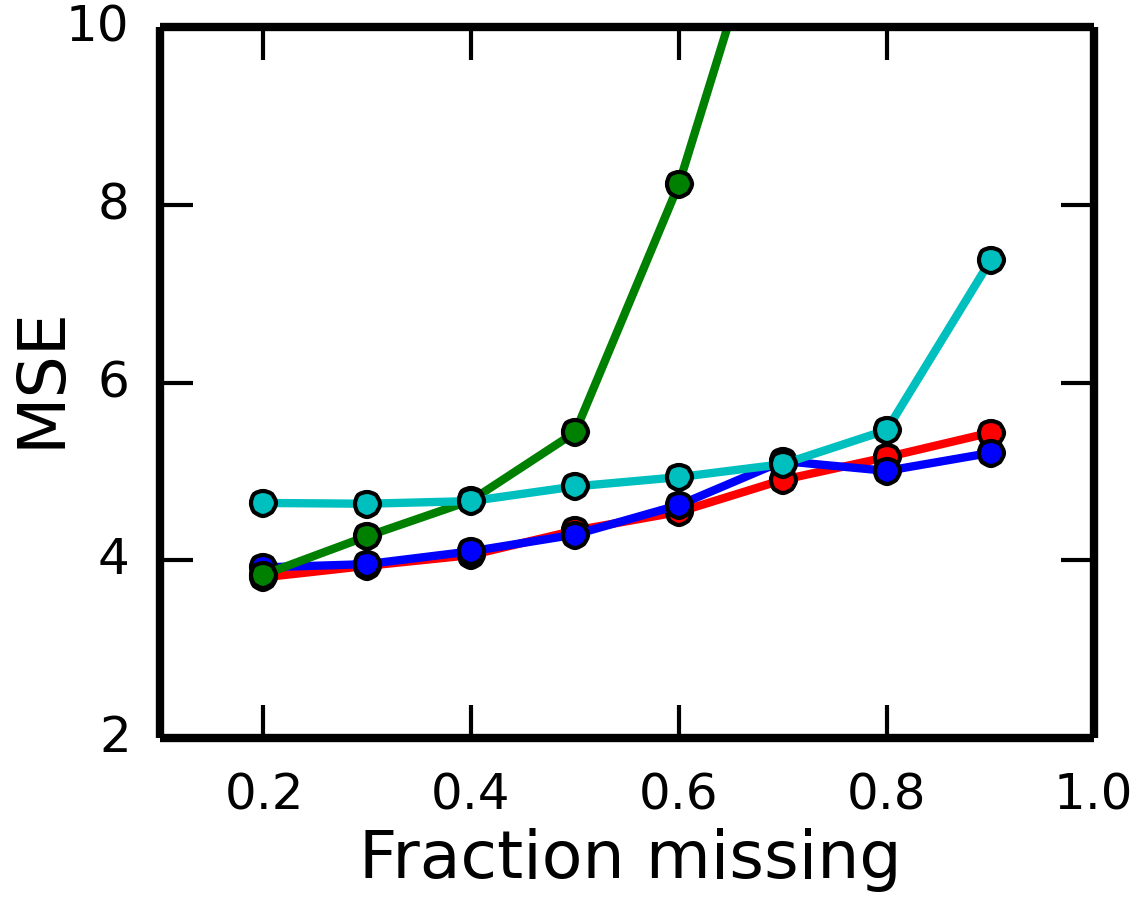}
					\captionsetup{width=0.9\columnwidth}
					\caption{CCLE $IC_{50}$, NMF} 
					\label{mse_nmtf_sparsity_test_ccle_ic}
					\vspace{5pt}
				\end{subfigure} %
				\begin{subfigure}{0.24 \columnwidth}
					\includegraphics[width=\columnwidth]{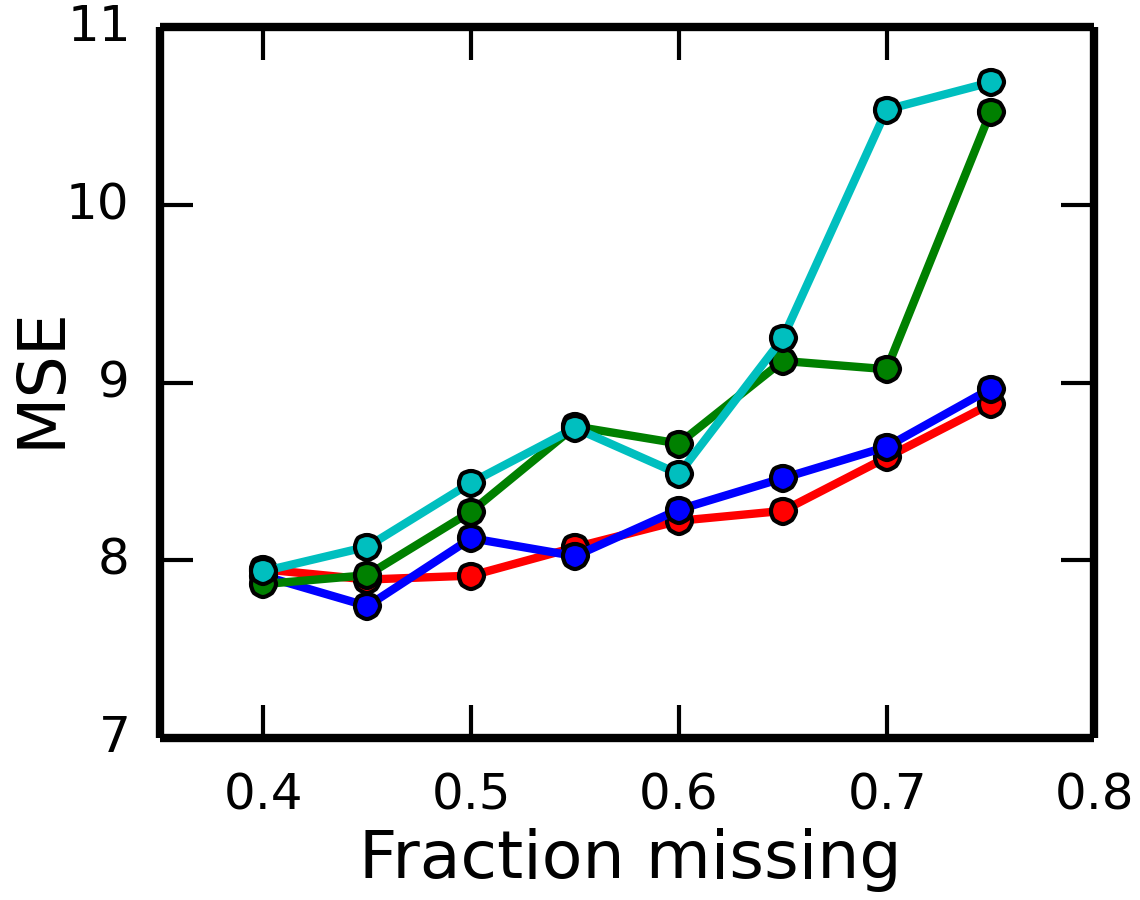}
					\captionsetup{width=0.9\columnwidth}
					\caption{CCLE $EC_{50}$, NMF} 
					\label{mse_nmtf_sparsity_test_ccle_ec}
					\vspace{5pt}
				\end{subfigure} %
				\captionsetup{width=1\columnwidth}
				\caption{Sparsity test performances on all four drug sensitivity datasets, measured by average predictive performance on test set (mean square error) for different sparsity levels. }
				\label{sparsity_results}
			\end{figure*}

			We conducted the same sparsity test as in the main paper on the remaining two drug sensitivity dataset. Results for both NMF and NMTF on all four datasets are given in Figure \ref{sparsity_results}. As with the other two datasets, on CCLE $IC_{50}$ and $EC_{50}$ we see that the VB and Gibbs versions are much more robust to sparsity than the ICM and NP methods. 
						
		\subsection{Sparsity test with ARD}
\begin{figure*}
				\centering
				\begin{subfigure}[t]{\columnwidth}
					\hspace{70pt}
					\includegraphics[width=0.7\columnwidth]{legend_ard.png}
					\vspace{3pt}
				\end{subfigure}
				\begin{subfigure}{0.24 \columnwidth}
					\includegraphics[width=\columnwidth]{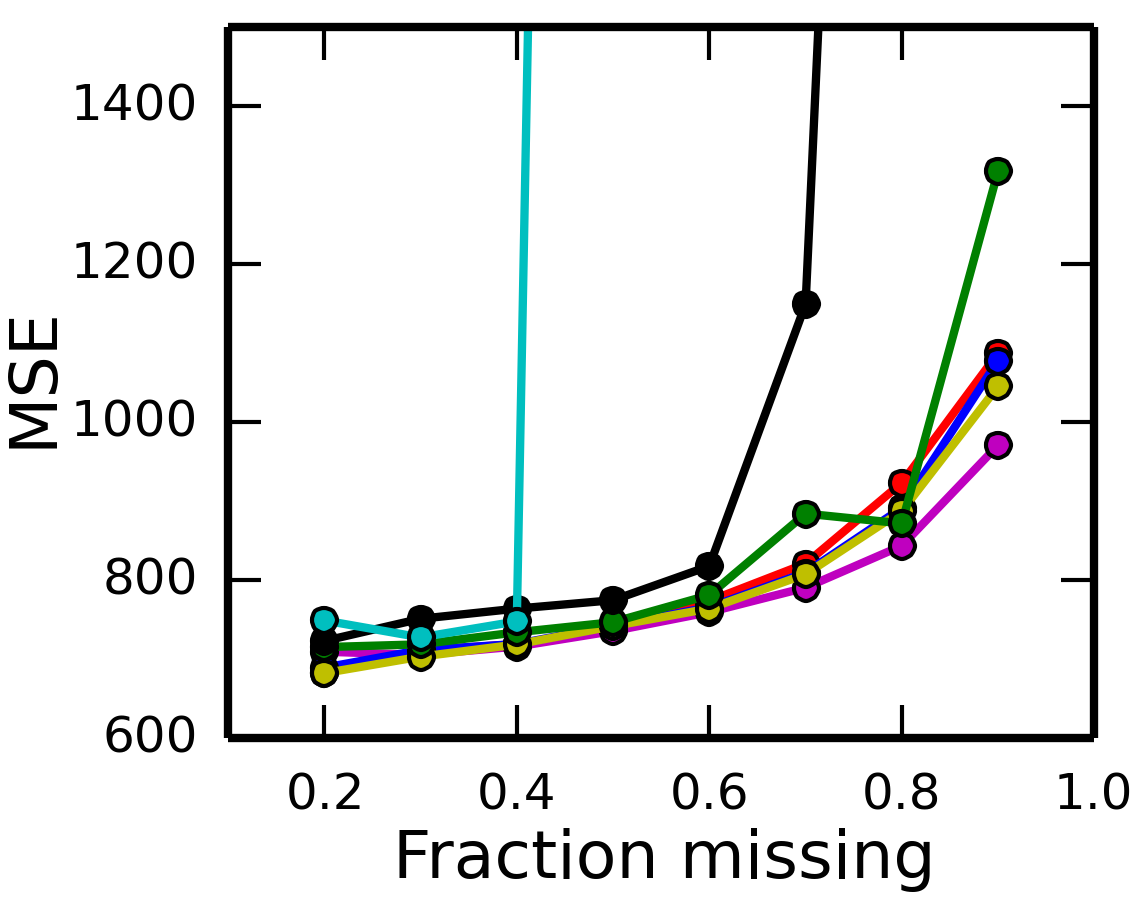}
					\captionsetup{width=0.9\columnwidth}
					\caption{GDSC, NMF} 
				\end{subfigure} %
				\begin{subfigure}{0.24 \columnwidth}
					\includegraphics[width=\columnwidth]{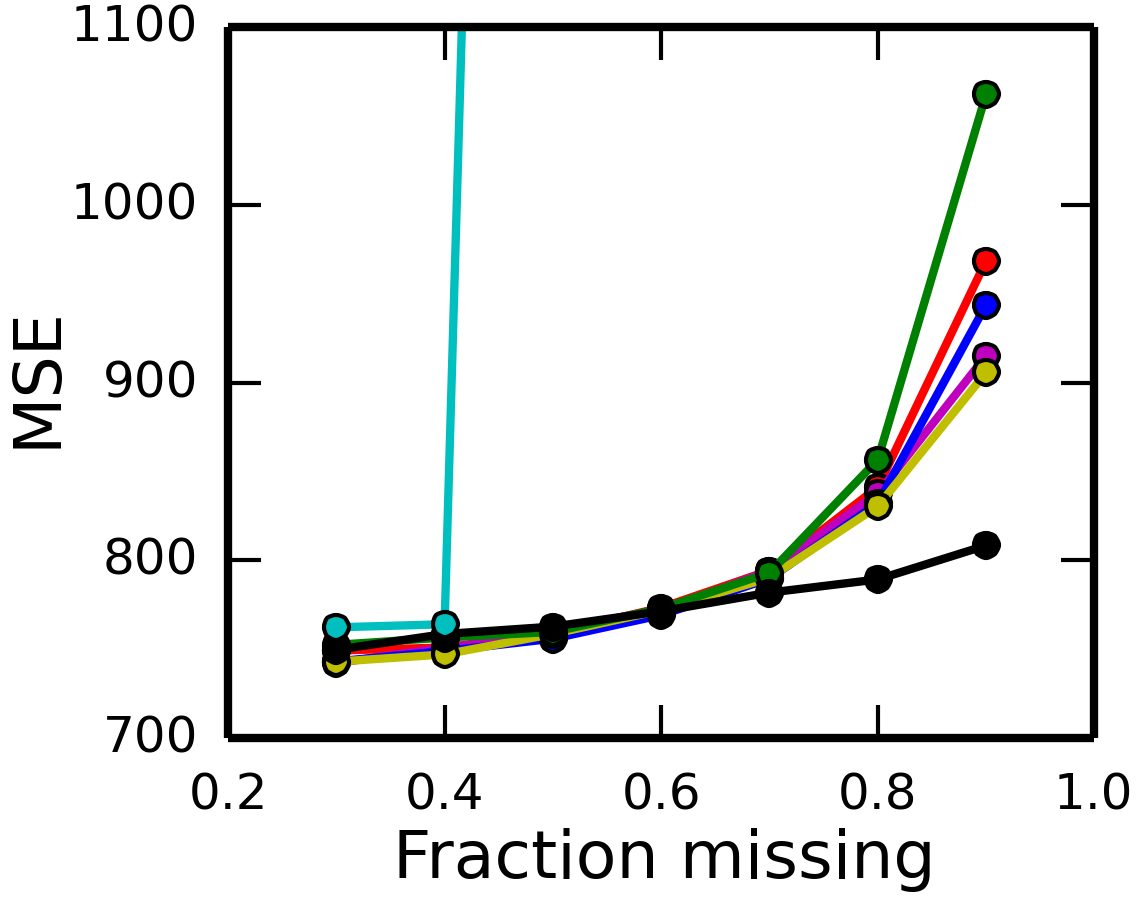}
					\captionsetup{width=0.9\columnwidth}
					\caption{CTRP, NMF} 
					\vspace{5pt}
				\end{subfigure} %
				\begin{subfigure}{0.24 \columnwidth}
					\includegraphics[width=\columnwidth]{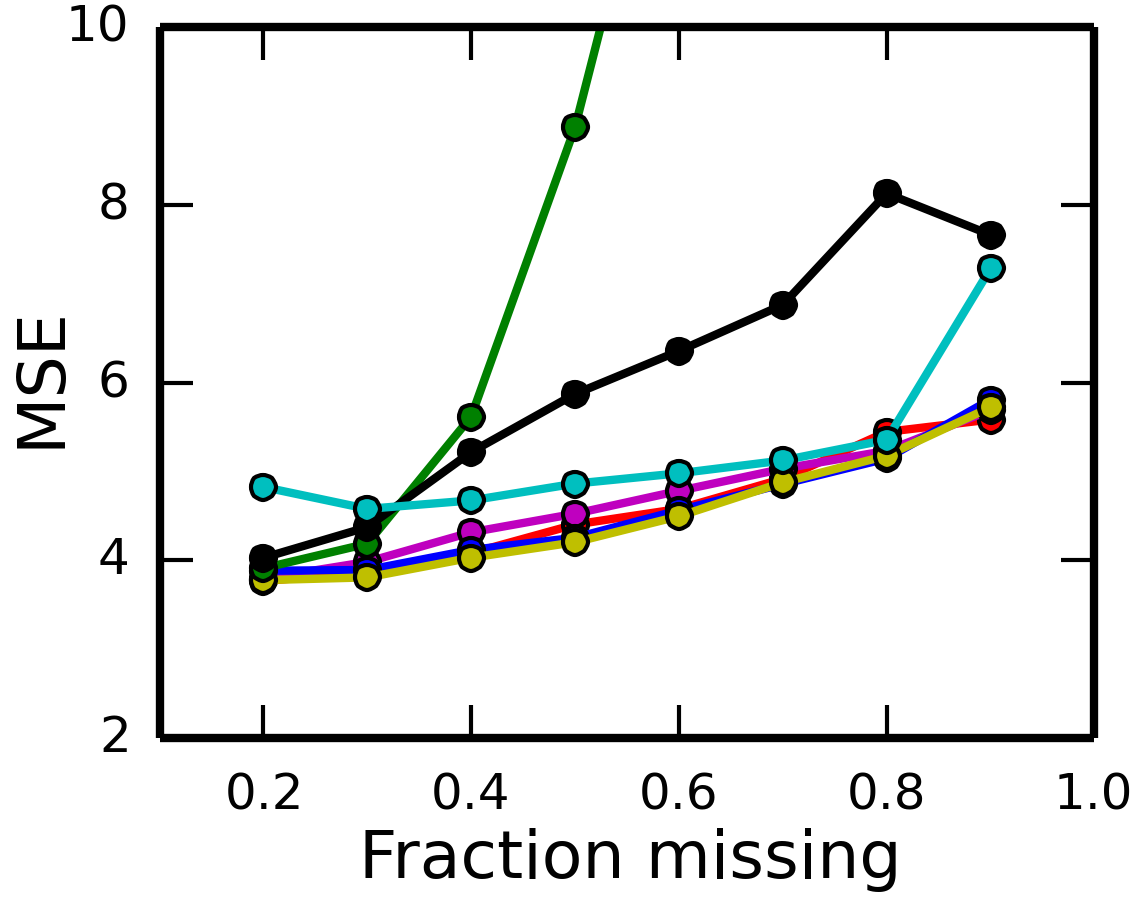}
					\captionsetup{width=0.9\columnwidth}
					\caption{CCLE $IC_{50}$, NMF} 
					\vspace{5pt}
				\end{subfigure} %
				\begin{subfigure}{0.24 \columnwidth}
					\includegraphics[width=\columnwidth]{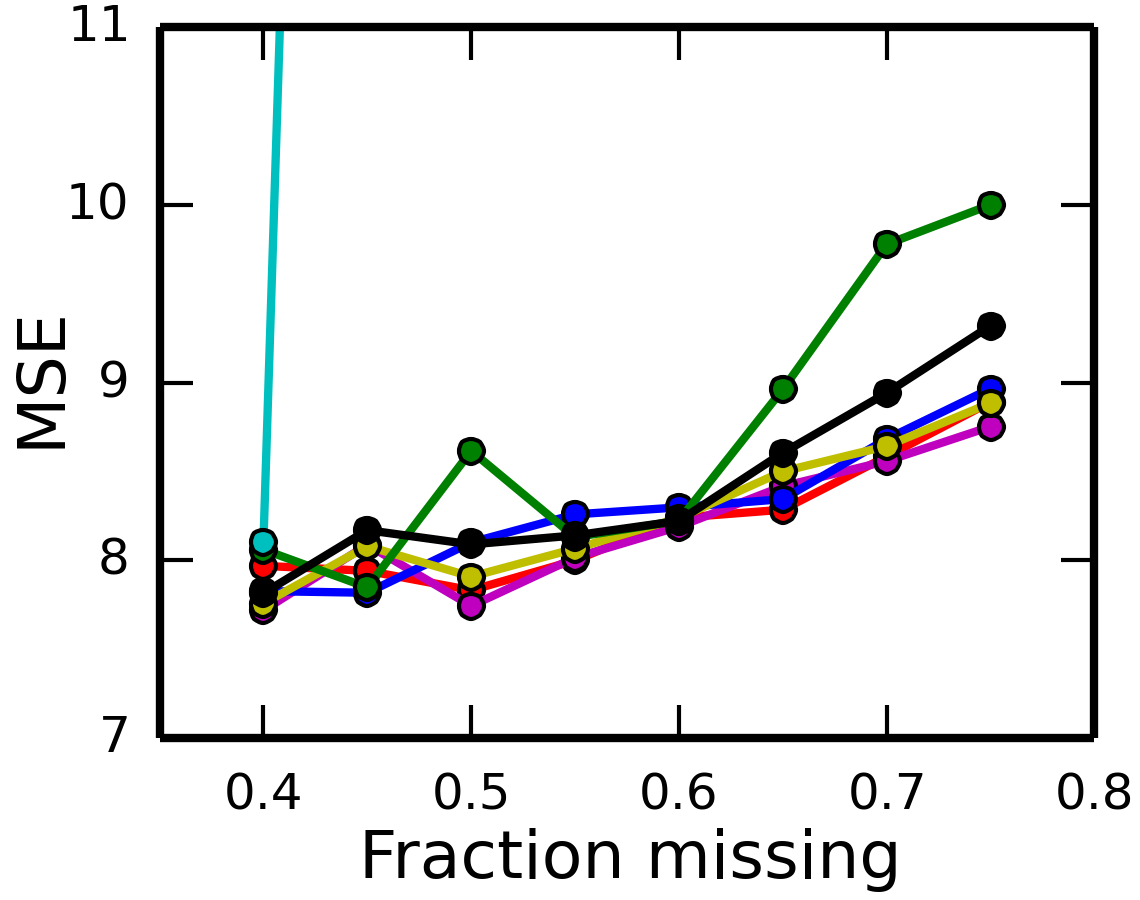}
					\captionsetup{width=0.9\columnwidth}
					\caption{CCLE $EC_{50}$, NMF} 
					\vspace{5pt}
				\end{subfigure} %
				\begin{subfigure}{0.24 \columnwidth}
					\includegraphics[width=\columnwidth]{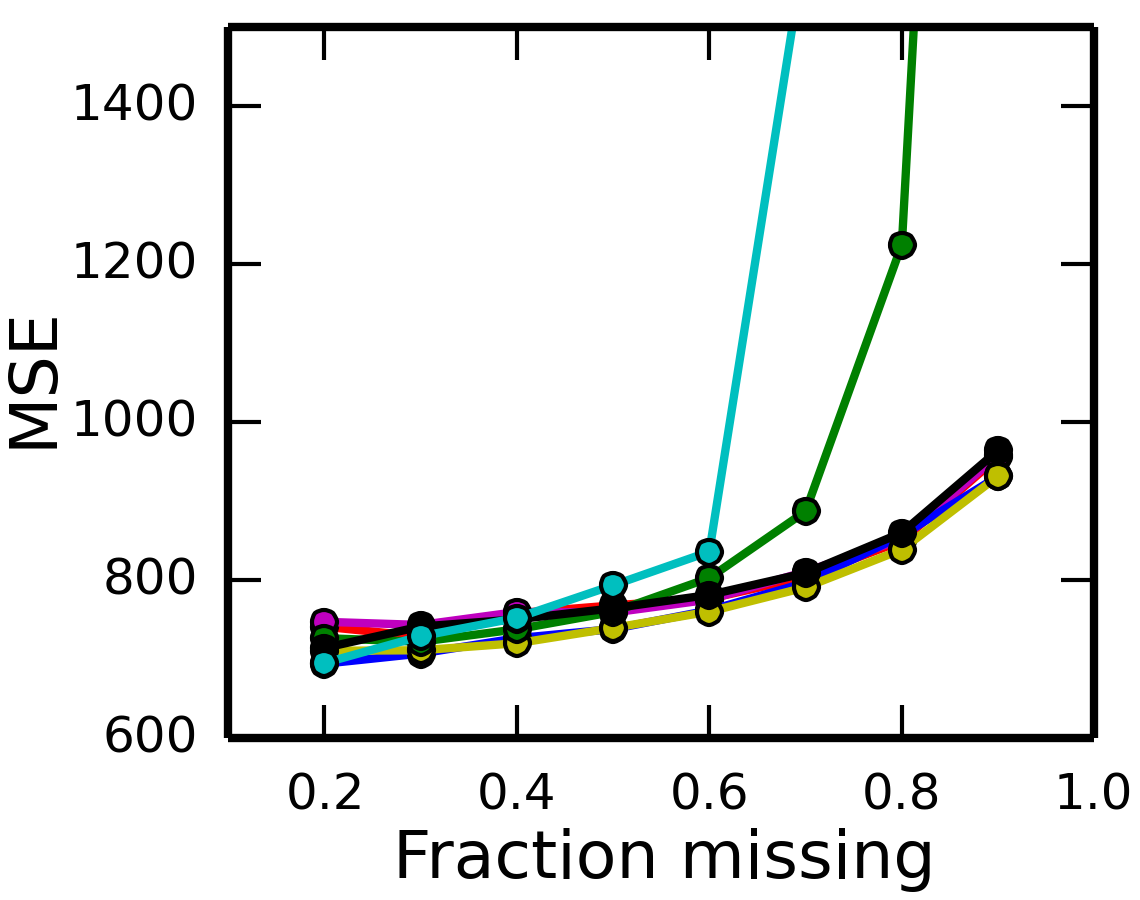}
					\captionsetup{width=0.9\columnwidth}
					\caption{GDSC, NMTF} 
				\end{subfigure}
				\begin{subfigure}{0.24 \columnwidth}
					\includegraphics[width=\columnwidth]{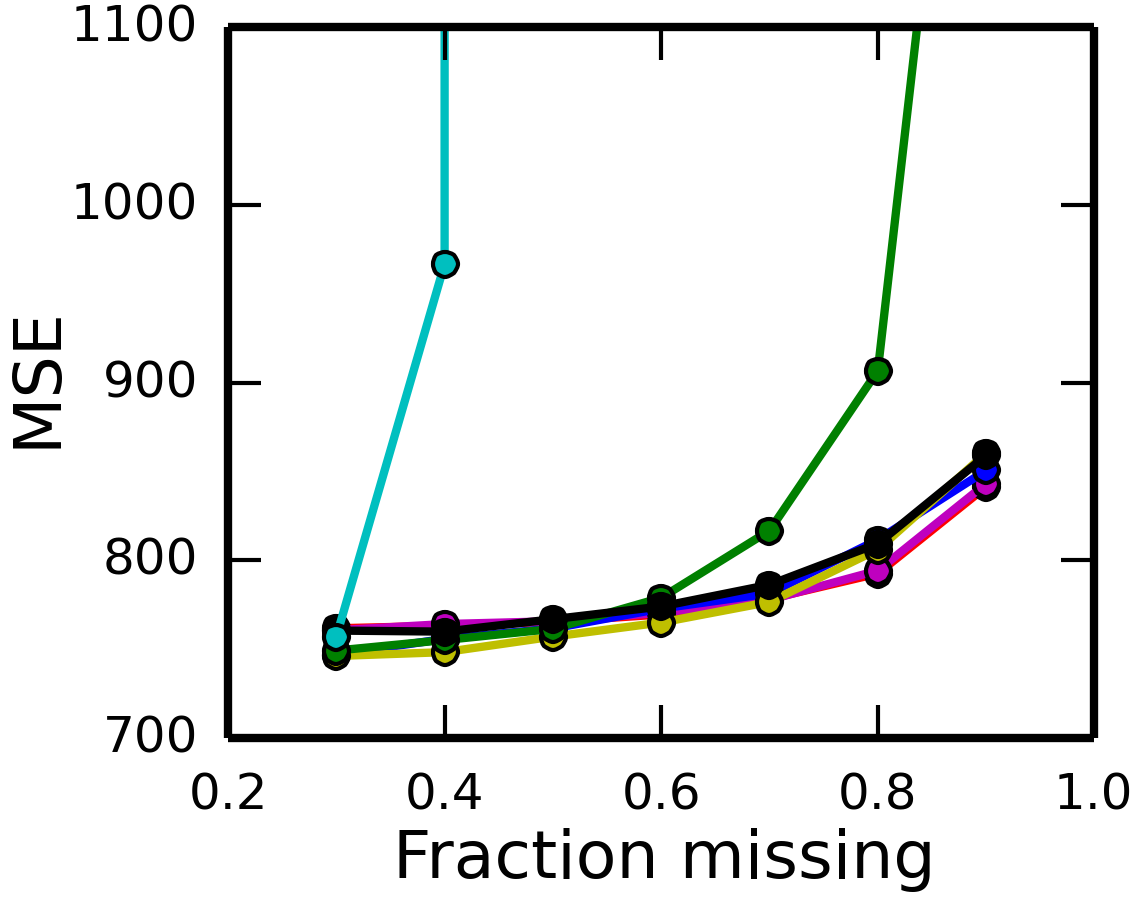}
					\captionsetup{width=0.9\columnwidth}
					\caption{CTRP, NMTF} 
				\end{subfigure}
				\begin{subfigure}{0.24 \columnwidth}
					\includegraphics[width=\columnwidth]{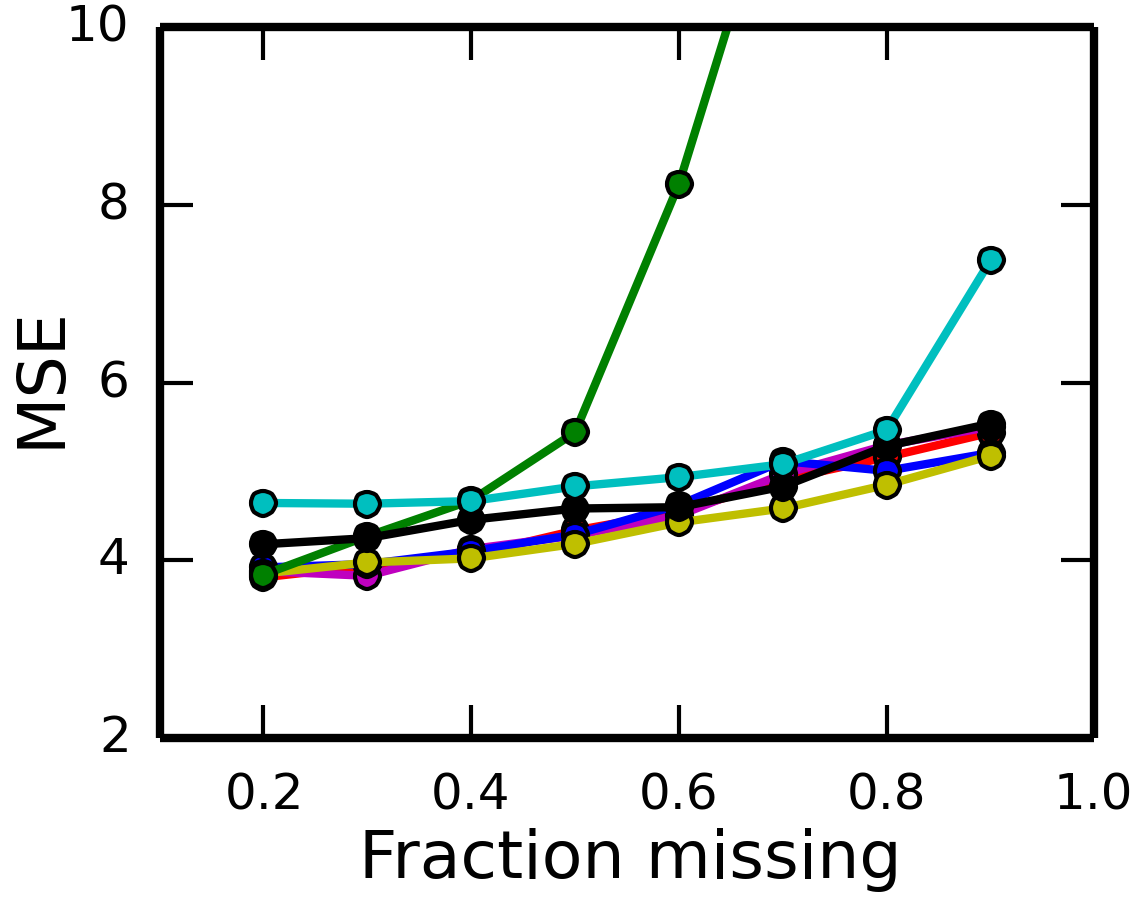}
					\captionsetup{width=0.9\columnwidth}
					\caption{CCLE $IC_{50}$, NMTF} 
					\vspace{5pt}
				\end{subfigure} %
				\begin{subfigure}{0.24 \columnwidth}
					\includegraphics[width=\columnwidth]{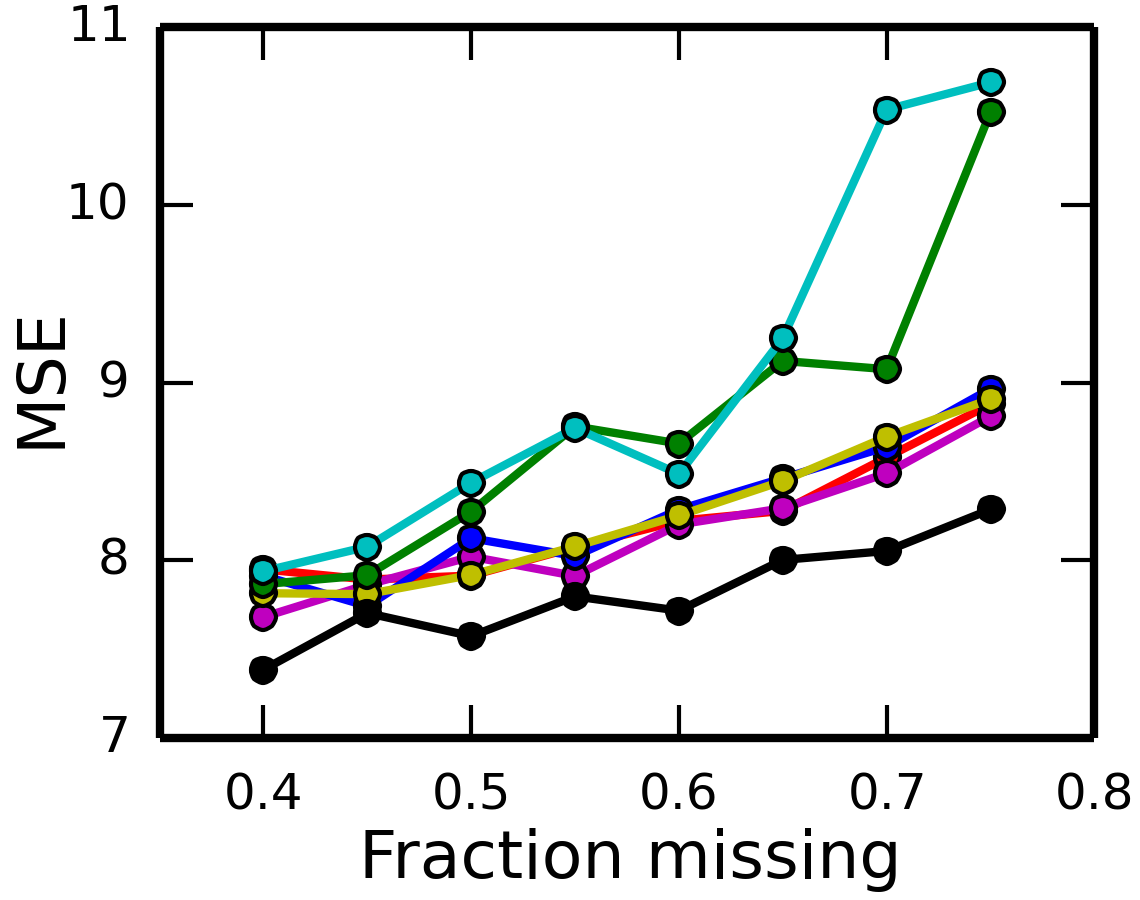}
					\captionsetup{width=0.9\columnwidth}
					\caption{CCLE $EC_{50}$, NMTF} 
					\vspace{5pt}
				\end{subfigure} %
				\captionsetup{width=1\columnwidth}
				\caption{Sparsity test performances for all methods (with and without ARD) on the four drug sensitivity datasets, measured by average predictive performance on test set (mean square error) for different sparsity levels. The top row gives the performances for NMF, and the bottom for NMTF.}
				\label{sparsity_results_ard}
			\end{figure*}

			We furthermore give the performances in the sparsity test on the four drug sensitivity datasets of the methods with ARD. We see in Figure \ref{sparsity_results_ard} that adding ARD makes no difference to the robustness to sparsity for the fully Bayesian models (VB and Gibbs), but for the ICM version it can greatly increase its robustness: notice how ICM with ARD (black line) often performs better than ICM (green line).
			
		\subsection{Model selection on other datasets}
\begin{figure*}[t]
				\centering
				\includegraphics[width=0.21\columnwidth]{gdsc_nmf_vb_model_selection.png}
				\includegraphics[width=0.21\columnwidth]{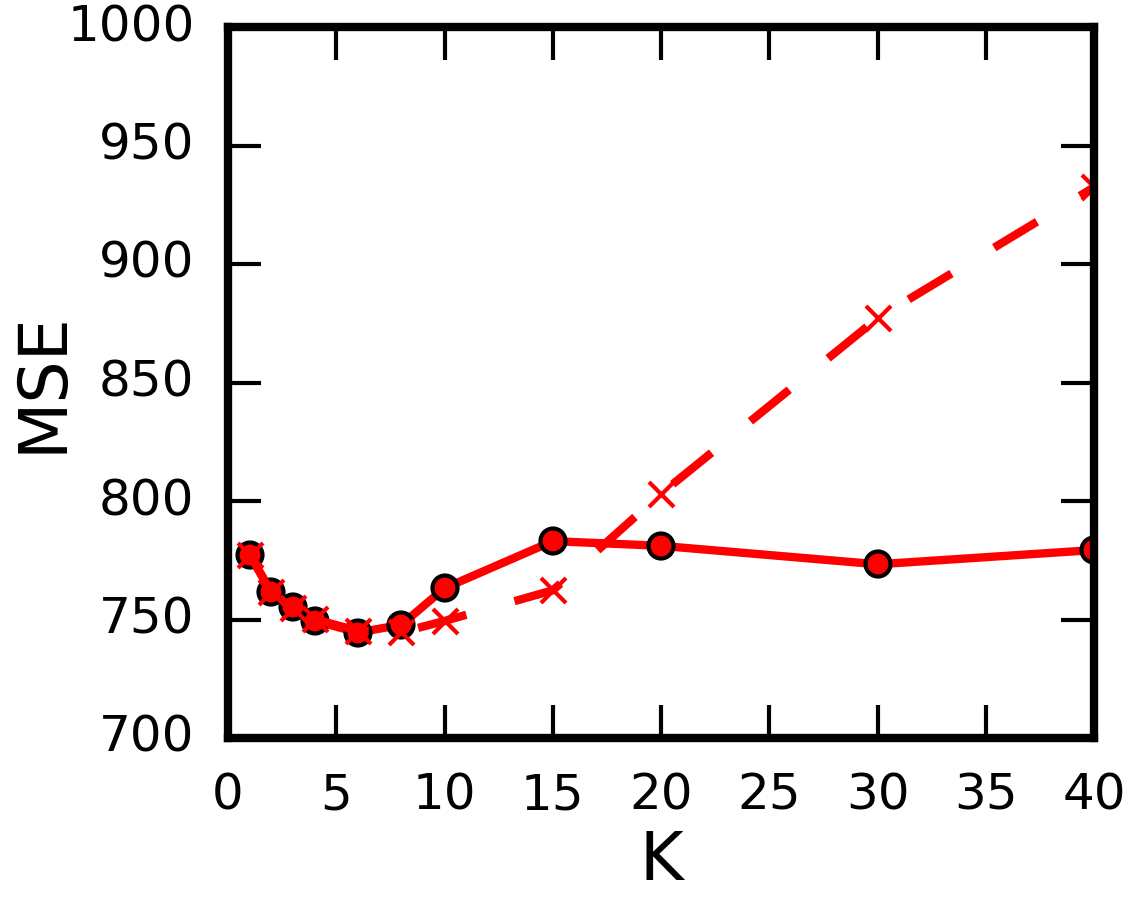}
				\includegraphics[width=0.21\columnwidth]{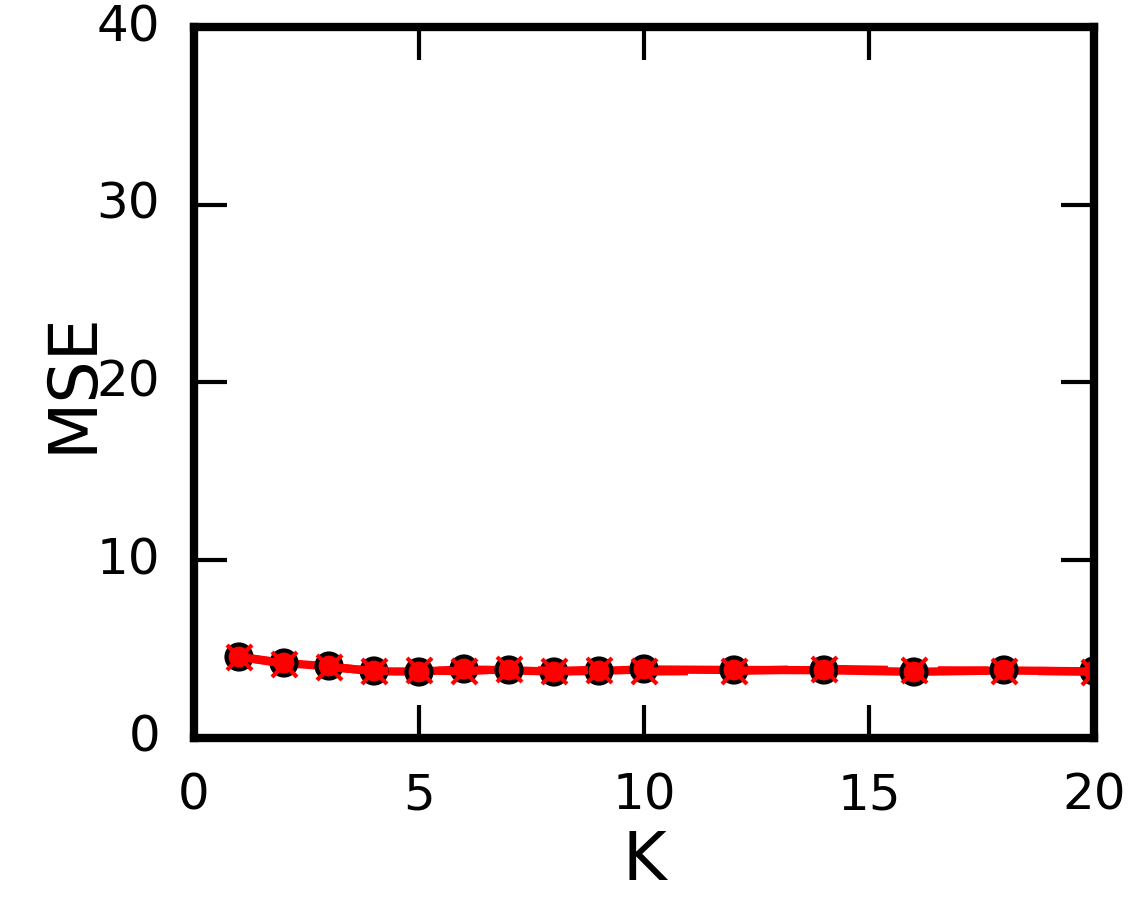}
				\includegraphics[width=0.21\columnwidth]{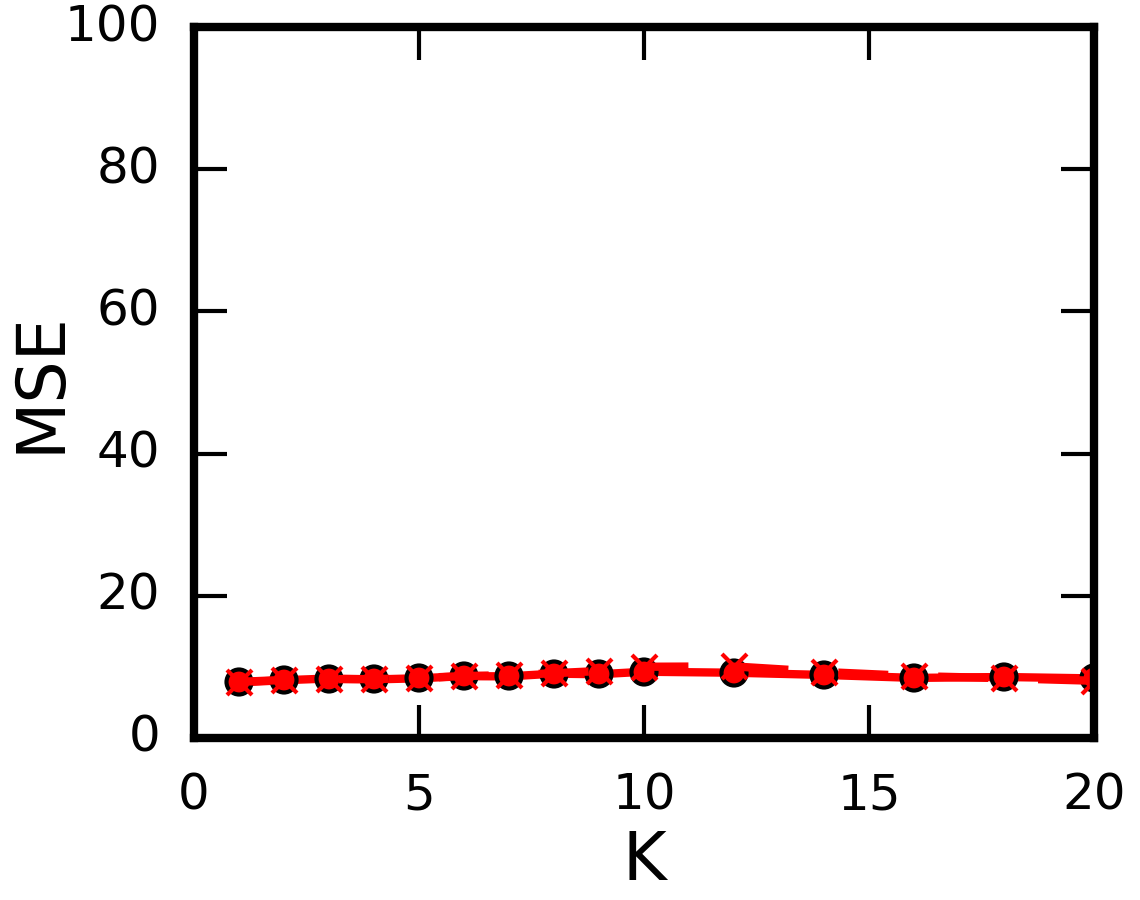}
				\includegraphics[width=0.21\columnwidth]{gdsc_nmf_gibbs_model_selection.png}
				\includegraphics[width=0.21\columnwidth]{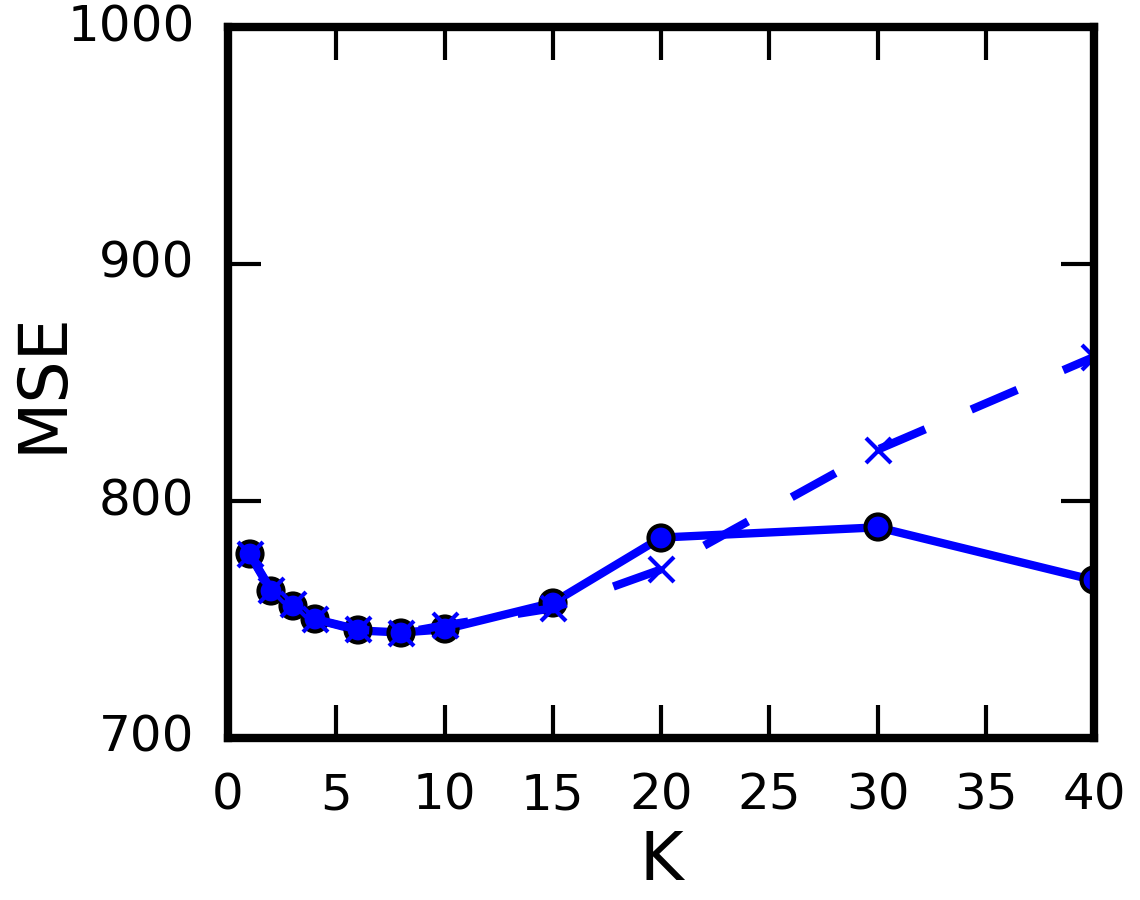}
				\includegraphics[width=0.21\columnwidth]{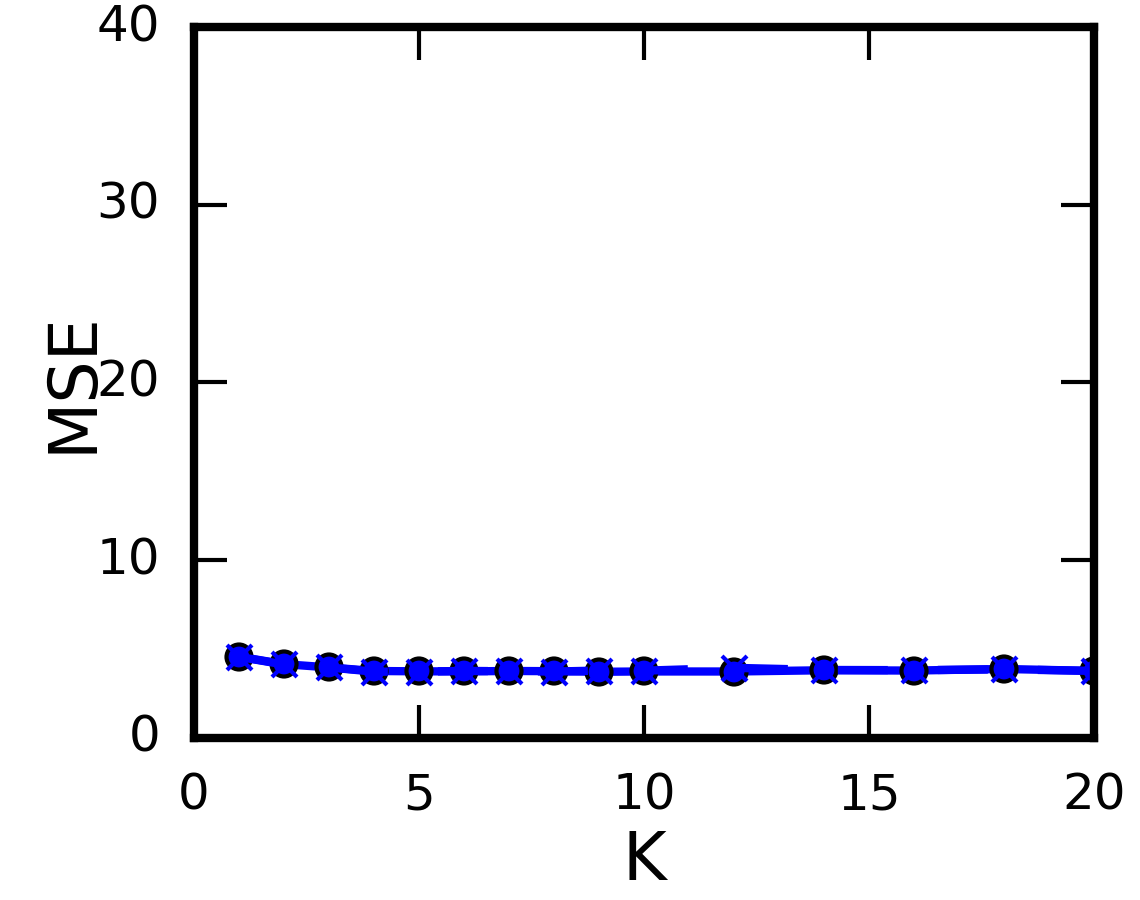}
				\includegraphics[width=0.21\columnwidth]{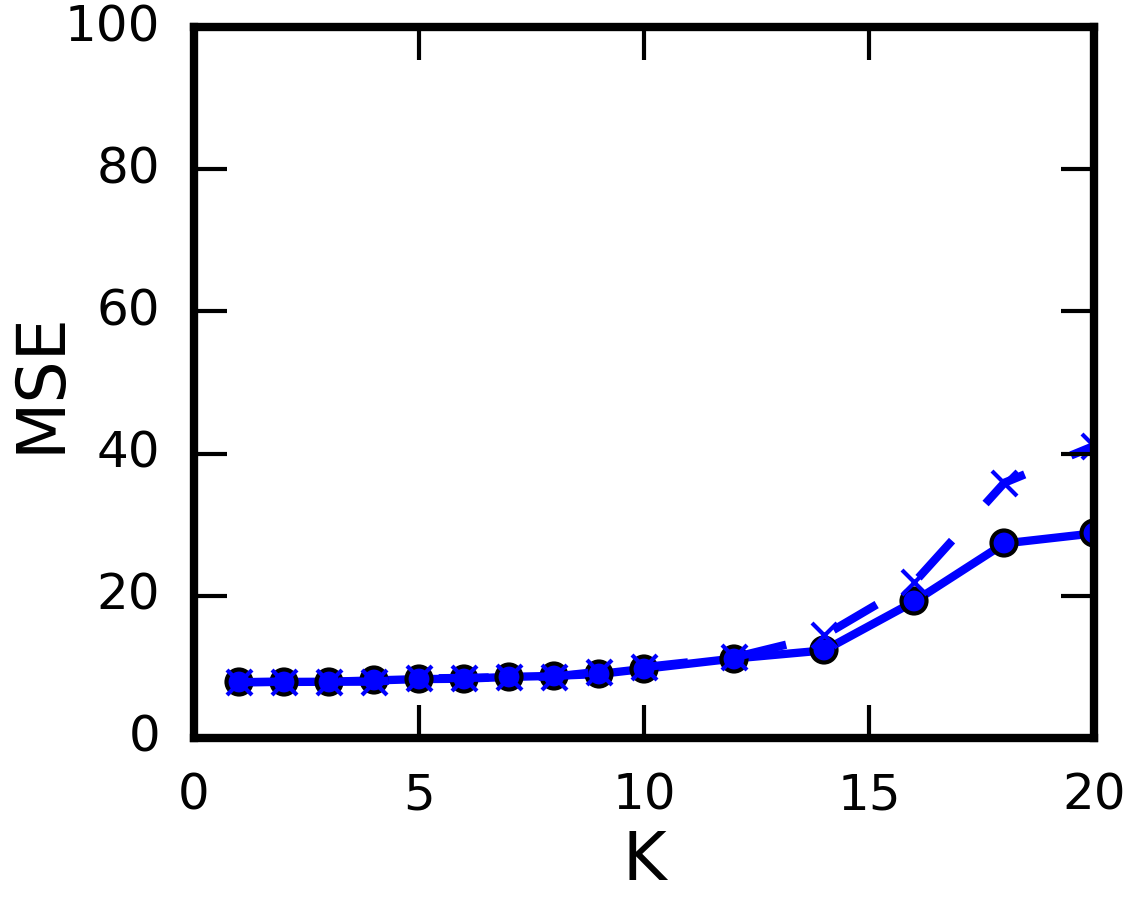}
				\begin{subfigure}{0.21 \columnwidth}
					\includegraphics[width=\columnwidth]{gdsc_nmf_icm_model_selection.png}
					\captionsetup{width=\columnwidth}
					\caption{GDSC, NMF}
				\end{subfigure}
				\begin{subfigure}{0.21 \columnwidth}
					\includegraphics[width=\columnwidth]{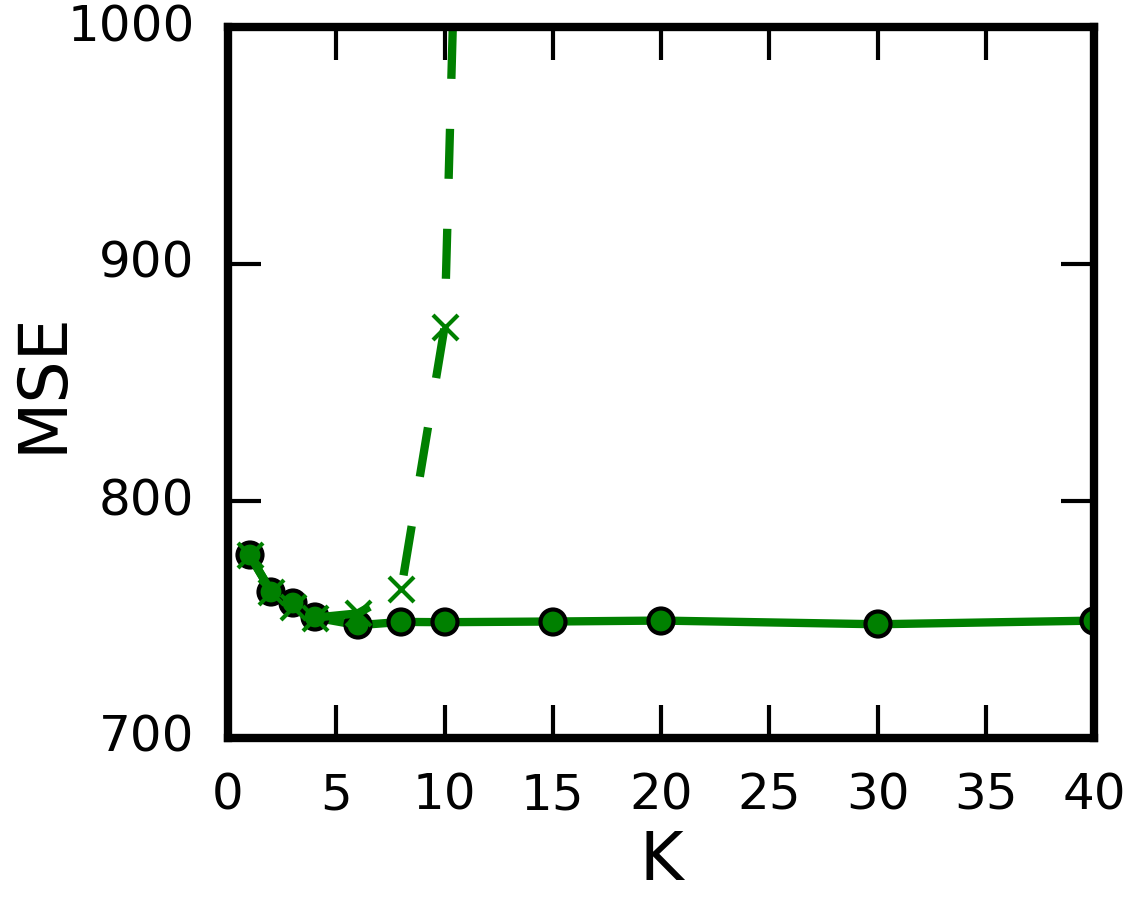}
					\captionsetup{width=\columnwidth}
					\caption{CTRP, NMF}
				\end{subfigure}
				\begin{subfigure}{0.21 \columnwidth}
					\includegraphics[width=\columnwidth]{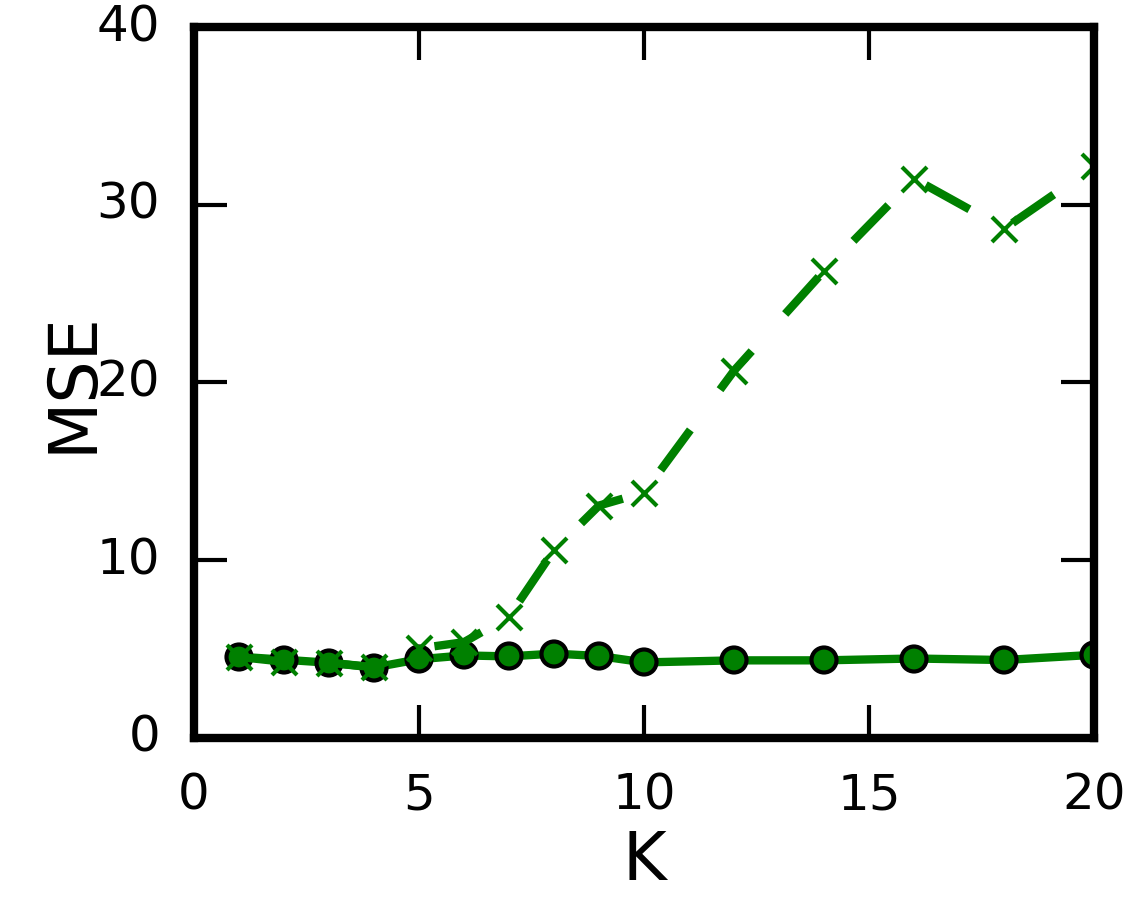}
					\captionsetup{width=\columnwidth}
					\caption{CCLE $IC_{50}$, NMF}
				\end{subfigure}
				\begin{subfigure}{0.21 \columnwidth}
					\includegraphics[width=\columnwidth]{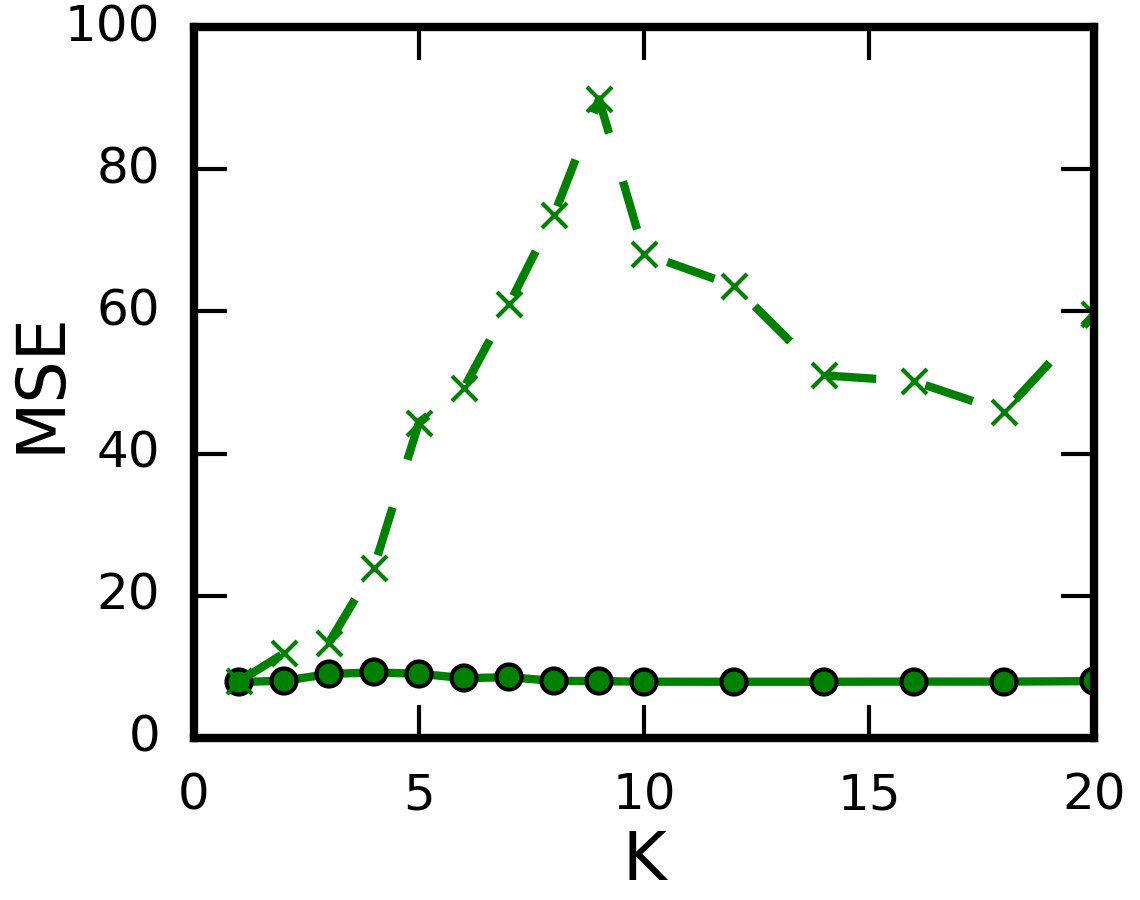}
					\captionsetup{width=\columnwidth}
					\caption{CCLE $EC_{50}$, NMF}
				\end{subfigure}
				\captionsetup{width=1\columnwidth}
				\caption{10-fold cross-validation performances of the Bayesian NMF models on the drug sensitivity datasets, where we vary the dimensionality $K$. The top row gives the performances for NMF VB, the middle row for NMF Gibbs, and the bottom row for NMF ICM. Performances for models without ARD are given by dotted lines and crosses (x), with ARD by circles (o).}
				\label{model_selection_nmf}
				\vspace{10pt}
				\includegraphics[width=0.21\columnwidth]{gdsc_nmtf_vb_model_selection.png}
				\includegraphics[width=0.21\columnwidth]{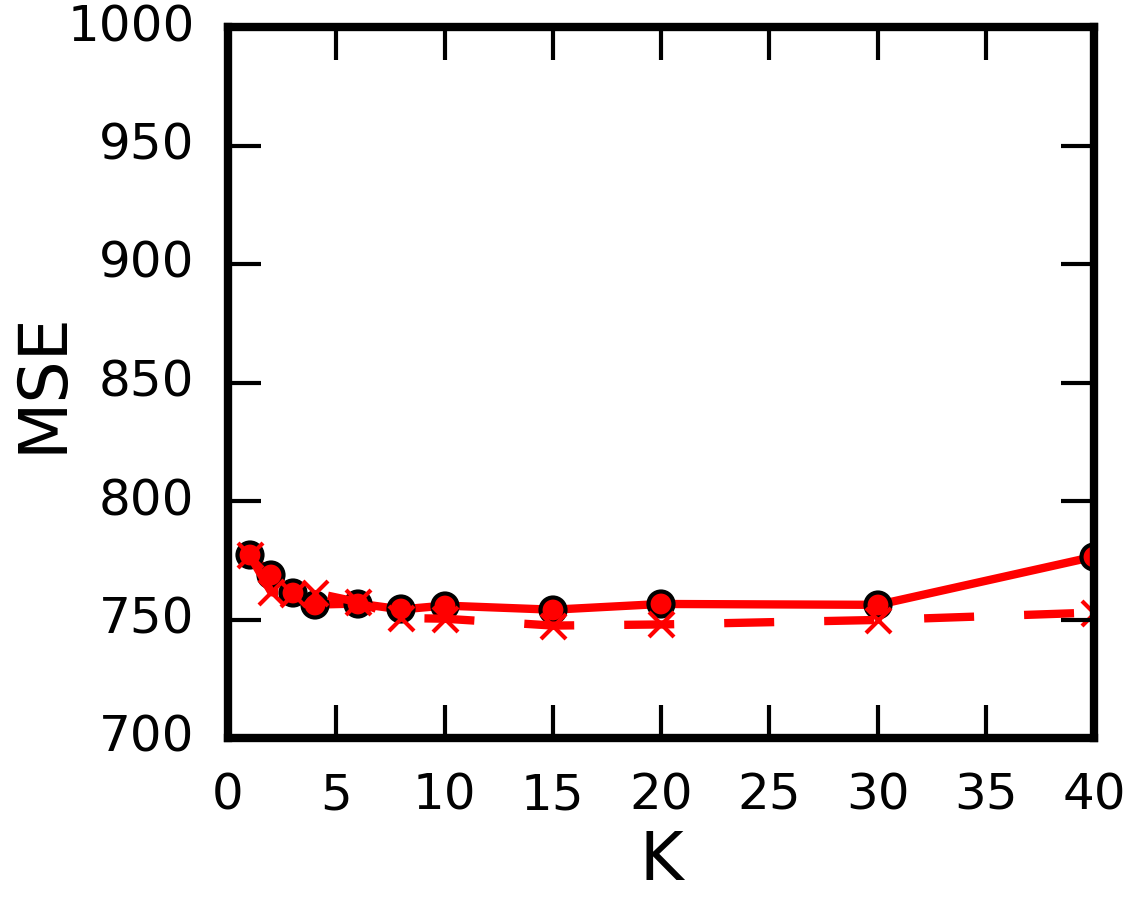}
				\includegraphics[width=0.21\columnwidth]{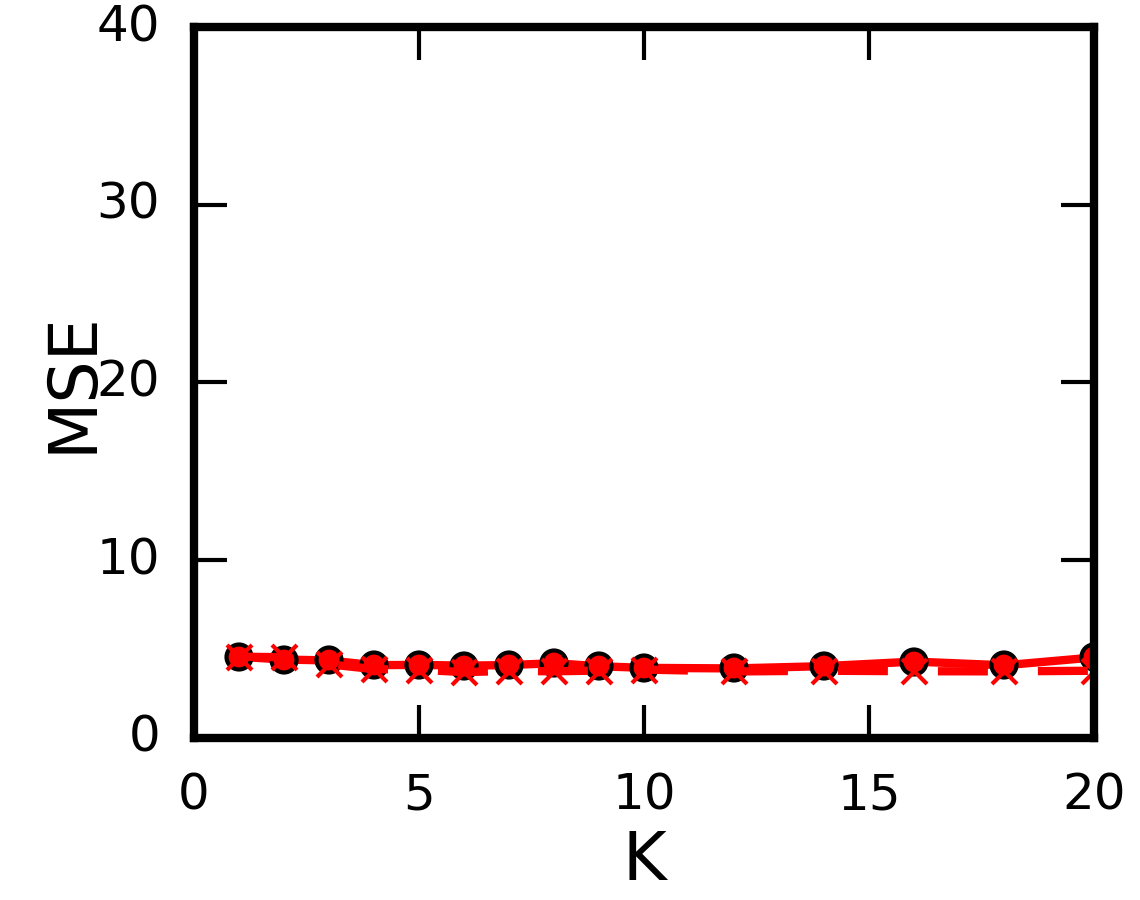}
				\includegraphics[width=0.21\columnwidth]{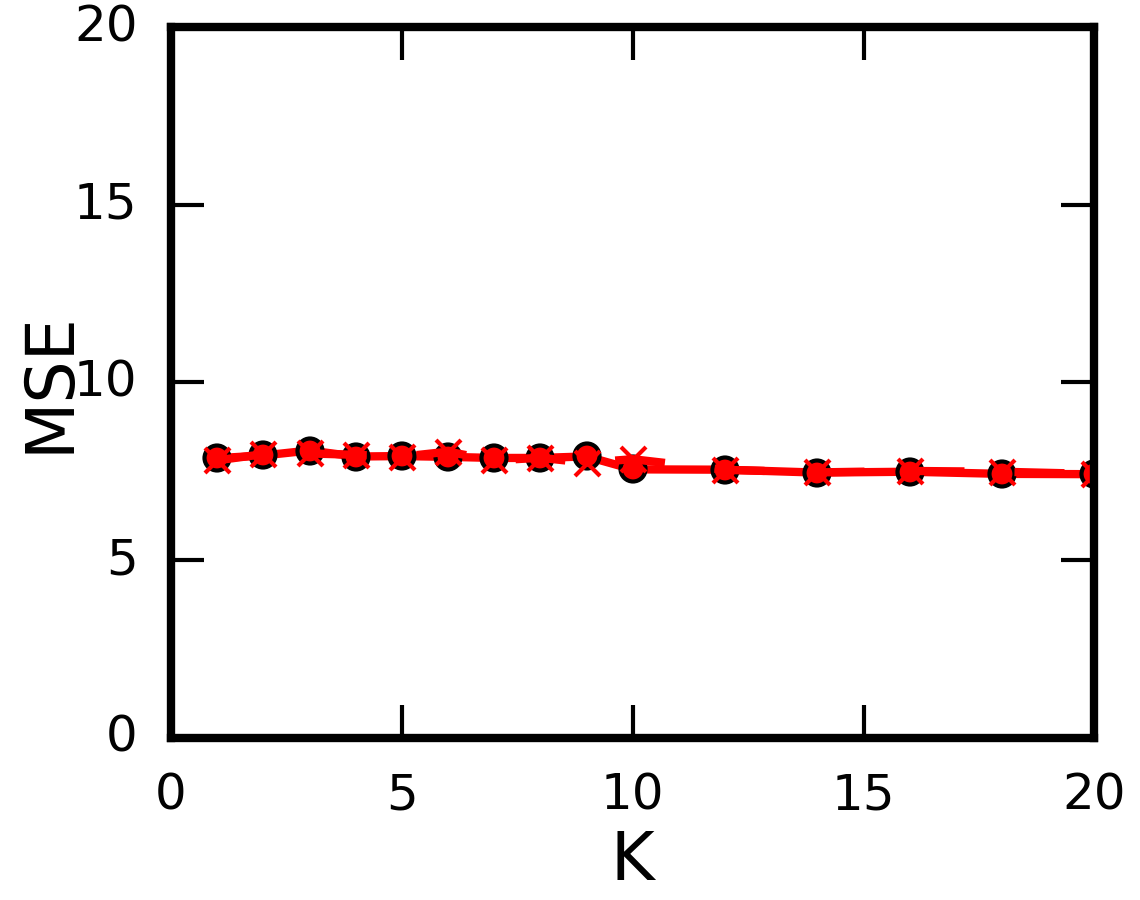}
				\includegraphics[width=0.21\columnwidth]{gdsc_nmtf_gibbs_model_selection.png}
				\includegraphics[width=0.21\columnwidth]{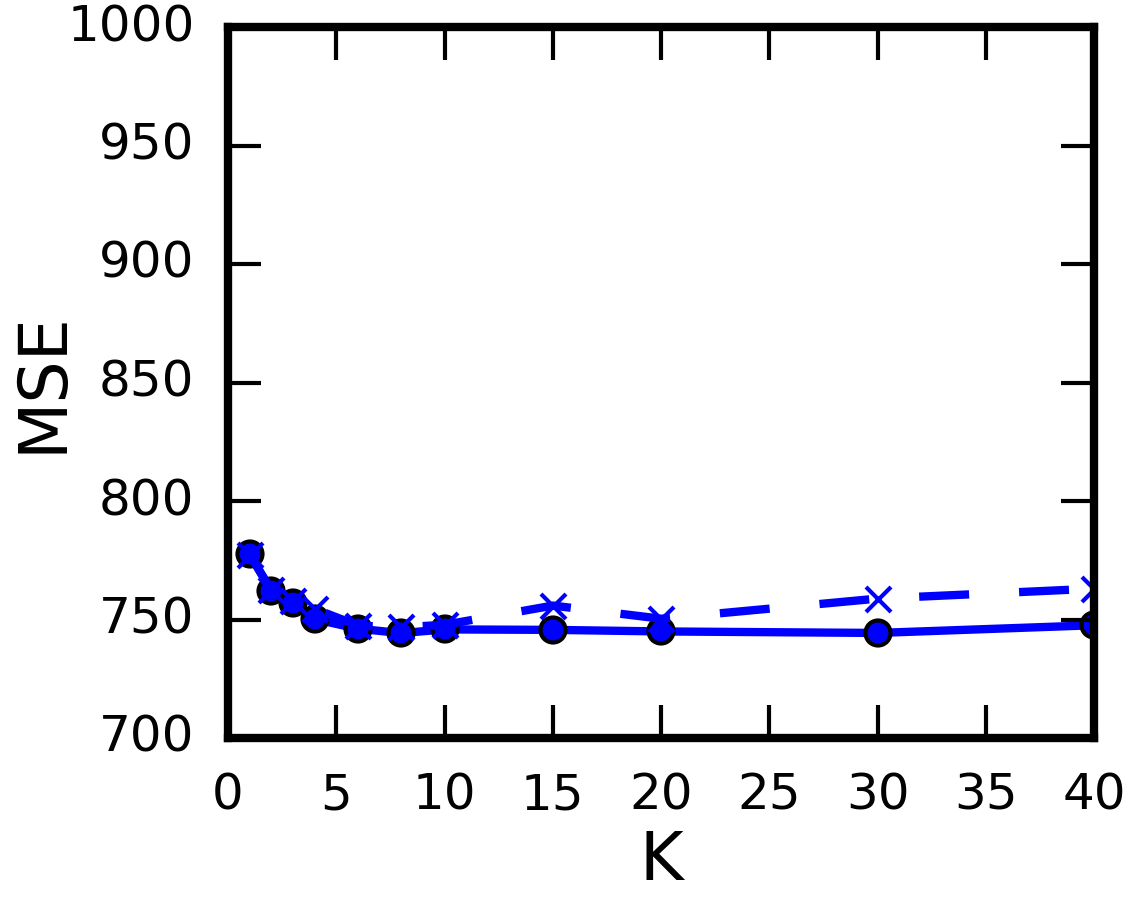}
				\includegraphics[width=0.21\columnwidth]{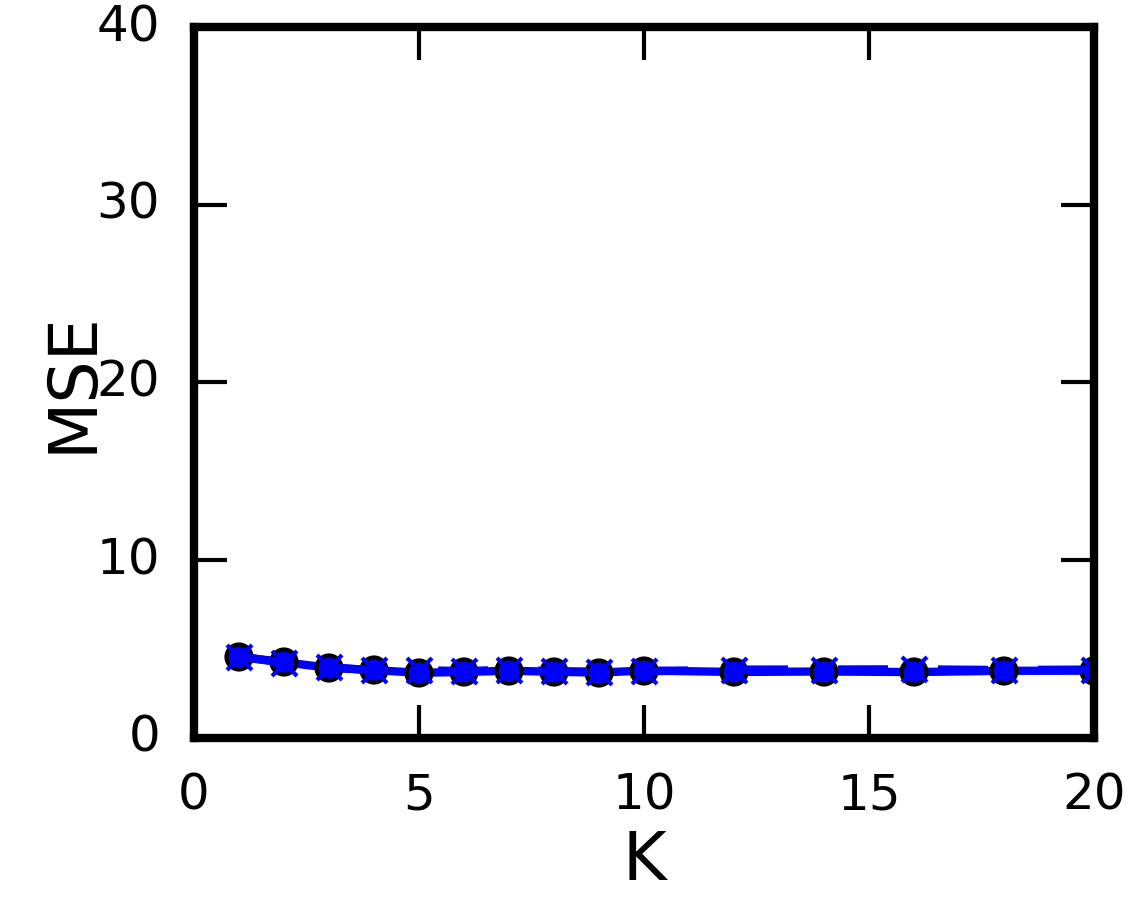}
				\includegraphics[width=0.21\columnwidth]{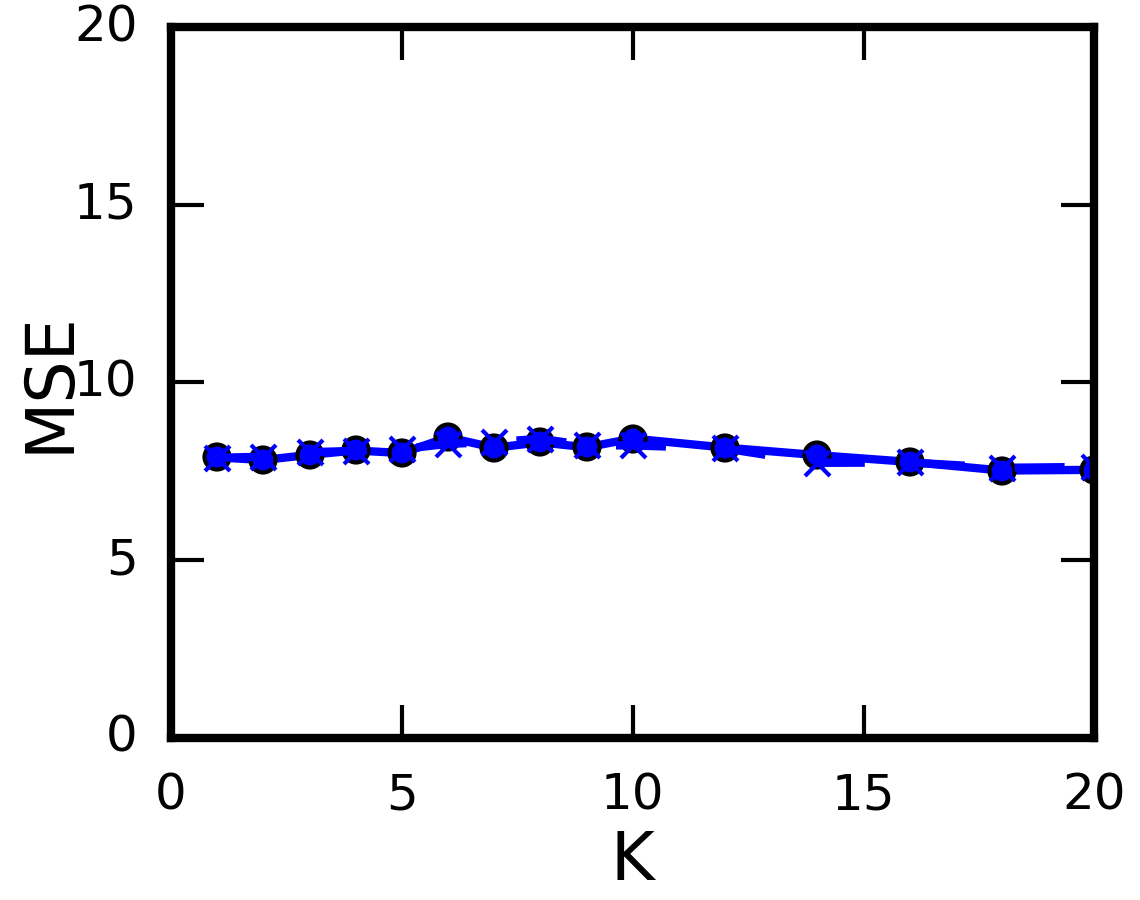}
				\begin{subfigure}{0.21 \columnwidth}
					\includegraphics[width=\columnwidth]{gdsc_nmtf_icm_model_selection.png}
					\captionsetup{width=\columnwidth}
					\caption{GDSC, NMTF}
				\end{subfigure}
				\begin{subfigure}{0.21 \columnwidth}
					\includegraphics[width=\columnwidth]{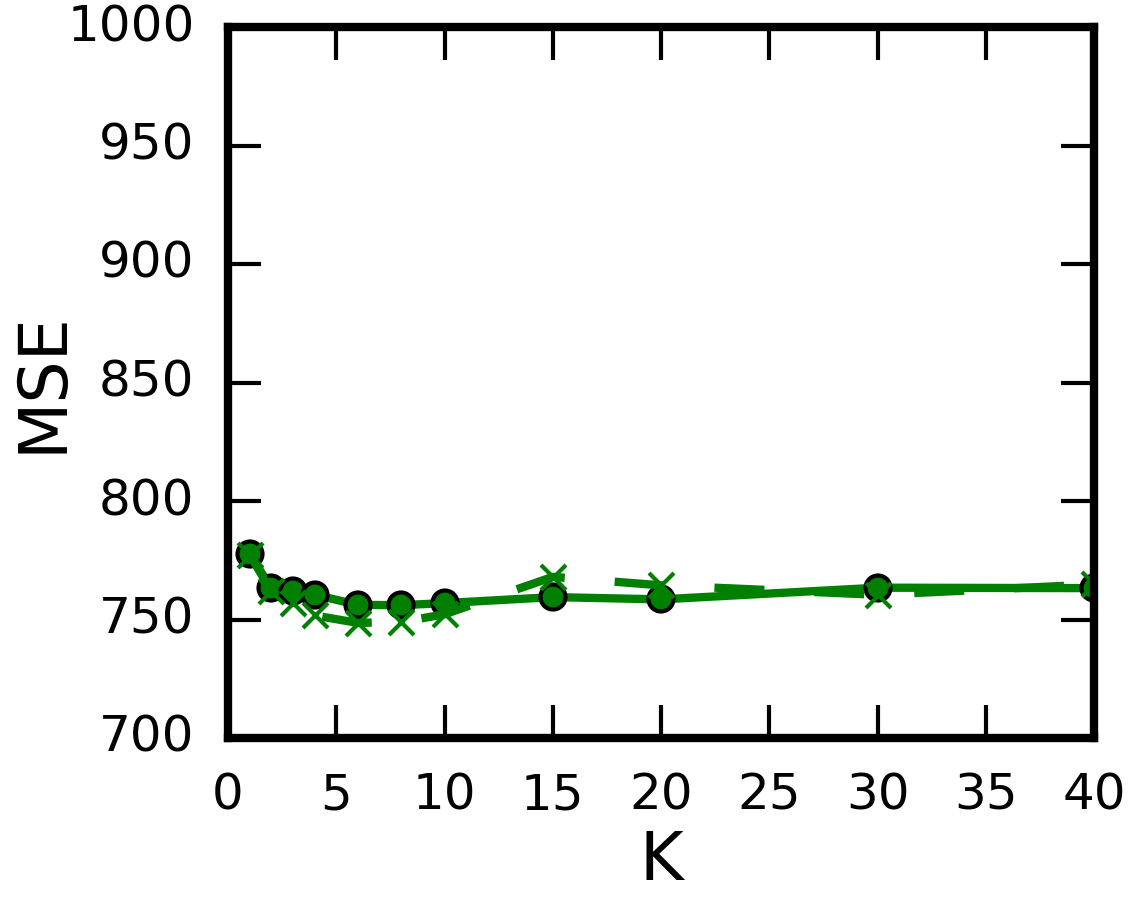}
					\captionsetup{width=\columnwidth}
					\caption{CTRP, NMTF}
				\end{subfigure}
				\begin{subfigure}{0.21 \columnwidth}
					\includegraphics[width=\columnwidth]{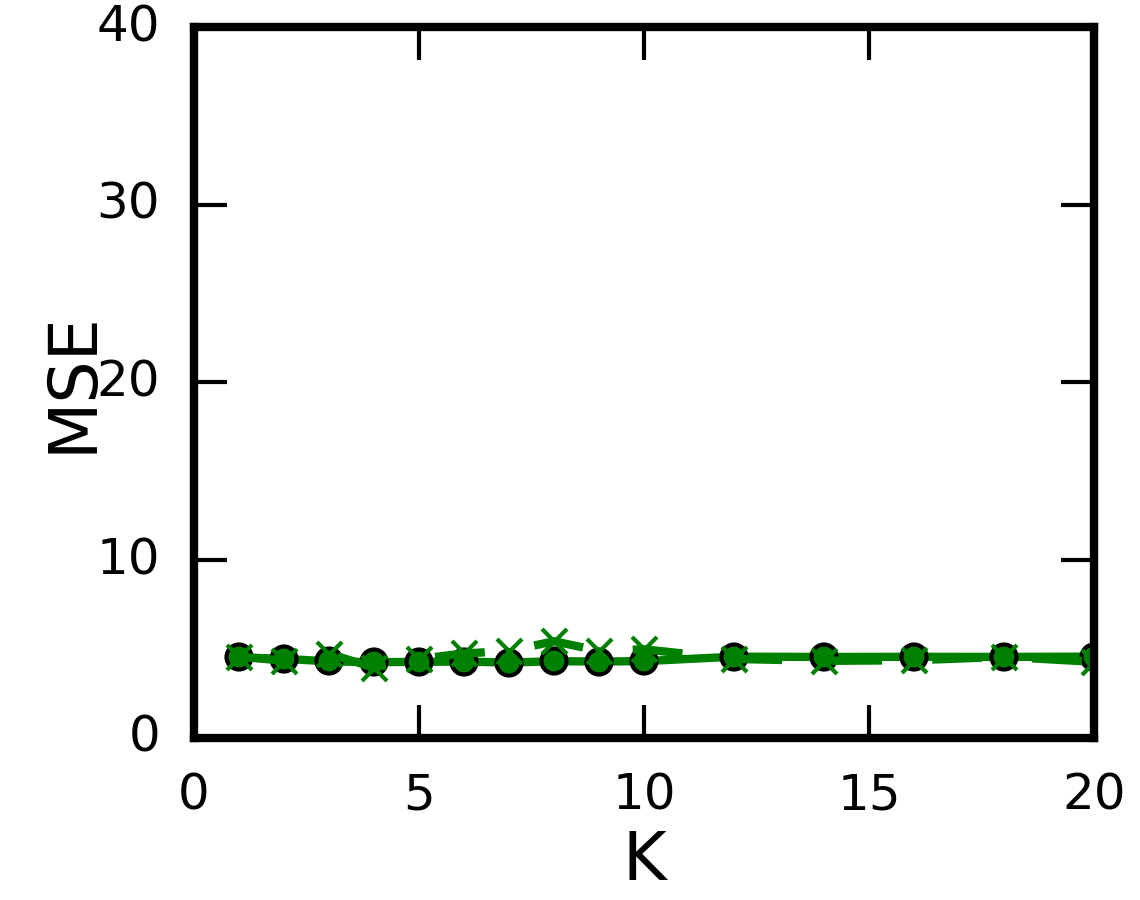}
					\captionsetup{width=\columnwidth}
					\caption{CCLE $IC_{50}$, NMTF}
				\end{subfigure}
				\begin{subfigure}{0.21 \columnwidth}
					\includegraphics[width=\columnwidth]{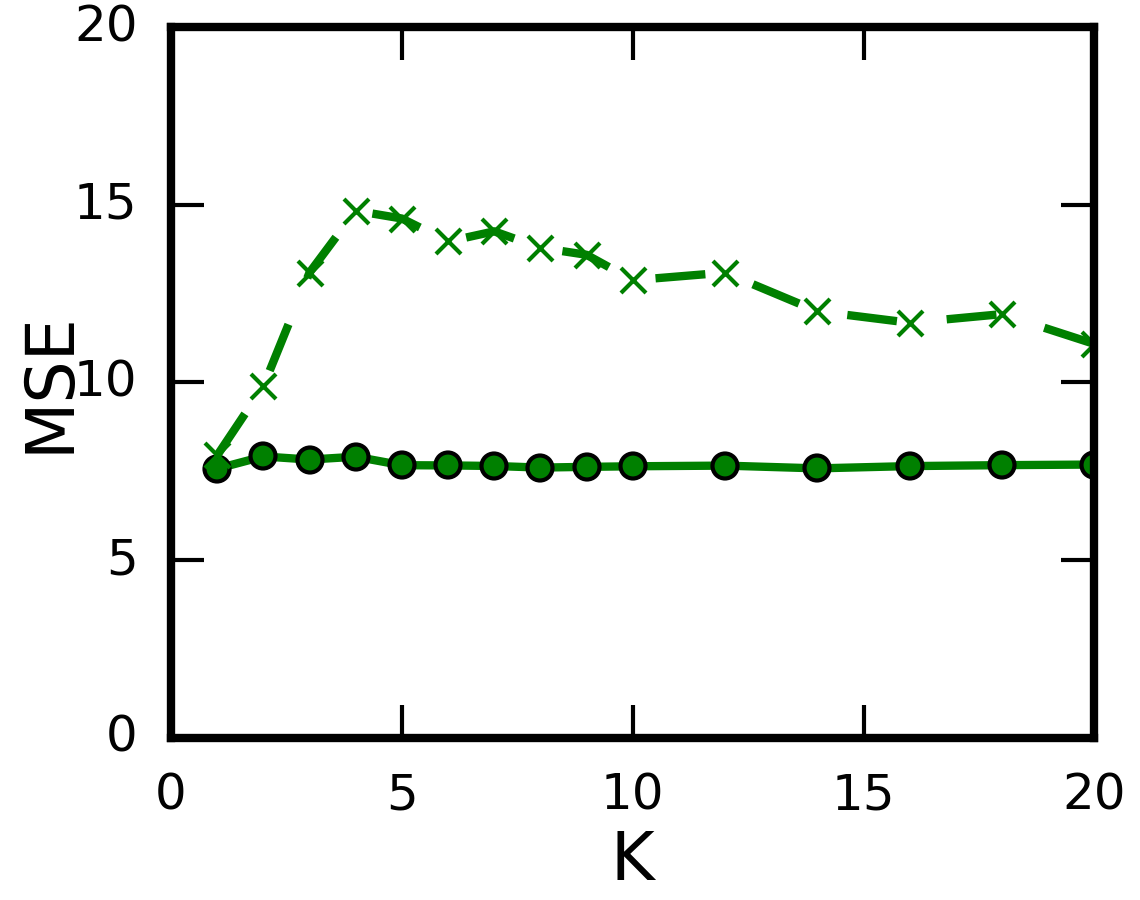}
					\captionsetup{width=\columnwidth}
					\caption{CCLE $EC_{50}$, NMTF}
				\end{subfigure}
				\captionsetup{width=1\columnwidth}
				\caption{10-fold cross-validation performances of the Bayesian NMTF models on the drug sensitivity datasets, where we vary the dimensionality $K$ and $L$ ($L=K$). The top row gives the performances for NMTF VB, the middle row for NMTF Gibbs, and the bottom row for NMTF ICM. Performances for models without ARD are given by dotted lines and crosses (x), with ARD by circles (o).}
				\label{model_selection_nmtf}
			\end{figure*}

			Finally, we conducted the model selection experiment on the remaining three drug sensitivity datasets. Results on all four are given in Figure \ref{model_selection_nmf} for NMF, and Figure \ref{model_selection_nmtf} for NMTF. We can see that the ARD also works very well on the other three datasets, particularly for the ICM approach. On the CCLE $IC_{50}$ and $EC_{50}$ datasets the Gibbs and VB models do not need the ARD to keep a flat line, demonstrating that the fully Bayesian approaches are naturally robust to overfitting already (although it can help---as can be seen on the CTRP and GDSC datasets).
			
%%%%%%%%%%%%%%%%%%%%%%%%%%%%%%%%%%%%%%%%%%%%%%%%%%%%%%%%%%%%%%%%

\clearpage
\section*{Bibliography} 
\bibliography{bibliography}
\bibliographystyle{abbrvnat}